%% file: main.tex
\documentclass[12pt]{book}

\usepackage{geometry}
\usepackage{pdfpages}
\usepackage{graphicx}
\usepackage{fullpage}
\usepackage{amsfonts}
\usepackage{amsmath}
\usepackage{algpseudocode}
\usepackage{algorithm}
\usepackage[round]{natbib}
\usepackage{subcaption}
\usepackage{amssymb}
\usepackage{appendix}
\usepackage{bbm}
\usepackage{multirow}

\usepackage{tikz}
\tikzset{>=latex}

\definecolor{c0}{HTML}{1F77B4}
\definecolor{c1}{HTML}{FF7F0E}
\definecolor{c2}{HTML}{2CA02C}
\definecolor{c3}{HTML}{D62728}
\definecolor{c4}{HTML}{9467BD}

\DeclareSymbolFont{bbold}{U}{bbold}{m}{n}
\DeclareSymbolFontAlphabet{\mathbbold}{bbold}
\newcommand{\ind}{\mathbbold{1}}

\newcommand{\term}[1]{\textbf{#1}}

\DeclareMathOperator*{\argmax}{\arg\!\max}
\DeclareMathOperator*{\argmin}{\arg\!\min}

\newcommand{\cost}{\mathrm{cost}}
\newcommand{\edit}{\mathrm{edit}}

\newcommand{\tail}{\mathrm{tail}}
\newcommand{\head}{\mathrm{head}}
\newcommand{\inedges}{\text{in}}
\newcommand{\outedges}{\text{out}}

\title{Sequence Prediction with Neural Segmental Models}
\author{Hao Tang\\\texttt{haotang@ttic.edu}}
\date{Toyota Technological Institute at Chicago}

\begin{document}

\frontmatter

\input{title}

\pagestyle{empty}
\newgeometry{left=0in,right=0in,top=0in,bottom=0in}
{\centering
\includegraphics[width=8in]{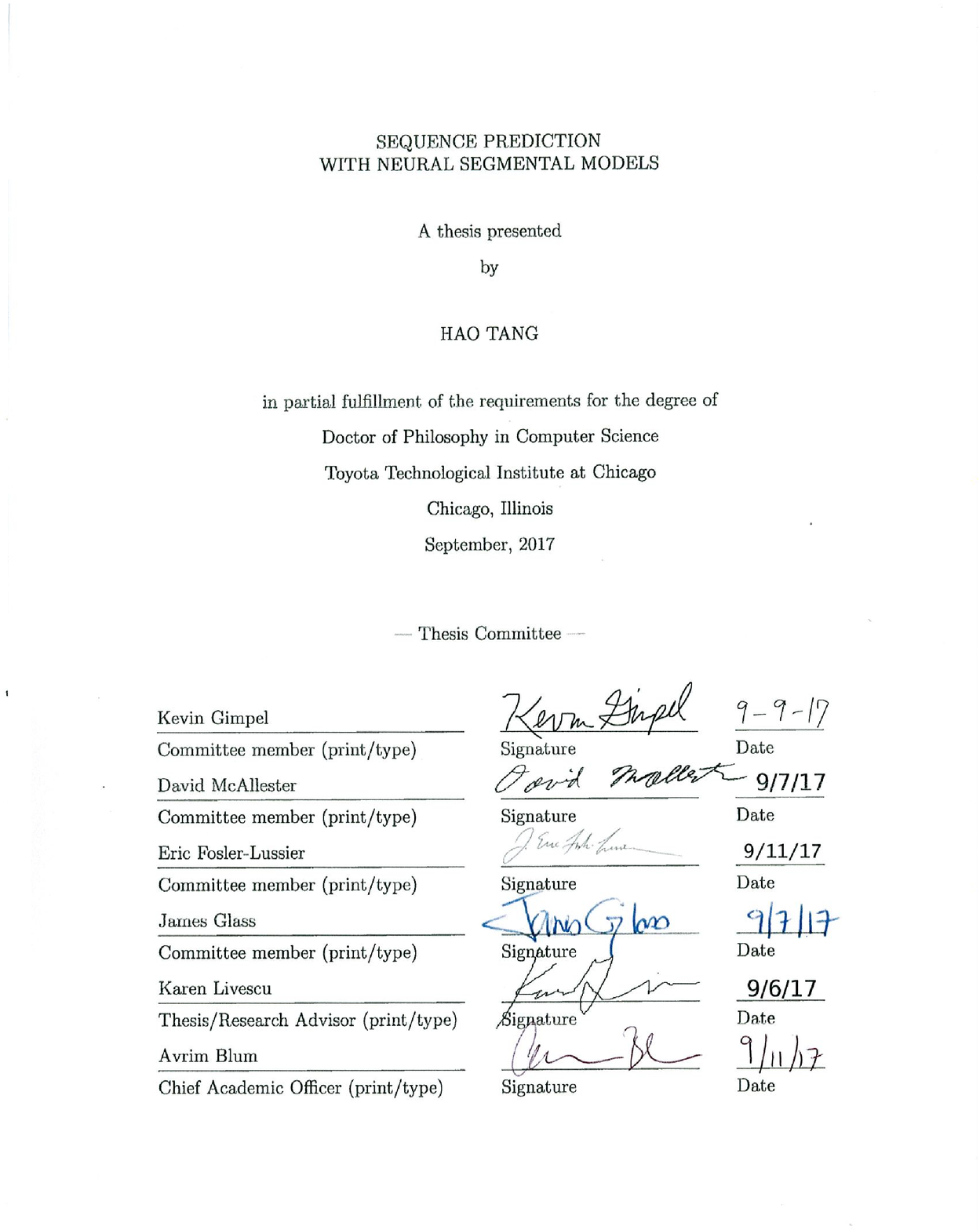}
}
\newpage
\restoregeometry

\pagenumbering{roman}

\input{abstract}

\input{ack}

\tableofcontents

\mainmatter

\include{intro}

\include{background}

\include{seg}

\include{dsc}

\include{e2e}

\include{related}

\include{unified}

\include{conclusion}

\bibliographystyle{plainnat}
\bibliography{string,ref}

\end{document}

%% file: title.tex
\begin{titlepage}

\begin{center}
{\large\bf SEQUENCE PREDICTION \\ WITH NEURAL SEGMENTAL MODELS} \\[1cm]
BY \\[0.16cm]
HAO TANG \\[1cm]
A thesis submitted \\[0.16cm]
in partial fulfillment of the requirements for \\[0.16cm]
the degree of \\[1cm]
Doctor of Philosophy in Computer Science \\[1cm]
at the \\[1cm]
TOYOTA TECHNOLOGICAL INSTITUTE AT CHICAGO \\[0.16cm]
Chicago, Illinois \\[1cm]
September, 2017 \\[3cm]
Thesis Committee: \\
Karen Livescu (Thesis Advisor) \\
Kevin Gimpel \\
David McAllester \\
Eric Fosler-Lussier \\
James Glass
\end{center}

\end{titlepage}

%% file: abstract.tex
\begin{center}
{\Large\bf Sequence Prediction with Neural Segmental Models} \\[0.1cm]
{\large by \\[0.1cm]
Hao Tang}
\end{center}

\vspace{1cm}

\section*{Abstract}

Segments that span contiguous parts of inputs, such as phonemes in
speech, named entities in sentences, actions in videos, occur
frequently in sequence prediction problems.  Segmental models, a class
of models that explicitly hypothesizes segments, have allowed the
exploration of rich segment features for sequence prediction.
However, segmental models suffer from slow decoding, hampering the use
of computationally expensive features.
In this thesis, we introduce discriminative
segmental cascades, a multi-pass inference framework that allows us to
improve accuracy by adding higher-order features and neural segmental
features while maintaining efficiency. 

Segmental models, similarly to conventional speech recognizers,
are typically trained in multiple stages.
In the first stage, a frame classifier is trained,
and in the second stage, segmental models are trained with
the outputs of the frame classifier.
Both training stages require manual alignments,
and obtaining manual alignments are time-consuming and expensive.
We explore end-to-end training for segmental models with
various loss functions, and show how end-to-end training
with marginal log loss can eliminate
the need for detailed manual alignments.

We draw the connections between the marginal log loss
and a popular end-to-end training
approach called connectionist temporal classification,
and present a unifying
framework for various end-to-end graph search-based models, such as
hidden Markov models, connectionist temporal classification, and
segmental models.
Finally, we discuss possible extensions of
segmental models to large-vocabulary sequence prediction tasks.

\vspace{1cm}

\noindent
Thesis Supervisor: Karen Livescu \\
Title: Associate Professor

\newpage

%% file: ack.tex
\section*{Acknowledgements}

First, I would like to thank my advisor, Karen Livescu. Karen has always
been very patient, letting me make mistakes after mistakes, mistakes that
she can foresee, mistakes that I have made many times already, mistakes
that have no easy fix, so that I can learn from them.  She taught me when
to explore novel ideas and when to persist and complete tedious tasks.
I have learned to think critically and argue constructively, while at the
same time to be honest and critical to my own work.  There are
countless other things I have learned from her including phonetics, spectrogram reading,
and English word usage, to name a few.  I can not express how much I appreciate
her guidance and mentorship.  I am really fortunate to be her first
PhD student.

Next, I want to thank my committee members, Kevin Gimpel, David
McAllester, Eric Fosler-Lussier, and Jim Glass. David's comments and
questions always encourage me to view my work in a broader context and in
the most general form possible.  I have also greatly benefited
from his mathematical rigor through his talks, his lectures, and interactions with him,
and have developed the habit of asking
whether a mathematical expression is type checked or not.
I am grateful to Eric for his insightful comments.  I have benefited a lot
from conversations we had in workshops and conferences.
I enjoyed his humorous and joyful attitude, and him
pushing me towards the finish line harder than my advisor does.
I thank Jim for keeping the speech science aspect of my research in mind.
My research is heavily influenced by his prior work,
and it is very special to work on something that he has been working on
for decades.

I am deeply grateful to Kevin.  He is like my second advisor,
caring and helpful in many ways.  I have benefited from his
expertise in natural language processing (NLP), and discovered
many similar approaches developed in parallel in both the NLP
and the speech community.  Besides the technical content,
he taught me to be humble yet to be bold when necessary.
I have also enjoyed his humorous comments on my writing
before stressful deadlines.  It has been a great pleasure
to have his company during my PhD.

I was fortunate to join Mark Hasegawa-Johnson's team in
the second Jelinek summer workshop.  I have benefited from
Mark's extensive knowledge in both speech science and machine learning.
It is fair to say that I always learn something new when talking to him.
I also thank him for his guidance after the workshop.
It has been a great pleasure to work with the people on the team.
In particular, I have benefited from interactions with Chunxi Liu,
Preethi Jyothi, Amit Das, Vimal Manohar, Paul Hager, Tyler Kekona, and Rose Sloan.
People I met during the workshop, including Yi Luan, Yangfeng Ji, Lingpeng Kong,
Guoguo Chen, and Trang Tran, also made the overall experience fun and pleasant.

In 2013, I worked as an intern with Shinji Watanabe
at the Mitsubishi Electric Research Laboratories (MERL).
I have benefited from the interactions with him and other people
at MERL, including John Hershey, Jonathan Le Roux, and Tim Marks.
In particular, I thank Shinji for taking care of me when I had Bell's palsy
during the internship.
The fellow interns at MERL, including Niao He and Lingling Tao,
also made the internship experience fun and rewarding.

My first research experience related to speech and language processing
was acquired from Lin-Shan Lee's guidance in National Taiwan University back in 2009.
Lin-Shan, even with his always busy schedule, has been very caring and helpful.
It would not have been possible for me to pursue a PhD without his help and encouragement.

I thank the people in Toyota Technological Institute at Chicago (TTIC) for making
my PhD journey wonderful.  In particular, I have benefited from Julia Chuzhoy's
and Madhur Tulsiani's lectures, acquiring the necessary vocabulary to talk to theory people.
I thank Nati Srebro for grilling me during
my qualifying exam, forcing me to always keep the precise language and mathematical terms in mind.
I enjoyed the interactions with Greg Shakhnarovich and his humor.
I thank Joseph Keshet for his guidance and mentorship during my first
few years of PhD.
I have benefited interactions and collaboration with Liang Lu.
I thank the fellow students, including Behnam Tavakoli, Somaye Hashemifar,
Taehwan Kim, Shubham Toshniwal, Shane Settle, Qingming Tang, Bowen Shi, Lifu Tu, Hai Wang,
Renyu Zhang, Jian Yao, Jianzhu Ma, Xing Xu, Avleen Bijral, Payman Yadollahpour, Feng Zhao, Zhiyong Wang,
Karthik Sridharan, and Peng Jian, for the interactions, cookie breaks,
random interruptions, and random questions.
I will never regret choosing a cubicle that invites so many interruptions.
I am grateful to Weiran Wang.  It is fair to say that much of my work would not be possible
without his help.
I have also greatly benefited from interactions with past post-docs,
including Raman Arora and Herman Kamper.
I thank visiting students, Taiki Kawano and Takayuki Yamabe,
from Toyota Technological Institute in Japan, for the fun time we had
in Chicago and in Tokyo.
Thanks to Chrissy Novak, Adam Bohlander, and the rest of the staff for making
my PhD life easy and smooth.
Lastly, I thank the president of TTIC, Sadaoki Furui, for his guidance,
and for pushing me to exercise and to practice my tennis skills with him.

Thanks to my family in Taiwan for their support over the years.
Thanks to Hsin-Yu, for everything.  And thanks to my parents,
whom I cannot possibly thank in words.

%% file: intro.tex
\chapter{Introduction}
\label{ch:intro}

\begin{figure}
\begin{center}
\begin{tikzpicture}

\node[left] at (-2, 3) {input};
\node[right] at (1, 3) {a sequence of tokens};

\node[left] (x0) at (0, 0) {\includegraphics[width=4cm]{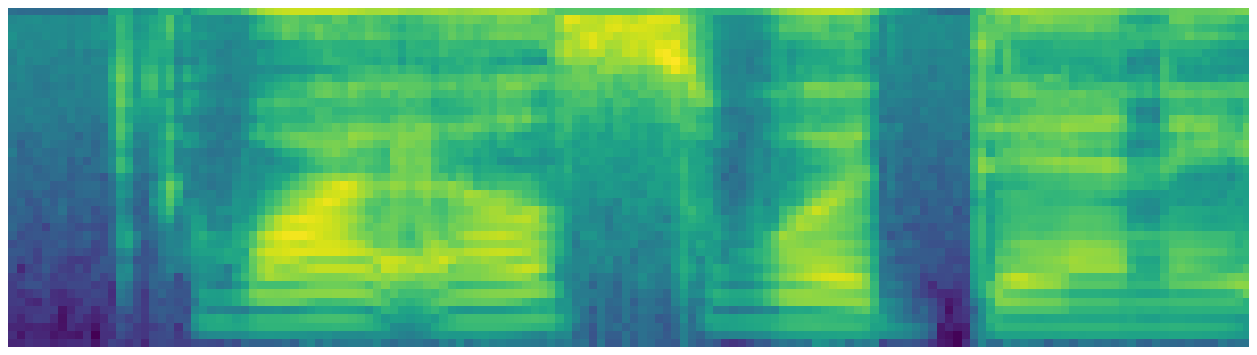}};
\node[right] (x1) at (1, 0) {why else would ...};
\draw[->] (x0) to (x1);

\draw[black!50] (-2.7, -0.6) rectangle (-1.9, 0.6);
\draw[black!50] (2, 0.2) edge[->,bend right] (-2.3, 0.6);

\node[left] (y0) at (0, -3) {
    \includegraphics[width=1cm]{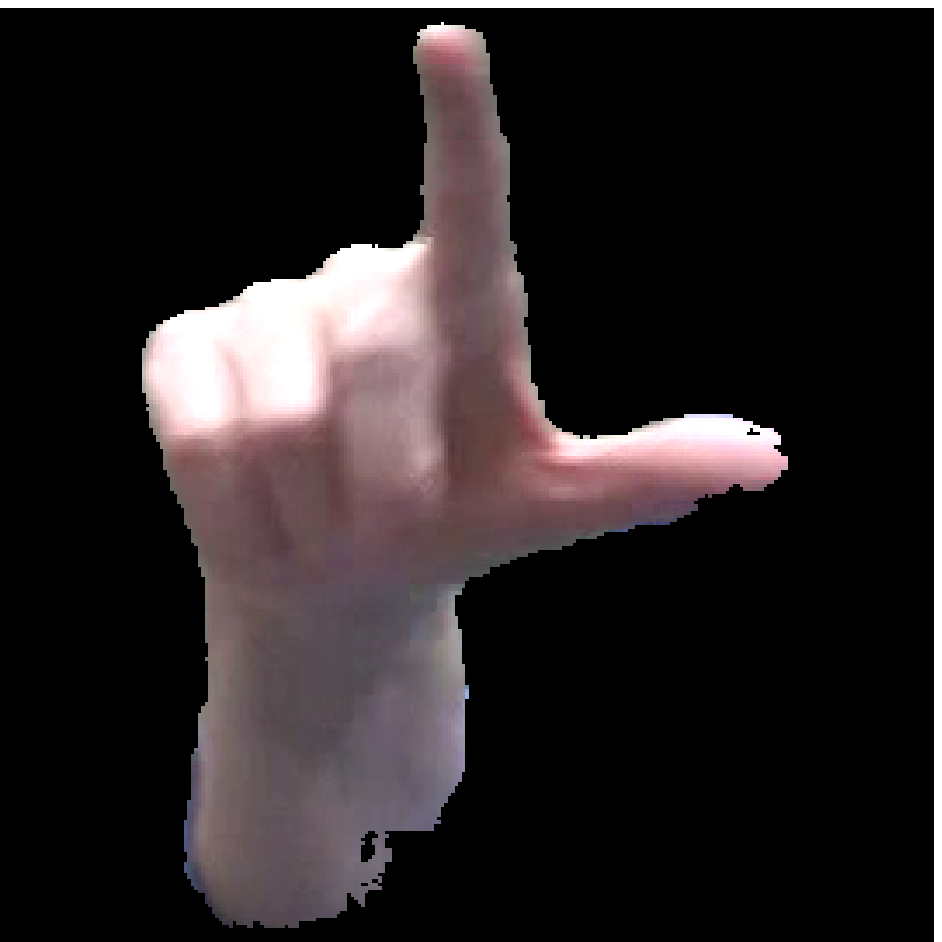}
    \includegraphics[width=1cm]{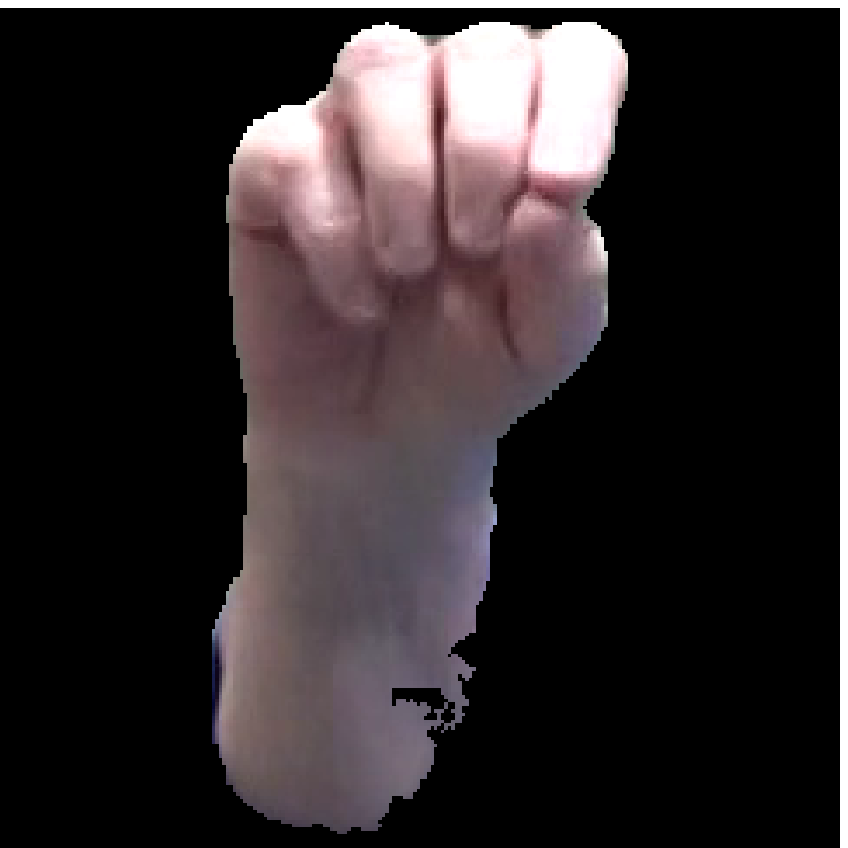}
    \includegraphics[width=1cm]{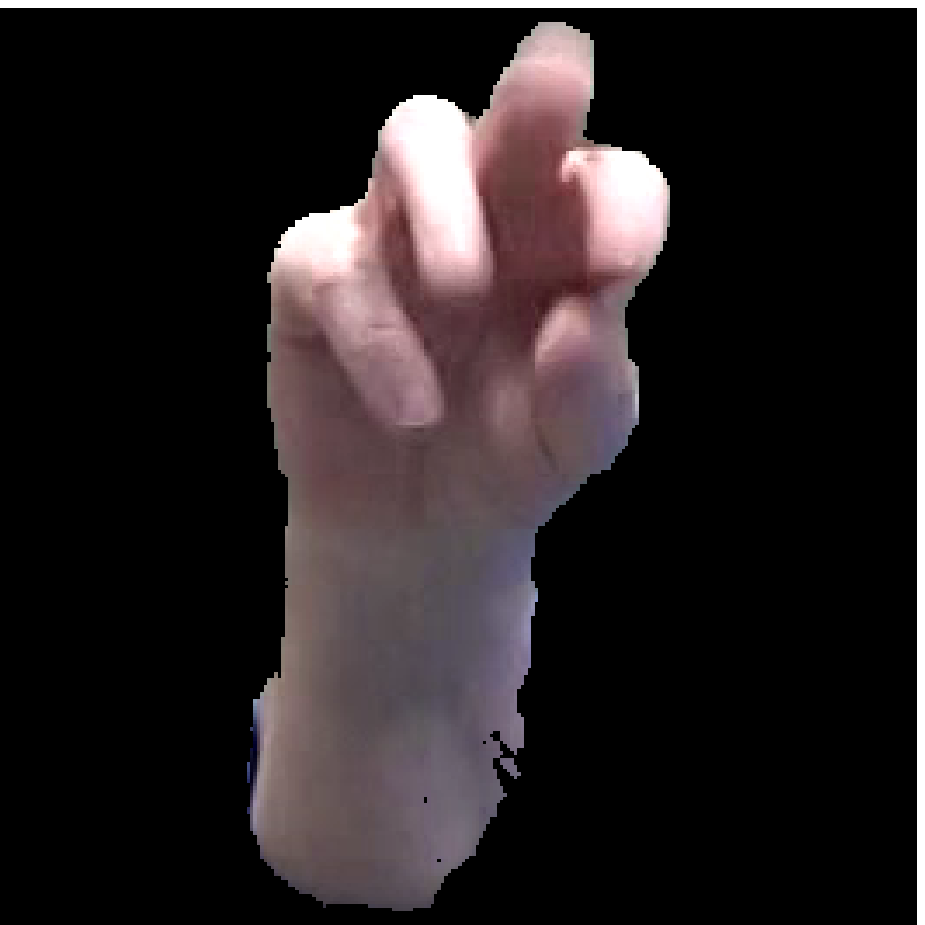}
    \includegraphics[width=1cm]{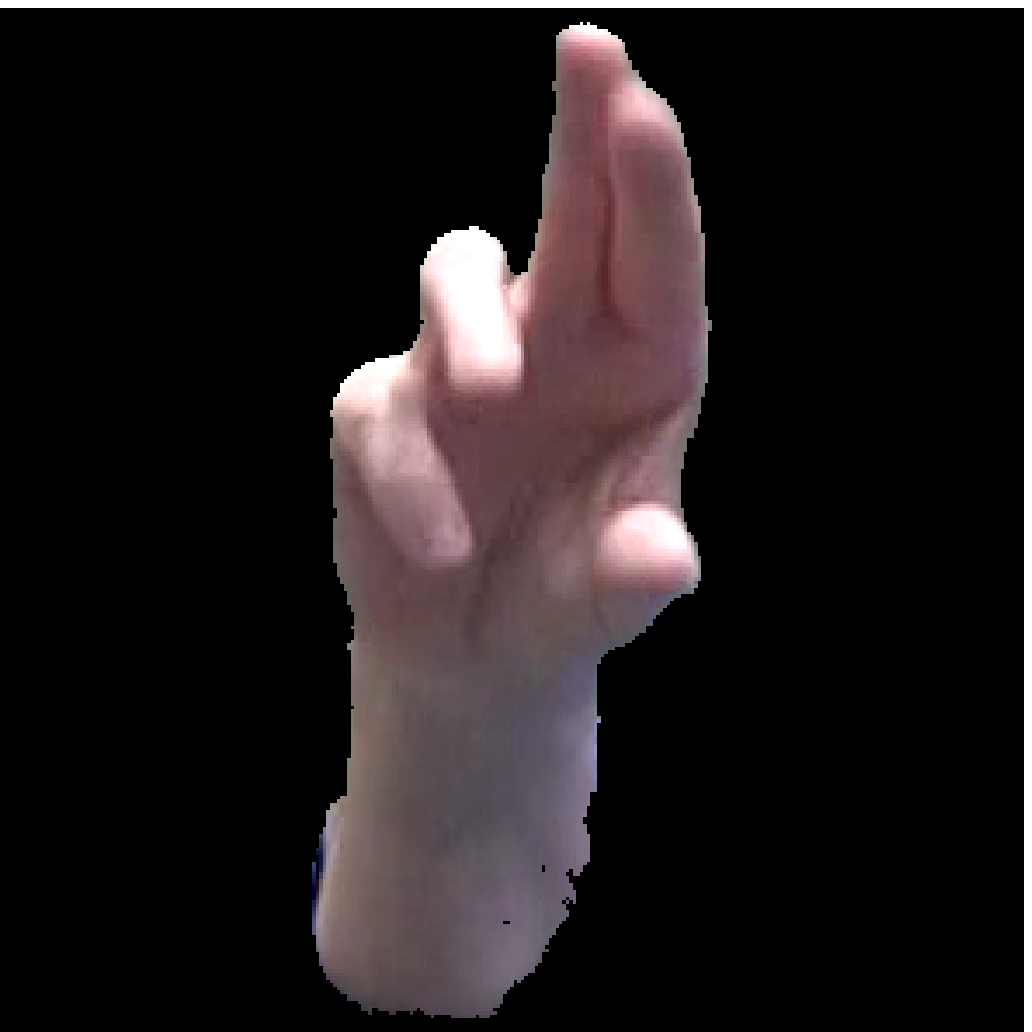}
    \includegraphics[width=1cm]{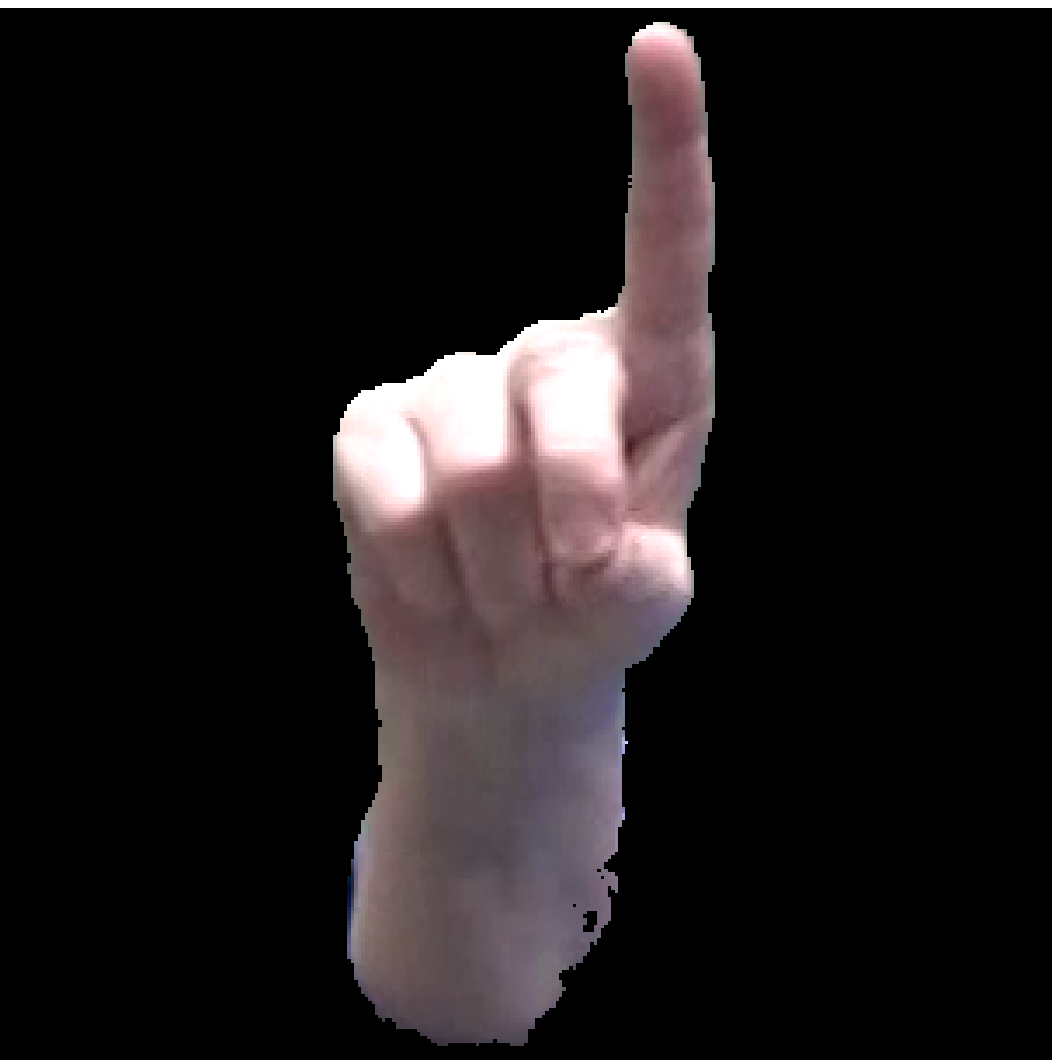}
};

\node[right] (y1) at (1, -3) {L O R D};
\draw[->] (y0) to (y1);

\node[left, font=\footnotesize] (z0) at (0, -6) {\begin{tabular}{l}
     Details emerge in deal to bring \\
     2028 Summer Olympics to \\
     Los Angeles.
     \end{tabular}};

\node[right, font=\footnotesize] (z1) at (1, -6) {\begin{tabular}{l}
     Details emerge in deal to bring \\
     $(\text{2028 Summer Olympics})_\text{event}$ to \\
     $(\text{Los Angeles})_{\text{place}}$.
     \end{tabular}};

\draw[->] (z0) to (z1);

\end{tikzpicture}
\end{center}
\caption{Examples of sequence prediction tasks.
    The first example is speech recognition, where the input is a sequence of acoustic
    signals and the output is a sequence of words.  The second example is American Sign
    Language fingerspelling recognition, where the input is a sequence of images and
    the output is a sequence of letters.  The third example is named entity recognition,
    where the input is a sequence of words and the output is the same sequence of words
    with named entities in parentheses.  In the example of speech recognition,
    the gray arrow connects the word ``else'' to its corresponding contiguous part
    in the input sequence.}
\label{fig:seq-pred}
\end{figure}

Segmental models, the subject of this thesis, are a family of models
that predict sequences of segments.
Segmental models are designed for a wide range of sequence prediction tasks,
such as named entity recognition \citep{SC2005}, speech recognition \citep{ODK1996, G2003},
and action recognition \citep{SNDC2010,HLD2011}.
Examples of these tasks are shown in Figure \ref{fig:seq-pred}.
The input of a sequence prediction task is commonly represented as
a sequence of real-valued vectors, and the output is a sequence of
discrete labels.
The goal is to find a function that maps the input vectors to
the output labels.
Every output label in the output sequence, such as a named entity, a phoneme,
or an action, typically has
a corresponding chunk of contiguous input vectors, called a segment.
Segments come in varying lengths.
As a consequence,
the lengths of the output sequences typically do not match
the lengths of the input sequences.

The predominant approach to solving these tasks is to break the varying-length
segments into pieces so that the lengths of the input sequences match the lengths of
the output sequences.  These smaller pieces are then
assembled back into varying-length segments with an additional
post-processing step.
Each individual element in the input sequence is referred to as a \term{frame}.
Models,
such as hidden Markov models \citep{R1989,J1998},
linear-chain conditional random fields \citep{GMAP2005,MF2009},
and recurrent neural networks \citep{GS2005},
that map input sequences to output sequences of the same lengths,
are called \term{frame-based models} \citep{ODK1996,G2003}.
To include a wider context, it is common to use windows of frames
instead of individual frames for prediction.
Nevertheless, the number of windowed frames is still
the same as the number of output labels.
Because frame-based models are
conceptually simple and computationally cheap,
they have enjoyed much success and received much attention
in the past few decades.

\begin{figure}
\begin{center}
\begin{subfigure}[b]{0.9\textwidth}
    \begin{center}
    \begin{tikzpicture}
    \node at (0, 0) {\includegraphics[height=2.5cm]{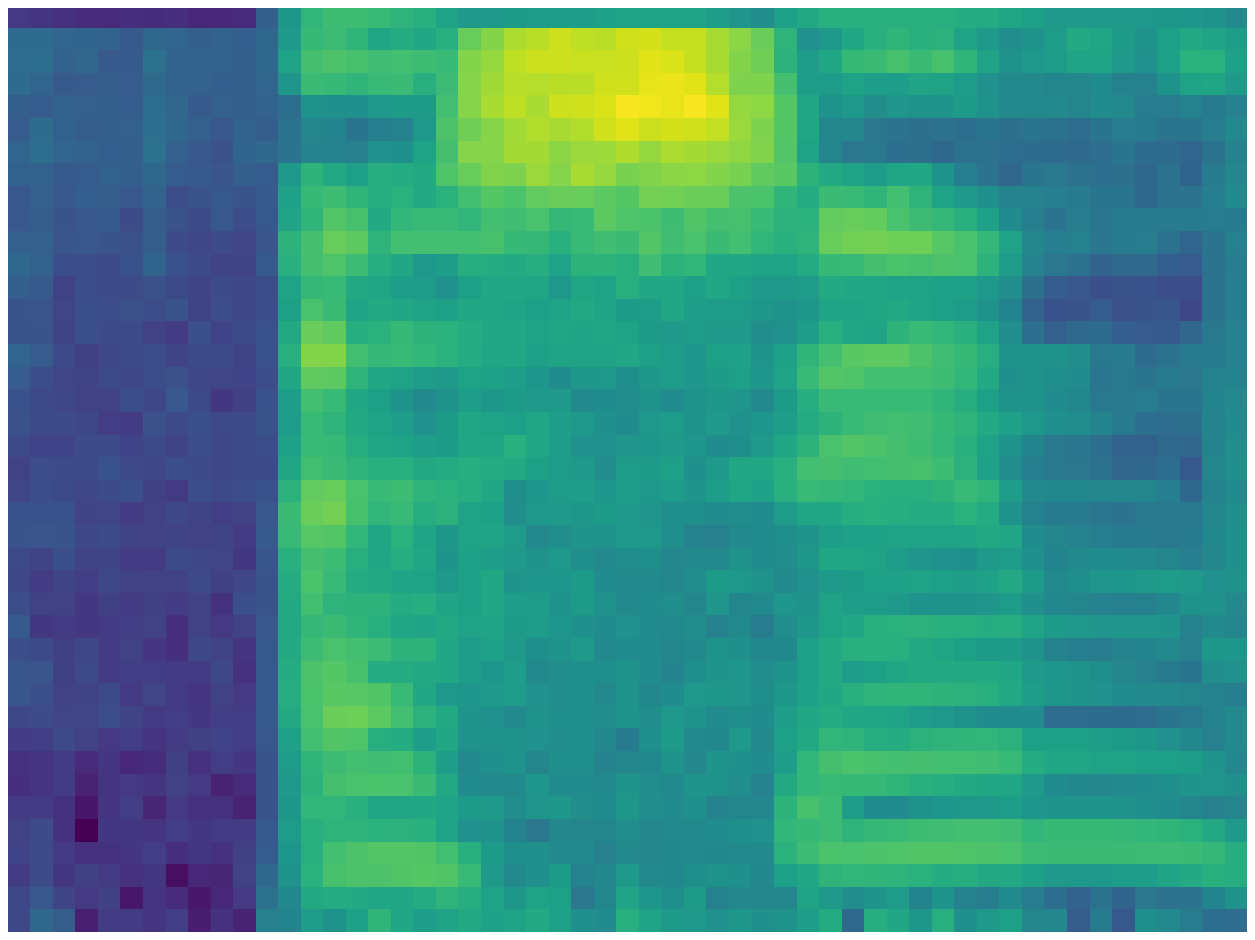}};
    \end{tikzpicture}
    \caption{A sequence of input frames.}
    \end{center}
    \vspace{0.5cm}
\end{subfigure}
\begin{subfigure}[b]{0.9\textwidth}
    \begin{center}
    \begin{tikzpicture}
    \node at (0, 0) {\includegraphics[height=2.5cm]{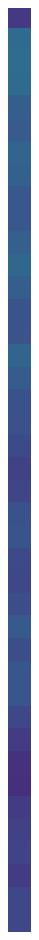}};
    \node at (0.5, 0) {\includegraphics[height=2.5cm]{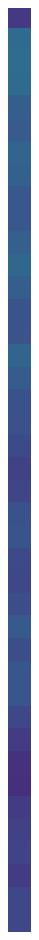}};
    \node at (1, 0) {\includegraphics[height=2.5cm]{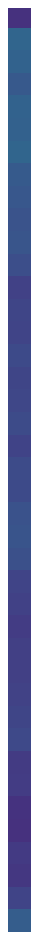}};
    \node at (1.5, 0) {$\dots$};
    \node at (2, 0) {\includegraphics[height=2.5cm]{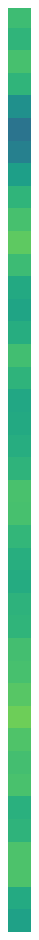}};
    \node at (2.5, 0) {\includegraphics[height=2.5cm]{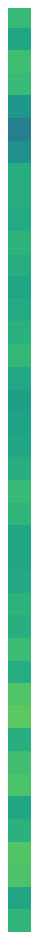}};
    \node at (3, 0) {\includegraphics[height=2.5cm]{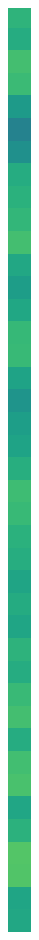}};
    \node at (3.5, 0) {$\dots$};
    \node at (4, 0) {\includegraphics[height=2.5cm]{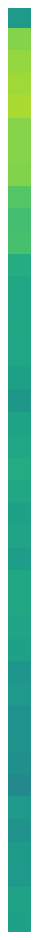}};
    \node at (4.5, 0) {\includegraphics[height=2.5cm]{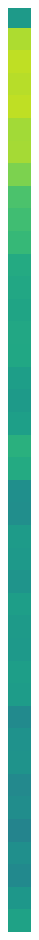}};
    \node at (5, 0) {\includegraphics[height=2.5cm]{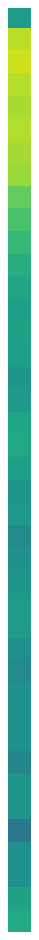}};
    \node at (5.5, 0) {$\dots$};
    \node at (6, 0) {\includegraphics[height=2.5cm]{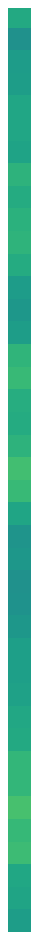}};
    \node at (6.5, 0) {\includegraphics[height=2.5cm]{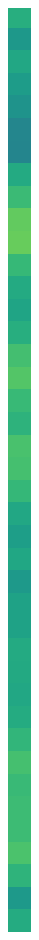}};
    \node at (7, 0) {\includegraphics[height=2.5cm]{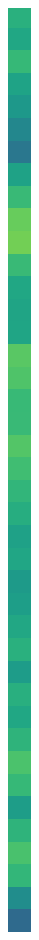}};
    \node at (7.5, 0) {$\dots$};
    \node at (8, 0) {\includegraphics[height=2.5cm]{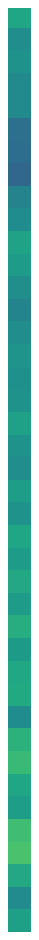}};
    \node at (8.5, 0) {\includegraphics[height=2.5cm]{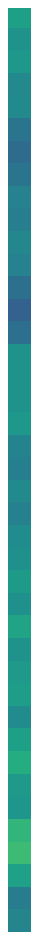}};
    \node at (9, 0) {\includegraphics[height=2.5cm]{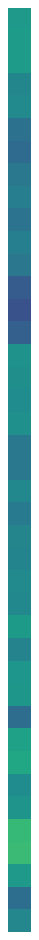}};
    
    \node at (0, -1.5) {sil};
    \node at (0.5, -1.5) {sil};
    \node at (1, -1.5) {sil};
    
    \node at (2, -1.5) {ax};
    \node at (2.5, -1.5) {ax};
    \node at (3, -1.5) {ax};
    
    \node at (4, -1.5) {s};
    \node at (4.5, -1.5) {s};
    \node at (5, -1.5) {s};
    
    \node at (6, -1.5) {ux};
    \node at (6.5, -1.5) {ux};
    \node at (7, -1.5) {ux};
    
    \node at (8, -1.5) {m};
    \node at (8.5, -1.5) {m};
    \node at (9, -1.5) {m};
    \end{tikzpicture}
    \caption{A frame-based approach with
        a label sequence of the same length as the input frames.}
    \end{center}
    \vspace{0.5cm}
\end{subfigure}
\begin{subfigure}[b]{0.9\textwidth}
    \begin{center}
    \begin{tikzpicture}
    \node at (0, 0) {\includegraphics[height=2.5cm]{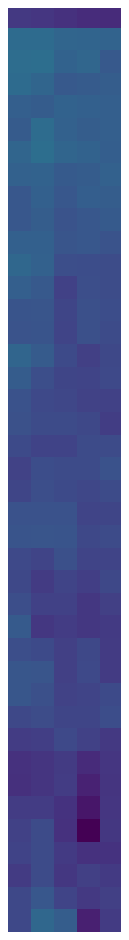}};
    \node at (0.5, 0) {\includegraphics[height=2.5cm]{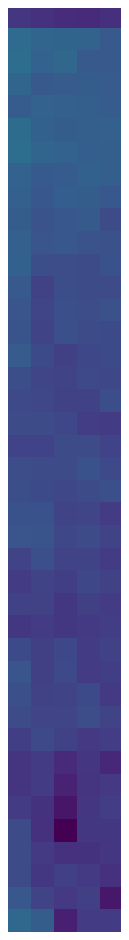}};
    \node at (1, 0) {\includegraphics[height=2.5cm]{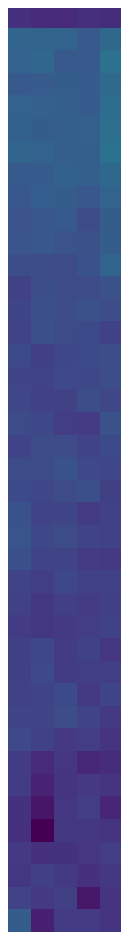}};
    \node at (1.5, 0) {$\dots$};
    \node at (2, 0) {\includegraphics[height=2.5cm]{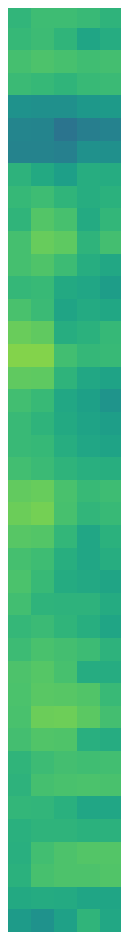}};
    \node at (2.5, 0) {\includegraphics[height=2.5cm]{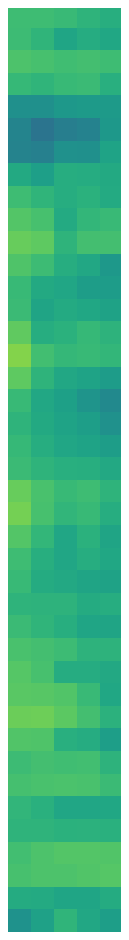}};
    \node at (3, 0) {\includegraphics[height=2.5cm]{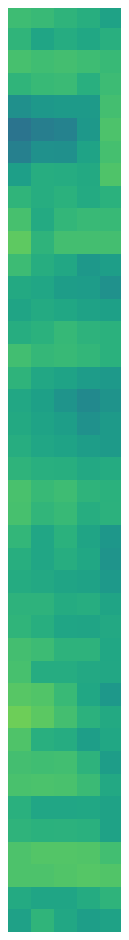}};
    \node at (3.5, 0) {$\dots$};
    \node at (4, 0) {\includegraphics[height=2.5cm]{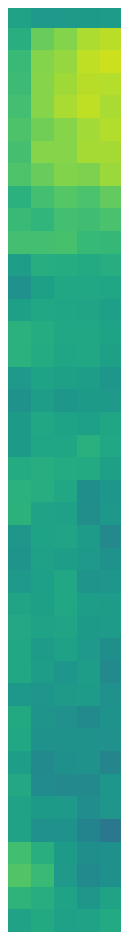}};
    \node at (4.5, 0) {\includegraphics[height=2.5cm]{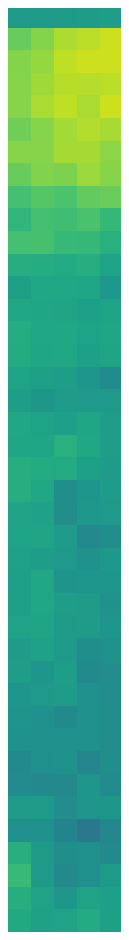}};
    \node at (5, 0) {\includegraphics[height=2.5cm]{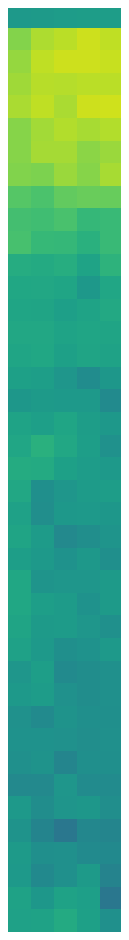}};
    \node at (5.5, 0) {$\dots$};
    \node at (6, 0) {\includegraphics[height=2.5cm]{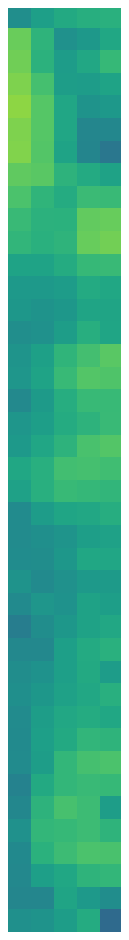}};
    \node at (6.5, 0) {\includegraphics[height=2.5cm]{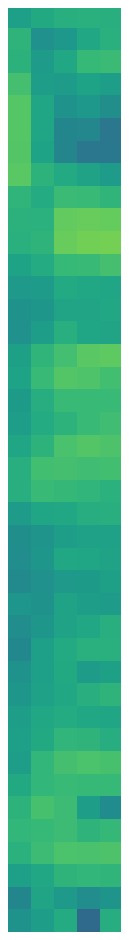}};
    \node at (7, 0) {\includegraphics[height=2.5cm]{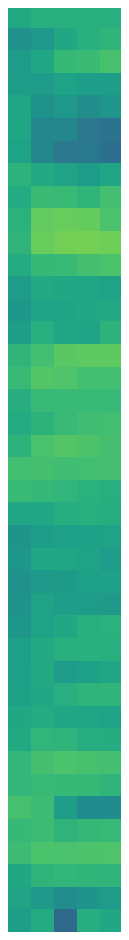}};
    \node at (7.5, 0) {$\dots$};
    \node at (8, 0) {\includegraphics[height=2.5cm]{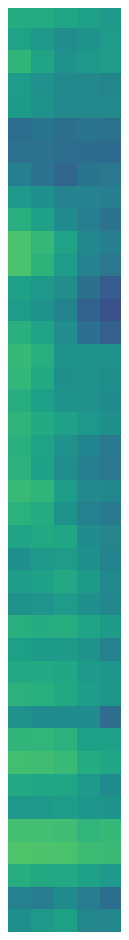}};
    \node at (8.5, 0) {\includegraphics[height=2.5cm]{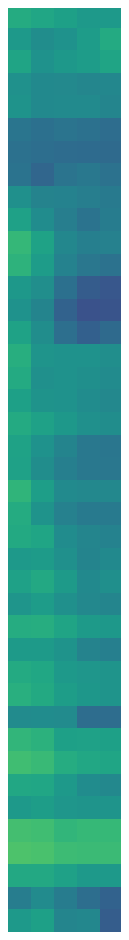}};
    \node at (9, 0) {\includegraphics[height=2.5cm]{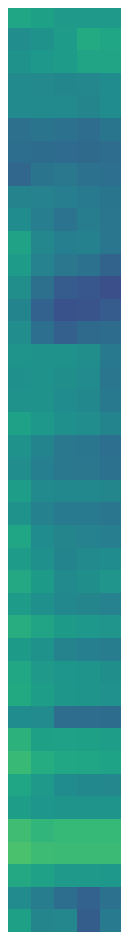}};
    
    \node at (0, -1.5) {sil};
    \node at (0.5, -1.5) {sil};
    \node at (1, -1.5) {sil};
    
    \node at (2, -1.5) {ax};
    \node at (2.5, -1.5) {ax};
    \node at (3, -1.5) {ax};
    
    \node at (4, -1.5) {s};
    \node at (4.5, -1.5) {s};
    \node at (5, -1.5) {s};
    
    \node at (6, -1.5) {ux};
    \node at (6.5, -1.5) {ux};
    \node at (7, -1.5) {ux};
    
    \node at (8, -1.5) {m};
    \node at (8.5, -1.5) {m};
    \node at (9, -1.5) {m};
    \end{tikzpicture}
    \caption{A frame-based approach with
        a label sequence of the same the length as the windowed frames.}
    \end{center}
    \vspace{0.5cm}
\end{subfigure}
\begin{subfigure}[b]{0.9\textwidth}
    \begin{center}
    \begin{tikzpicture}
    \node at (0, 0) {\includegraphics[height=2.5cm]{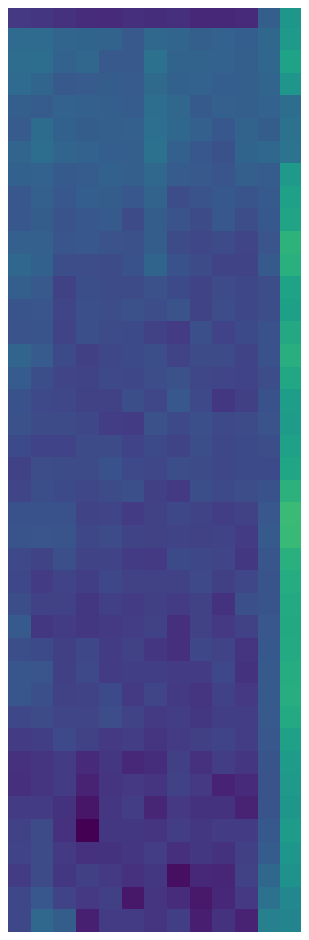}};
    \node at (2, 0) {\includegraphics[height=2.5cm]{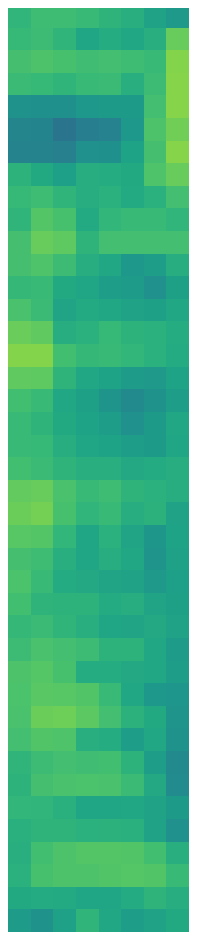}};
    \node at (4, 0) {\includegraphics[height=2.5cm]{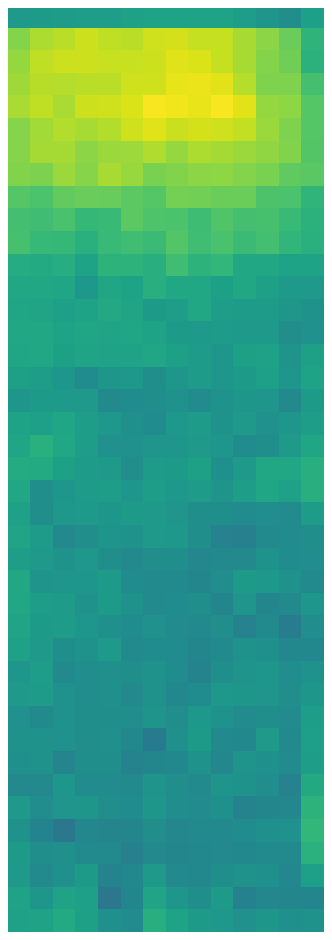}};
    \node at (6, 0) {\includegraphics[height=2.5cm]{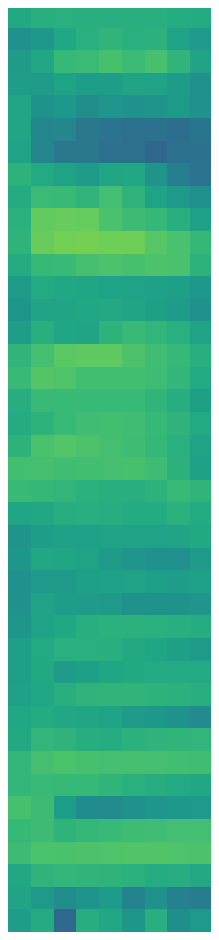}};
    \node at (8, 0) {\includegraphics[height=2.5cm]{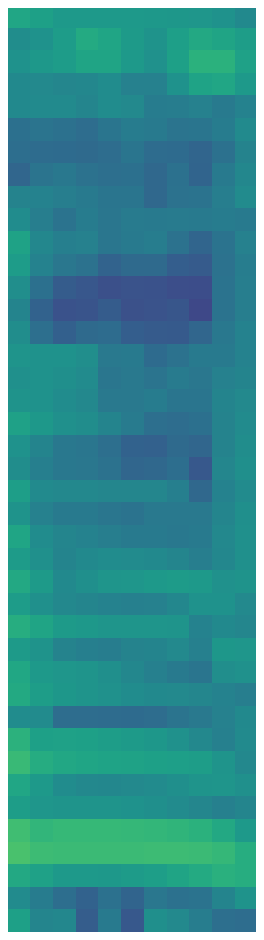}};
    
    \node at (0, -1.5) {sil};
    \node at (2, -1.5) {ax};
    \node at (4, -1.5) {s};
    \node at (6, -1.5) {ux};
    \node at (8, -1.5) {m};
    \end{tikzpicture}
    \caption{A segment-based approach with
        a label sequence aligned to varying-length segments
        in the input frames.}
    \end{center}
\end{subfigure}
\end{center}
\caption{Frame-based approaches and segment-based approaches for sequence prediction.}
\label{fig:frame-vs-seg}
\end{figure}

Model accuracies for prediction tasks are heavily influenced
by the set of features used to make prediction,
where a feature is a real-value function of the input that correlates well with a label
or a subset of labels.
Breaking up varying-length segments is an extremely successful heuristic,
but it also makes extracting certain features from segments difficult.
For example, computing durations is difficult for frame-based models.
Variants of frame-based models, such as variable-duration HMMs \citep{F1980},
are proposed in order to overcome
this difficulty, but it is impractical to construct a model
for every new type of feature.
In contrast, segment-based approaches do not assume anything about
the segments, so the users are free to use arbitrary information
about the segments during prediction.
For example, given a segment with a precise start time and end time,
computing its duration is trivial.
For speech recognition, energy distribution at the left and right boundaries
are useful features.
For named entity recognition, we can check whether
a phrase (segment) is in the dictionary.
The flexibility makes segment-based models appealing,
and many promising results have been reported in the past \citep{ZGPS1989,ODK1996,G2003}.
A comparison of the two approaches is shown in Figure \ref{fig:frame-vs-seg}.

\begin{figure}
\begin{center}
\begin{tikzpicture}[ver/.style={circle,draw}]

\node[ver, ultra thick] (x0) at (0, 0) {};
\node[ver] (x1) at (2, 0) {};
\node[ver] (x2) at (4, 0) {};
\node[ver] (x3) at (8, 0) {};
\node[ver, double] (x4) at (10, 0) {};

\node[ver] (x5) at (6, 1) {};
\node[ver] (x6) at (8, 1) {};

\draw[->] (x0) edge[out=30, in=150] node [above, font=\strut] {s} (x1);
\draw[->] (x1) edge node [above, font=\strut] {p} (x2);
\draw[->] (x2) edge[very thick] node [above, font=\strut] {iy} (x3);
\draw[->] (x3) edge node [above, font=\strut] {ch} (x4);

\draw[->] (x0) edge[out=-30, in=210] node [below, font=\strut] {z} (x1);
\draw[->] (x2) edge[out=90, in=180] node [above left, font=\strut] {ih} (x5);
\draw[->] (x5) edge node [above, font=\strut] {t} (x6);
\draw[->] (x6) edge[out=0, in=90] node [above right, font=\strut] {sh} (x4);

\node at ( 0.5, -3) {$\begin{bmatrix}\vdots\end{bmatrix}$};
\node at ( 1.5, -3) {$\begin{bmatrix}\vdots\end{bmatrix}$};
\node at ( 2.5, -3) {$\begin{bmatrix}\vdots\end{bmatrix}$};
\node at ( 3.5, -3) {$\begin{bmatrix}\vdots\end{bmatrix}$};
\node at ( 4.5, -3) {$\begin{bmatrix}\vdots\end{bmatrix}$};
\node at ( 5.5, -3) {$\begin{bmatrix}\vdots\end{bmatrix}$};
\node at ( 6.5, -3) {$\begin{bmatrix}\vdots\end{bmatrix}$};
\node at ( 7.5, -3) {$\begin{bmatrix}\vdots\end{bmatrix}$};
\node at ( 8.5, -3) {$\begin{bmatrix}\vdots\end{bmatrix}$};
\node at ( 9.5, -3) {$\begin{bmatrix}\vdots\end{bmatrix}$};

\node at ( 0.5, -4) {$x_1$};
\node at ( 1.5, -4) {$x_2$};
\node at ( 2.5, -4) {     };
\node at ( 3.5, -4) {     };
\node at ( 4.5, -4) {$x_s$};
\node at ( 5.5, -4) {     };
\node at ( 6.5, -4) {     };
\node at ( 7.5, -4) {$x_t$};
\node at ( 8.5, -4) {     };
\node at ( 9.5, -4) {$x_T$};

\draw (4, -2.5) -- (4, -2.3) -- (8, -2.3) -- (8, -2.5);
\draw[->] (6, -2.3) to node [right] {$w((\text{iy}, s, t))$} (6, 0);

\end{tikzpicture}
\caption{An example of a segmental model. A search space is built based on the
    input frames. Each edge (segment) has a start time, an end time, and a label,
    which the weight function can make use of. Once the weights of edges
    are computed, decoding is the problem of finding the maximum-weight path.}
\label{fig:seg}
\end{center}
\end{figure}
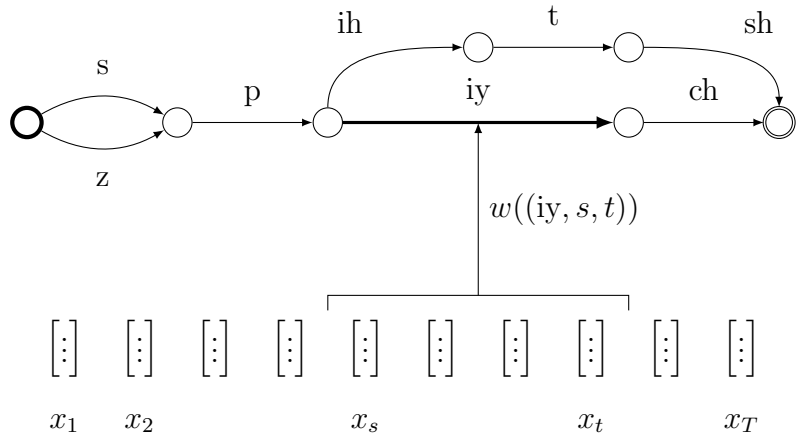

More formally, a segment is a tuple of a start time, an end time,
and possibly a label.
At a high level, segmental models make predictions by first
creating a search space consisting of connected segments, or paths.
Segments are given weights based on the start time, end time,
and label.  The weight values reflect how well the labels match the input.
Finding the sequence of segments that best matches the input
is equivalent to finding the maximum-weight path.
The prediction process is also referred to as \term{decoding}.
Note that the segment weights can be computed by any function
as long as it only makes use of the start time, the end time, the label,
and the input sequence.
An example is shown in Figure \ref{fig:seg}.

Despite their success and attractive properties, segmental models
still fall behind frame-based models on many tasks,
such as speech recognition \citep{Z2012}.
There are many possible reasons behind this.
The choice of features extracted from the segments
might play a role.
However, complex features are typically computationally expensive.
The large number of segments we need to consider
prohibits the use of computationally expensive segment features.
The central goal of this thesis is to improve the efficiency of
segmental models, allowing us to explore more computationally
expensive features and to improve the performance of
segmental models.

This thesis studies segmental models along three directions:
segment features, inference, and learning.
We introduce discriminative segmental cascades to speed up
inference, allowing us to explore computationally expensive
segment representations, such as ones computed with neural networks.
We study how segmental models perform when
they are trained with various loss functions.
Different loss functions have different training requirements,
some of them allowing us to train segmental models
without manual alignments.
We also study whether it is possible to train segmental models
end to end from random initialization.
We focus on two tasks, phonetic recognition and American Sign Language
fingerspelling recognition.

\section{Preliminaries}

In this section, we briefly describe
automatic speech recognition (ASR), a task for which segmental models
have been studied extensively.
We describe a few features that were shown to be useful
for disambiguating phonemes, the basic sound units that distinguish words.
We then describe why it is hard to incorporate these
features in frame-based models, and some of the past efforts
to overcome this difficulty.

\subsection{Automatic speech recognition}

In brief, human speech production is the process of mapping
a sequence of discrete linguistic units, such as words or phonemes, into waveforms,
and automatic speech recognition is about inverting
this process, mapping speech waveforms to sequences of discrete linguistic units.
On one side of the process, we have
speech waveforms, commonly represented as sequences of real-valued
vectors called frames.
On the other side of the process, we have
the discrete units, namely segments, representing
phonemes or words.

A waveform can be represented as a sequence
of real-valued vectors in multiple ways.
One common representation is the amount of sinusoids
at different frequencies appearing in the signal.
A waveform is first converted into small overlapping chunks.
Typically the size of the chunk is 25ms, and a chunk is
created every 10ms.
Each chunk of waveform is represented
as a linear combination of
different sinusoids at different frequencies.
In particular, a frame in this representation
is the vector of coefficients of the linear combination.
This representation is known as the \term{spectrogram}.
Examples of spectrograms are shown in Figure \ref{fig:frame-vs-seg}.

Given a sequence of frames, the goal of ASR
is to predict what is being said in the waveform,
in the form of a sequence of discrete linguistic units,
with words and phonemes being the two most common discrete units.
Phonemes are defined as the sound units that distinguish words \citep{La2005},
and a sequence of phonemes can be converted into words by looking up
the pronunciations of words in a lexicon.
In most settings,
the set of phonemes and the lexicon are assumed to be given.

A standard evaluation metric for ASR is the \term{Levenshtein
distance} between the predicted label sequence and
the ground-truth sequence.  Levenshstein distance
measures the minimum number of edits that needs to be performed
to transform the predicted label sequence
to the ground-truth sequence; hence it is also called the \term{edit
distance}.
Formally, the edit distance of two label sequences $y$ and $\hat{y}$
is defined as
\begin{align}
\text{edit}(\hat{y}, y) = \frac{I + D + S}{|y|},
\end{align}
where $|y|$ is the length of $y$, $I$, $D$, and $S$ are the number of insertion,
deletion, and substitution, respectively.

ASR is a difficult task due to the wide range
of variabilities in the process of speech production.
Even in the most controlled setting, the same isolated word
can hardly be pronounced the same way twice by the same speaker.
Speech production is a highly context-dependent process.
Phonemes are pronounced differently when the neighboring
phonemes are different \citep{La2005}.  Similarly, words are pronounced
differently in different contexts.
Different speakers pronounce words and phonemes differently
due to the differences in their speech organs \citep{LW1995}.
Different speaking styles also affect pronunciations.
For example, in conversational speech, words are
seldom pronounced in the canonical way presented in the lexicon
due to the casual speaking style \citep{Li2005}.
All of the above variabilities make ASR difficult.
It is even more difficult when the speech signals
are degraded by noise \citep{HP2000}.

\subsection{Segment features}

Since the goal of speech recognition is to predict words
based on their pronunciations,
much work has been dedicated to finding features,
sometimes referred to as acoustic cues,
that identify phonemes and differentiate phonemes.
See \citep{H+2005} and citations therein.
Some features are correlated with how humans
identify phonemes \citep{MN1955}.
In general, there are many such features that are useful for
speech recognition.
Each feature alone is probably not enough to identify
a phoneme, but a combination of features might be able to
\citep{H+2005}.
Below we describe two features as examples.

Duration is one of the features that differentiate phonemes.
For example, the duration of the long vowel
/iy/\footnote{The phonemes are written in ARPAbet.}
in the word ``seat'' is typically longer than short vowels
/ih/ in the word ``sit'' \citep{PL1960}.
In this case, duration serves as a good feature
to differentiate /iy/ from /ih/.
Unvoiced fricatives, such as /f/, /s/, and /sh/, typically have
longer duration than voiced fricatives, such /v/, /z/, and /zh/ \citep{U1977}.
Durations have also been shown to help improve recognition results
\citep{ASS1995}.

Perhaps the most salient feature for distinguishing phonemes
by looking at the spectrogram is the distribution of energy \citep{La2005}.
For example, fricatives, such as /s/ and /f/, have evenly spread energy
across frequency.
Stops, such as /t/ and /p/,
have sudden sharp bursts of energy.
Bands of high energy in spectrograms, commonly referred to as formants,
are another type of energy patterns
useful for identifying vowels \citep{La2005}.
The first and second formants (counting from low frequencies)
are commonly used to differentiate vowels \citep{HGCW1995}.
Formants are also useful for speech recognition
\citep{HHG1997}.

Integrating arbitrary segment features into frame-based models
has always been considered a difficult task.
Integrating even just the duration
requires nontrivial modification to the models \citep{L1986}.
For example, we can expand the label set with durations,
but this approach only works for discrete features
and generates unnecessarily large search spaces.

Integrating these segment features is also one of the driving
forces behind the use and development of segmental models \citep{ZGPS1989}.
While hidden Markov models assume that frame labels follow a Markov process,
hidden semi-Markov models assume that segments follow a Markov process.
Semi-Markov processes
have been used to incorporate duration in frame-based models \citep{RM1985}.
Hidden semi-Markov models have been
applied to speech recognition with the same reason in mind \citep{L1986,RC1987}.
The segment-based speech recognizer SUMMIT, is
designed with the goal of integrating arbitrary segment features \citep{ZGPS1989}.

\subsection{Other sequence prediction tasks}

Segmental models have also been applied to other sequence prediction tasks,
such as named entity recognition \citep{SC2005}, American Sign Language fingerspelling recognition \citep{KSL2013},
and action recognition \citep{SNDC2010, HLD2011}.
For named entity recognition, the input is a sequence of words,
and the output is a sequence of named entity tags, such as person name, organization name,
and location name.
For American Sign Language fingerspelling recognition, the input sequence is a sequence of video frames,
and the output is a sequence of letters.
For action unit detection, the input sequence is a sequence of video frames,
and the output is a sequence of actions, such as walk, jump, sit, and stand.
In general, these tasks follow a
left-to-right (or right-to-left) order
in both the input sequence and the output sequence.

A task is said to be linear or monotonic
if the task has a left-to-right (or right-to-left) order.
One sequence prediction task that is not monotonic is translation \citep{C2005}.
Depending on the source and the target languages, the order of words
in the source language can be different from
the order in the target language.
In this thesis, we only focus on segmental models for monotonic sequence
prediction tasks.

\section{Motivations}

We have seen why segmental models are favored over frame-based
models when incorporating complex segment features is needed.
However, the flexibility of segmental models comes with a price.
Enumerating segments of different
lengths and of different labels can be time-consuming.
Consider an input sequence of length 300,
and a label set size of 50.
The number of possible segments of length up to 30
is around $30 \times 50 \times 300 = 450000$.
If we make only a single decision at every time point
as for frame-based models,
we only need to consider $50 \times 300 = 15000$ different
decisions.  The hypothesis space for segmental models is considerably larger,
and as a consequence learning and inference for segmental models
are significantly slower than for frame-based models.

Many researchers have attacked the problem of large hypothesis
spaces either implicitly or explicitly.
The most common approach is to use frame-based models to generate
a set of high-confidence segments, and then use segmental models
to rescore the segments with segment features \citep{G2003, ZN2009}.
\cite{OMTT2006} explicitly train a model for pruning
segments, while \cite{VLL2011} reuse the computation of
features whenever possible.
The first approach needs to have two separate models
and the second approach depends on the exact form of the features.
We would like to design segmental models that do not depend on
other external models, and can handle large hypothesis spaces
in a feature-oblivious manner.

Designing and implementing
state-of-the-art frame-based models requires
a significant amount of engineering effort.
Phonemes are modeled with three-state hidden Markov models \citep{SCRKM1984}.
Since phonemes are influenced heavily by nearby phonemes,
phoneme labels that include the previous and the next phonemes, referred to as triphones,
are introduced \citep{SCRKM1984}.
The large number of triphones makes estimating their
probabilities difficult,
so triphone parameters are shared using with decision trees \citep{YOW1994}.
These design decisions in model structure makes engineering difficult.
In contrast to frame-based models, the engineering effort in segmental models
is put into designing segment features, while
the model structure remains the same.

In sum, we aim to design segmental models that perform well
on sequence prediction tasks, have a clean and modular mathematical definition,
are efficient to train and to decode,
and do not depend on other models.

\section{Previous work}

The concept of segmental models was not
explicitly defined before the late 1980s,
and little work has had segmental models
as the primary focus.
\cite{WMMZ1975} were one of the early
attempts in the 1970s to build speech recognizers
based on segment features.
In the early 1980s, \cite{CSPBPS1983} developed a system
for recognizing isolated English letters based on segment features.
Isolated digits as segments were considered in \citep{KB1985}.
\cite{ZGPS1989} introduced SUMMIT,
a segment-based speech recognizer that can handle
arbitrary segment features.
While the SUMMIT system was later formulated
into a probabilistic framework \citep{GCM1996},
others were not probabilistic,
and have very few parameters to be estimated.

The probabilistic view of segmental models can also be traced back to the 1980s.
As mentioned earlier, \cite{RM1985} introduced semi-Markov processes
and the follow-up \citep{RC1987} introduced hidden semi-Markov models
to the speech recognition community.
These two studies were mainly motivated by the need to include duration
in frame-based models.
\cite{OR1989} formalized segmental models
as a probabilistic framework to include arbitrary features,
but the authors only used frame-based probabilities
with one additional duration feature.
\cite{R1993} and \cite{GY1993} further proposed
segmental hidden Markov models, but the general form
stayed mostly the same.
Segmental models proposed in these studies were generative,
and were mostly restricted to frame-based Gaussian distributions.
This line of work has been summarized in \citep{ODK1996}.

Into the 2000s, studies of segmental models shifted from generative
to discriminative.
\cite{SC2005} proposed semi-Markov conditional random fields (CRF),
the first type of discriminative segmental models.
Semi-Markov CRFs require the use of manual alignments
to train the models, making them less applicable
to speech recognition.
\cite{ZN2009} proposed to use
a different training loss for training semi-Markov CRFs,
marginalizing over all possible segmentations.
This approach alleviated the need for manual alignments to train segmental models.
Later, \cite{Z+2010} used segmental models
as second-pass rescoring models for word recognition,
showcasing the flexibility of segmental models
for incorporating a wide variety of features.

A segmental model
is said to be a first-pass model if it
exhaustively searches over the entire hypothesis space.
Before \citep{Z2012}, all of the segmental models for
speech recognition were not first-pass.  
The early version of SUMMIT \citep{ZGPS1989} used
a heuristic approach to estimate phoneme boundaries,
instead of searching over the entire search space.
A more recent version of SUMMIT \citep{GCM1996} avoided searching
exhaustively by using a frame-based model to generate high confidence
hypotheses.
Semi-Markov CRFs are first-pass segmental models, but have
only been applied to natural language tasks, where
sequences are typically short and label set size is small.

\cite{Z2012} showed that it is feasible
to use segmental models as first-pass speech recognizers.
Since then, studies of segmental models have moved
on to first-pass recognition.
At the same time, advances in neural networks
have allowed us to learn generic feature representations,
so many studies have been devoted to finding better
segment representations to replace hand-crafted features.
\cite{HF2012} improved upon \citep{Z2012} by using
a 3-layer multilayer perceptron to produce phoneme probabilities.
\cite{ADYJ2013} further improved the results by using
a deep convolutional neural network, and it
was also the first to train segmental models and
neural networks end to end.
Along the same line, \cite{LKDSR2016} proposed end-to-end segmental models
with segment representations computed with long short-term memory networks.

\section{Contributions}

In this thesis, we make the following contributions
to the development of segmental models.

\begin{itemize}

\item We introduce discriminative segmental cascades, allowing us to improve sequence
    prediction performance using rich features while maintaining
    efficiency.

\item We develop improved understanding of training approaches and
    training requirements for segmental models.
    We compare segmental models trained end to end and
    ones trained in multiple stages, and
    compare segmental models trained with and without manual alignments

\item Along the way, we present segmental models in a general, modular fashion.
    We present a unifying framework
    that encompasses many end-to-end
    models, such as hidden Markov models, connectionist temporal classification,
    as special cases.

\item We use phonetic recognition and American Sign Language
    fingerspelling recognition as test beds for comparing segmental models
    and frame-based models.  Segmental models are able to outperform
    frame-based models in many settings for both sequence prediction tasks.

\end{itemize}

\section{Thesis outline}

In Chapter 2, we review finite-state transducers (FST),
an important tool for describing the hypothesis spaces
of segmental models.
For the second part of Chapter 2,
we review the essential components in speech recognizers,
such as language models, lexicons, and hidden Markov models,
represented with FSTs.
In Chapter 3, we formally define segmental models.
The definition is modular and covers many existing segmental models
as special cases.
In Chapter 4, we introduce discriminative segmental cascades,
which allow us to explore various segment representations
while maintaining efficiency.
In Chapter 5, we study segmental models trained in different
training conditions, comparing end-to-end training and
multi-stage training.
In Chapter 6, we review several variants of segmental models,
and describe how they relate to our definition of segmental models.
In Chapter 7, we propose a unified framework
encompassing many end-to-end frame-based models
and segmental models.
Finally, in Chapter 8, we discuss possible ways to extend
segmental models.

%% file: background.tex
\chapter{Background}
\label{ch:background}

This chapter describes finite-state transducers (FST),
a useful tool for representing
and manipulating the search space of a sequence
prediction task.
This chapter also includes
a complete example of a standard speech recognizer
based on hidden Markov models
represented as FSTs.
See \citep{M2009}
for a comprehensive review of FSTs and their algorithms,
and \citep{MPR1996} for their applications to speech
recognition.

\section{Finite-State Transducers}

\begin{figure}
\begin{center}
\begin{minipage}{10cm}
\begin{tikzpicture}[ver/.style={circle,draw}]

\node[ver] (x0) at (0, 0) {0};
\node[ver] (x1) at (4, 1.5) {1};
\node[ver] (x2) at (4, -1.5) {2};
\node[ver] (x3) at (8, 0) {3};

\draw[->] (x0) edge[bend left] node[above] {0} (x1);
\draw[->] (x0) edge[bend right] node[above] {1} (x1);
\draw[->] (x0) edge[bend right] node[above] {2} (x2);
\draw[->] (x1) edge[bend left] node[above] {3} (x3);
\draw[->] (x2) edge[bend right] node[above] {4} (x3);

\end{tikzpicture}
\end{minipage}
\begin{minipage}{4cm}
\begin{tabular}{l|ll}
$e$   & tail   & head   \\
\hline
0     & 0      & 1      \\
1     & 0      & 1      \\
2     & 0      & 2      \\
3     & 1      & 3      \\
4     & 2      & 3      
\end{tabular}
\end{minipage}
\caption{A multigraph, where $V = \{0, 1, 2, 3\}$ and $E = \{0, 1, 2, 3, 4\}$.
    The functions of tail and head are shown in the table on the right.}
\label{fig:multigraph}
\end{center}
\end{figure}
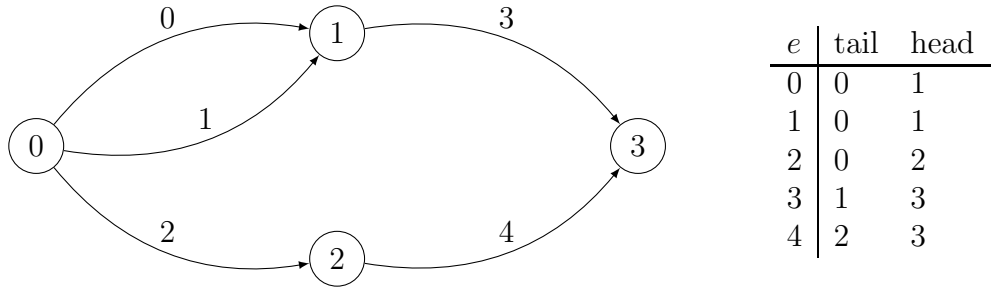

We define FSTs based on multigraphs
(graphs that allow multiple edges between any two vertices)
instead of regular graphs.
A \term{multigraph} $G$ is a tuple $(V, E, \tail, \head)$,
where $V$ is a set of vertices,
$E$ is a set of edges,
$\tail: E \to V$ is a function that
associates an edge to its tail vertex,
and $\head: E \to V$ is a function that
associates an edge to its head vertex.
Note that $E$ is not a subset of $V \times V$
but a general set, because
we allow many edges to have the same tail and head,
and a pair of vertices is not enough to uniquely
identify an edge.
An example of a multigraph is shown in Figure~\ref{fig:multigraph}.
We say that $e_1 \in E$ and $e_2 \in E$ are
connected if $\head(e_1) = \tail(e_2)$.
In addition, we define a \term{path} of length $n$ as a sequence of
connected edges $(e_1, e_2, \dots, e_n)$ where $\head(e_i) = \tail(e_{i+1})$
for $i = 1, \dots, n$.
A sub-path is a sequence of connected edges within a path.
For example, $(e_3, e_4, e_5)$ is a sub-path of $(e_1, e_2, \dots, e_7, e_8)$.

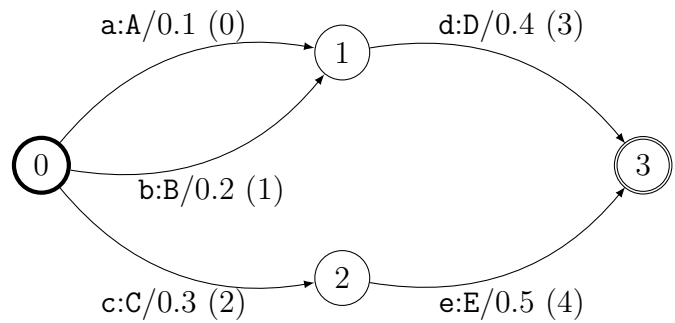
\begin{figure}
\begin{center}
\begin{tikzpicture}[ver/.style={circle,draw}]

\node[ver, ultra thick] (x0) at (0, 0) {0};
\node[ver] (x1) at (4, 1.5) {1};
\node[ver] (x2) at (4, -1.5) {2};
\node[ver, double] (x3) at (8, 0) {3};

\draw[->] (x0) edge[bend left] node[above=2pt] {\texttt{a}:\texttt{A}/0.1 (0)} (x1);
\draw[->] (x0) edge[bend right] node[below=2pt] {\texttt{b}:\texttt{B}/0.2 (1)} (x1);
\draw[->] (x0) edge[bend right] node[below=2pt] {\texttt{c}:\texttt{C}/0.3 (2)} (x2);
\draw[->] (x1) edge[bend left] node[above=2pt] {\texttt{d}:\texttt{D}/0.4 (3)} (x3);
\draw[->] (x2) edge[bend right] node[below=2pt] {\texttt{e}:\texttt{E}/0.5 (4)} (x3);

\end{tikzpicture}
\caption{An FST based on the multigraph in Figure~\ref{fig:multigraph},
    with $\Sigma = \{\texttt{a}, \texttt{b}, \texttt{c}, \texttt{d}, \texttt{e}\}$ and
    $\Lambda = \{\texttt{A}, \texttt{B}, \texttt{C}, \texttt{D}, \texttt{E}\}$.
    The initial vertex is shown in bold, and the final vertex
    is shown with a doubled circle, i.e., $I = \{0\}$ and $F = \{3\}$.
    We use $\sigma$:$\lambda$/w to denote an edge
    with input symbol $\sigma$, output symbol $\lambda$, and weight $w$.
    The number in parentheses is an identifier of an edge.}
\label{fig:fst}
\end{center}
\end{figure}

A \term{finite-state transducer} is a tuple $(G, \Sigma, \Lambda, I, F, i, o, w)$,
where $G$ is a multigraph,
$\Sigma$ is a set of input symbols,
$\Lambda$ is a set of output symbols,
$I \subseteq V$ is a set of initial vertices,
$F \subseteq V$ is a set of final vertices,
$i: E \to \Sigma$ is a function that
associates an edge to its input symbol,
$o: E \to \Lambda$ is a function that
associates an edge to its output symbol,
and $w: E \to \mathbb{R}$ is
a function that puts weights on edges.
An example is shown in Figure~\ref{fig:fst}.

An FST can be seen as a function that maps strings to strings,
where each path in the graph defines an input-output pair.
Specifically, a path $(e_1, e_2, \dots, e_n)$
associates the input string $i(e_1)i(e_2)\cdots i(e_n)$
with the output string $o(e_1)o(e_2)\cdots o(e_n)$.
For example, the FST shown in Figure~\ref{fig:fst}
defines the mapping $\{(\texttt{ad}, \texttt{AD}), (\texttt{bd}, \texttt{BD}),
\allowbreak (\texttt{ce}, \texttt{CE})\}$.

\begin{figure}
\begin{center}
\begin{tikzpicture}[ver/.style={circle,draw}]

\node[ver, ultra thick] (x0) at (0, 0) {0};
\node[ver] (x1) at (4, 1.5) {1};
\node[ver] (x2) at (4, -1.5) {2};
\node[ver, double] (x3) at (8, 0) {3};

\draw[->] (x0) edge[bend left] node[above=2pt] {\texttt{a}:\texttt{A}/0.1 (0)} (x1);
\draw[->] (x0) edge[bend right] node[below=2pt] {\texttt{b}:\texttt{B}/0.2 (1)} (x1);
\draw[->] (x0) edge[bend right] node[below=2pt] {\texttt{c}:\texttt{C}/0.3 (2)} (x2);
\draw[->] (x1) edge[bend left] node[above=2pt] {\texttt{d}:\texttt{D}/0.4 (3)} (x3);
\draw[->, red] (x1) edge[bend right] node[below=2pt] {$\epsilon$:\texttt{F}/0.6 (4)} (x3);
\draw[->] (x2) edge[bend right] node[below=2pt] {\texttt{e}:\texttt{E}/0.5 (5)} (x3);

\end{tikzpicture}
\caption{An FST with empty symbols.  The edge
    with the empty symbol is highlighted in red.}
\label{fig:fst2}
\end{center}
\end{figure}
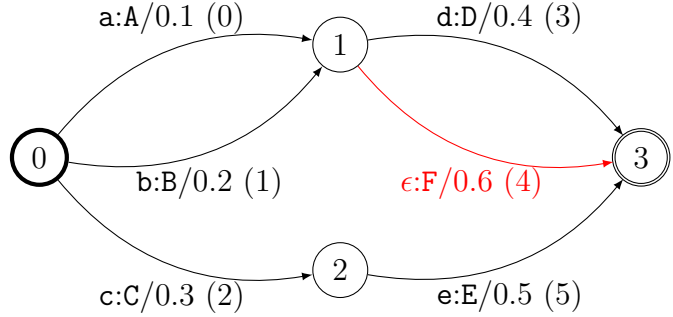

There is a special symbol called the empty symbol, denoted $\epsilon$.
Any string concatenated with an empty symbol is itself,
i.e., $\sigma_1\sigma_2\cdots \sigma_n\epsilon = \epsilon\sigma_1\sigma_2\cdots \sigma_n
= \sigma_1\sigma_2\cdots \sigma_n$.
If the input symbol of an edge is the empty symbol,
then we can traverse the edge without consuming
any input.
If the output symbol of an edge is the empty symbol,
then we do not produce any output when traversing the edge.
As an example, the FST in Figure~\ref{fig:fst2}
defines the mapping $\{(\texttt{ad}, \texttt{AD}), (\texttt{a}, \texttt{AF}),
(\texttt{bd}, \texttt{BD}), (\texttt{b}, \texttt{BF}), (\texttt{ce}, \texttt{CE})\}$.

We assume there is one unique start vertex and
one unique end vertex, i.e., $|I| = |F| = 1$.
If an FST has more than one initial vertices,
we can always create an additional vertex
and connect all initial vertices
to the additional vertex with empty symbols (and possibly zero weights)
without affecting the string mapping of the FST.
For convenience, we also define $\inedges(v) = \{e \in E : \head(e) = v\}$
and $\outedges(v) = \{e \in E: \tail(e) = v\}$.

We have reviewed FSTs as graphs and string functions.  The third view of
FSTs is as state machines.  Take the FST in Figure \ref{fig:fst2}
for example.  To get the output \texttt{AD} from the input \texttt{ad}, we first
feed the character \texttt{a} and traverse from vertex 0 to vertex 1,
and then feed the character \texttt{d} and traverse from vertex 1 to vertex 3.
Similarly, to get \texttt{BF} from \texttt{b}, we first feed the character \texttt{b}
and traverse from vertex 0 to vertex 1.  Due to the empty symbol $\epsilon$,
we get \texttt{F} and traverse from vertex 1 to vertex 3 without
feeding any character.
The view of state machines is especially useful when we talk about FST compositions.

\subsection{Semiring}

Other than just manipulating strings, we are also interested in the weights
on the edges, especially when paths are considered.
We introduce two operators $\otimes$ and $\oplus$,
where $\otimes$ is for combining edge weights and $\oplus$ is for
combining path weights.
For example, by traversing the path $(0, 3)$ that produces \texttt{AD} in Figure \ref{fig:fst2},
we collect the weight 0.1 from edge 0 and 0.4 from edge 3,
and the weight of the path $(0, 3)$ is defined
as $w(0) \otimes w(3) = 0.1 \otimes 0.4$.

Suppose in general $S$ is the set of weights.
It is desirable to have certain properties, such as commutativity and associativity,
for the two operators.  For $a \in S$, $b \in S$, and $c \in S$,
we say that an operator $*$ is \term{commutative} if $a * b = b * a$,
and that an operator $*$ is \term{associative} if $(a * b) * c = a * (b * c)$.
The element $0 \in S$ is an \term{identity} with respect to the operator $*$
if $0 * a = a * 0 = a$.
The element $0 \in S$ annihilates $S$ with respect to
$*$ if $0 * a = a * 0 = 0$.
We say that the operator $*$ distributes over the operator $+$ if
$a * (b + c) = a * b + a * c$ and $(a + b) * c = a * c + b * c$.

The tuple $(S, \oplus, \otimes)$ is a \term{semiring} \citep{M2002} if
$\oplus$ is associative and commutative with identity $0 \in S$,
$\otimes$ is associative with identity $1 \in S$,
$\otimes$ distributes over $\oplus$,
and the element $0 \in S$ annihilates $S$ with respect to $\otimes$.
An example of semiring is $(\mathbb{R}, +, \times)$ with 0
being the identity of $+$, and $1$ being the identity of $\times$.
Another example is the \term{tropical semiring} $(\mathbb{R} \cup \{\infty\}, \min, +)$
with $\infty$ being the identity of $\min$, and $0$ being the identity of $+$.
Yet another example is the \term{log semiring} $(\mathbb{R} \cup \{-\infty\}, \text{logadd}, +)$
where $\text{logadd}(a, b) = \log(\exp(a) + \exp(b))$,
$-\infty$ is the identity of $\text{logadd}$, and $0$ is the identity of $+$.

Defining the operators for combining weights
in a general way encourages algorithm reuse.
As we will see in later chapters, FST algorithms tend to look very similar,
and typically the only difference is how the weights
are combined.
By using general operators, we get a family of algorithms.
In other words, we create algorithms almost for free
by choosing a proper semiring.

\subsection{Shortest-path algorithm}

Given an FST, we are interested in finding a shortest path from the initial vertex
to the final vertex.  For example, if the weights are negative
log probabilities traversing from one vertex to another,
then a shortest path corresponds to one of the most likely paths.
We assume there are multiple shortest paths, but finding one of them is enough.
We also assume the underlying graph is acyclic, meaning
that every path can only traverse a vertex at most once.
For acyclic graphs, we have a nice necessary condition for
shortest paths---every sub-path within a shortest path is also
a shortest path.  The argument for the necessary condition
is simple.  If a sub-path is not a shortest path, then
we can always substitute the sub-path with another shorter
sub-path to create a shorter path overall.

The necessary condition can be viewed as a recursive condition.
Let $\mathcal{P}(u, v)$ be the set of paths from vertex $u$ to vertex $v$,
and let $w(p)$ be a shorthand of $\sum_{e \in p} w(e)$.
Suppose we want to compute a shortest path from the vertex $u$ to vertex $v$.
We examine the set of edges $\inedges(v) = \{e \in E: \head(e) = v\}$ leading into vertex $v$.
If an edge $e \in \inedges(v)$ is part of the shortest path,
then by the necessary condition every path from $u$ to $\tail(e)$ should
also be a shortest path.  In other words,
\begin{align}
\min_{p \in \mathcal{P}(u, v)} w(p)
    & = \min_{e \in \inedges(v)} \min_{p' \in \mathcal{P}(u, \tail(e))}
        [ w(e) + w(p') ] \\
    & = \min_{e \in \inedges(v)}
        \Bigg[ w(e) + \min_{p' \in \mathcal{P}(u, \tail(e))} w(p') \Bigg]
\end{align}
The recursive condition can be computed efficiently with dynamic programming
if we store the shortest distance from vertex $u$ to every other vertex.
Specifically, let $d(v)$ be the shortest distance from $u$ to $v$, i.e.,
\begin{align}
d(v) = \min_{p \in \mathcal{P}(u, v)} w(p).
\end{align}
By the recursive condition, we have
\begin{align}
d(v) = \min_{e \in \inedges(v)} [ w(e) + d(\tail(e)) ]. \label{eq:shortest}
\end{align}
Before computing the minimization in \eqref{eq:shortest},
we need to make sure the shortest sub-paths are computed.
Formally, to compute $d(v)$ for vertex $v$, we need
to make sure $d(\tail(e))$ are computed for $e \in \inedges(v)$.
In other words, if there is a path from vertex $w$ to vertex $v$,
then $d(w)$ should be computed before $d(v)$.
We are looking for an order in which $w$ should come before $v$
if there is a path from $w$ to $v$.
An order in which $w$ comes before $v$
if there is a path from $w$ to $v$ is called a \term{topological order}.
To be precise, a topological order is a function $f: V \to \{1, 2, \dots, |V|\}$
such that $f(w) < f(v)$ if there is a path from $w$ to $v$.
Given an directed acyclic graph, there exist many topological orders.
Any one of them would suffice for our dynamic programming.

One way to find a topological order is to run \term{depth-first search} (DFS)
on the reversed graph.  The depth-first search algorithm is shown
in Algorithm \ref{alg:dfs}.
We maintain a stack $S$.
We say that a vertex is traversed if it has been put on $S$,
and we use a set $T$ to track the traversed vertices.
When a vertex $v$ is put on stack,
we also put a variable indicating whether
we have tried to put its neighbors on the stack,
where the set of neighbors is defined by $\inedges(\cdot)$.
We say that a vertex is expanded if we have tried to
put its neighbors on the stack.
It is not hard to see that every vertex is put on the stack twice, once before
it is expanded and once after.
When a vertex $v$ popped out from the stack is expanded,
all vertices that can be traversed from $v$
are also expanded.
Since we put a vertex $v$ in $O$ when
$v$ and all vertices traversable from $v$ are expanded,
the order $O$ is exactly a topological order.

Algorithm \ref{alg:dfs} is an instance of DFS
specifically for computing
a topological order by defining the set of neighbors
with $\inedges(\cdot)$.
The algorithm can be modified for other purposes,
for example, letting the set of neighbors be $\outedges(\cdot)$
(and changing $\tail(\cdot)$ to $\head(\cdot)$ accordingly)
if we want to search the graph in a forward direction
instead of backwards.
Due to the use of a stack, DFS prefers to follow one
path until none of the vertices can be expanded;
hence the name depth-first search.

\begin{algorithm}
\caption{Depth-First Search (DFS)}
\label{alg:dfs}
\begin{algorithmic}
\Require{Let $r$ be the vertex we start to search.}
\Ensure{A list of vertices $O$ is returned in topological order.}
\State $S = \{(r, \texttt{false})\}$, $T = \{r\}$, $O = ()$
\While{$S$ is not empty}
    \State pop $(v, z)$ from $S$

    \If{$z$}
        \State append $v$ to $O$
    \Else
        \State push $(v, \texttt{true})$ to $S$
        \For{$e \in \inedges(v)$} \Comment{Expand $v$.}
            \If{$\tail(e) \not\in T$}
                \State push $(\tail(e), \texttt{false})$ to $S$
                \Comment{$\tail(e)$ is traversed.}
                \State add $v$ to $T$
            \EndIf
        \EndFor
    \EndIf
\EndWhile
\end{algorithmic}
\end{algorithm}

\begin{algorithm}
\caption{Shortest-Path Algorithm for Directed Acyclic Graphs}
\label{alg:shortest}
\begin{algorithmic}
\Require{Let $t$ be the ending vertex of all paths.}
\Ensure{The map $d$ contains the shortest distances to $v$ for all vertices.}

\State Let $O = (o_1, o_2, \dots, o_n)$ be a topological order by running DFS starting from $t$.

\For{$i = 1, \dots, n$}
    \State $d(o_i) = \min_{e \in \inedges(o_i)} [ w(e) + d(\tail(e)) ]$
    \State $\pi(o_i) = \argmin_{e \in \inedges(o_i)} [ w(e) + d(\tail(e)) ]$
\EndFor
\end{algorithmic}
\end{algorithm}

\begin{algorithm}
\caption{Backtracking}
\label{alg:backtracking}
\begin{algorithmic}
\Require{Let $s$ be the initial vertex and $t$ be the final vertex.}
\Ensure{A shortest path $p$ from $s$ to $t$.}
\State $v \gets t$
\State $p = ()$
\While{$v \neq s$}
    \State add $\pi(v)$ to $p$
    \State $v \gets \tail(\pi(v))$
\EndWhile
\State reverse $p$
\end{algorithmic}
\end{algorithm}

\begin{algorithm}
\caption{Generalized Shortest-Distance Algorithm for Directed Acyclic Graphs}
\label{alg:gen-shortest}
\begin{algorithmic}
\Require{Let $t$ be the ending vertex of all paths.}
\Ensure{The map $d$ contains the shortest distances to $v$ for all vertices.}

\State Let $O = (o_1, o_2, \dots, o_n)$ be a topological order by running DFS starting from $t$.

\For{$i = 1, \dots, n$}
    \State $d(o_i) = \bigoplus_{e \in \inedges(o_i)} [ w(e) \otimes d(\tail(e)) ]$
\EndFor
\end{algorithmic}
\end{algorithm}

Up to this point, we are interested in finding a path that is
shortest.  The weight of a path is defined to be the sum of
the edge weights, and path weights are combined
with the $\min$ operator.
These operators are the ones used in the tropical semiring,
and can be generalized to other semirings,
where edge weights are combined with $\otimes$
and path weights are combined with $\oplus$.
To be precise, the weight of a path $p$ is defined
as $w(p) = \bigotimes_{e \in p} w(e)$, and the distance can be written as
\begin{align}
d(v) & = \bigoplus_{p \in \mathcal{P}(u, v)} w(p)
       = \bigoplus_{p \in \mathcal{P}(u, v)} \bigotimes_{e \in p} w(e)
       = \bigoplus_{e \in \inedges(v)} \bigoplus_{p' \in \mathcal{P}(u, \tail(e))}
           [ w(e) \otimes w(p') ] \\
     & = \bigoplus_{e \in \inedges(v)} \left[ w(e) \otimes \bigoplus_{p' \in \mathcal{P}(u, \tail(e))} w(p') \right] \\
     & = \bigoplus_{e \in \inedges(v)} [w(e) \otimes d(\tail(e))].
\end{align}
The generalized shortest-distance algorithm is shown in
Algorithm~\ref{alg:gen-shortest}.
See \citep{M2002} for a general treatment of shortest-distance algorithms.

The recursive condition \eqref{eq:shortest} only gives us the distance
to the final vertex.
To obtain the path, we can record down the edge that
achieves the minimum while computing the distances.  Specifically, let
$\pi(v)$ be the edge that achieves the minimum, i.e.,
\begin{align}
\pi(v) = \argmin_{e \in \inedges(v)} [ w(e) + d(\tail(e)) ].
\end{align}
We can iteratively collect the optimal edge with $\pi$.
Since the algorithm goes from the final vertex to the initial vertex,
it is commonly called backtracking.
The final shortest-path algorithm and the backtracking algorithm are shown
in Algorithm \ref{alg:shortest} and \ref{alg:backtracking}.

\subsection{Composition}

Recall that FSTs can be regarded as functions that map
strings to strings.
It is natural to have multiple FSTs with one FST consuming the outputs
of another FST, similar to function composition.
Let $T(x, y)$ be the weight of a path in the FST $T$ consuming $x$ as input
and producing $y$ as output.
For any two FSTs $T_1$ and $T_2$, the weight of a path with input $x$ and output $y$
in the composed FST $T_1 \circ T_2$
is defined as
\begin{align} \label{eq:fst-comp}
(T_1 \circ T_2)(x, y) = \bigoplus_{z \in \mathcal{Z}(x)} T_1(x, z) \otimes T_2(z, y),
\end{align}
where $\mathcal{Z}(x)$ is the set of output strings that can be produced
by feeding $x$ to $T_1$ \citep{AM2009}.  The weight is undefined if $\mathcal{Z}(x)$ is
an empty set or there is no path with input $z$ and output $y$ in $T_2$.

\begin{figure}
\centering
\begin{subfigure}[b]{0.5\textwidth}
\centering
\begin{tikzpicture}[ver/.style={circle,draw}]

\node[ver, ultra thick] (x0) at (0, 0) {0};
\node[ver] (x1) at (2, 0) {1};
\node[ver] (x2) at (4, 0) {2};
\node[ver, double] (x3) at (6, 0) {3};

\draw[->] (x0) edge[bend left] node[above] {\texttt{a}:\texttt{A}} (x1);
\draw[->] (x0) edge[bend right] node[below] {\texttt{b}:\texttt{B}} (x1);

\draw[->] (x1) edge[bend left] node[above] {\texttt{a}:\texttt{A}} (x2);
\draw[->] (x1) edge[bend right] node[below] {\texttt{b}:\texttt{B}} (x2);

\draw[->] (x2) edge[bend left] node[above] {\texttt{a}:\texttt{A}} (x3);
\draw[->] (x2) edge[bend right] node[below] {\texttt{b}:\texttt{B}} (x3);

\end{tikzpicture}
\vspace{0.5cm}
\caption{FST $T_1$}
\end{subfigure}
\begin{subfigure}[b]{0.45\textwidth}
\centering
\begin{tikzpicture}[ver/.style={circle,draw}]

\node[ver, ultra thick] (x0) at (0, 0) {0};
\node[ver] (x1) at (2, 0) {1};
\node[ver, double] (x2) at (4, 0) {2};

\draw[->] (x0) edge node[above] {\texttt{A}:$\alpha$} (x1);
\draw[->] (x1) edge[loop] node[above] {\texttt{A}:$\alpha$} (x1);
\draw[->] (x1) edge node[above] {\texttt{B}:$\beta$} (x2);
\draw[->] (x2) edge[loop] node[above] {\texttt{B}:$\beta$} (x2);

\end{tikzpicture}
\vspace{0.5cm}
\caption{FST $T_2$}
\end{subfigure}

\vspace{2cm}
\begin{subfigure}[b]{0.9\textwidth}
\centering
\begin{tikzpicture}[ver/.style={circle,draw}]
    
\node[ver, ultra thick] (x00) at (0, 0) {};
\node[ver] (x01) at (3, 0) {};
\node[ver] (x02) at (6, 0) {};
\node[ver] (x03) at (9, 0) {};
\node[ver] (x10) at (0, -2) {};
\node[ver] (x11) at (3, -2) {};
\node[ver] (x12) at (6, -2) {};
\node[ver] (x13) at (9, -2) {};
\node[ver] (x20) at (0, -4) {};
\node[ver] (x21) at (3, -4) {};
\node[ver] (x22) at (6, -4) {};
\node[ver, double] (x23) at (9, -4) {};

\node (y00) at (0, -0.5) {(0, 0)};
\node (y01) at (3, -0.5) {(0, 1)};
\node (y02) at (6, -0.5) {(0, 2)};
\node (y03) at (9, -0.5) {(0, 3)};
\node (y10) at (0, -2.5) {(1, 0)};
\node (y11) at (3, -2.5) {(1, 1)};
\node (y12) at (6, -2.5) {(1, 2)};
\node (y13) at (9, -2.5) {(1, 3)};
\node (y20) at (0, -4.5) {(2, 0)};
\node (y21) at (3, -4.5) {(2, 1)};
\node (y22) at (6, -4.5) {(2, 2)};
\node (y23) at (9, -4.5) {(2, 3)};

\draw[->] (x00) edge node[right] {\texttt{a}:$\alpha$} (x11);
\draw[->] (x11) edge node[above] {\texttt{a}:$\alpha$} (x12);
\draw[->] (x12) edge node[right] {\texttt{b}:$\beta$} (x23);
\draw[->] (x11) edge node[left] {\texttt{b}:$\beta$} (x22);
\draw[->] (x22) edge node[above] {\texttt{b}:$\beta$} (x23);

\end{tikzpicture}
\vspace{0.5cm}
\caption{FST $T_1 \circ_\sigma T_2$}
\end{subfigure}
\vspace{0.5cm}
\caption{An example of structured composition.
    Vertices in $T_1 \circ_\sigma T_2$ are labeled with pairs of vertices from $T_1$ and $T_2$.}
\label{fig:fst-sigma-comp}
\end{figure}
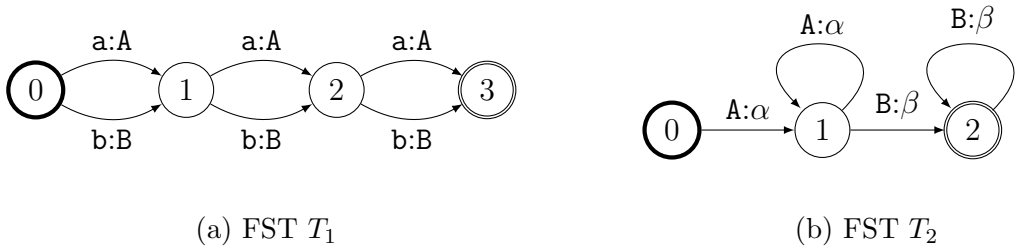
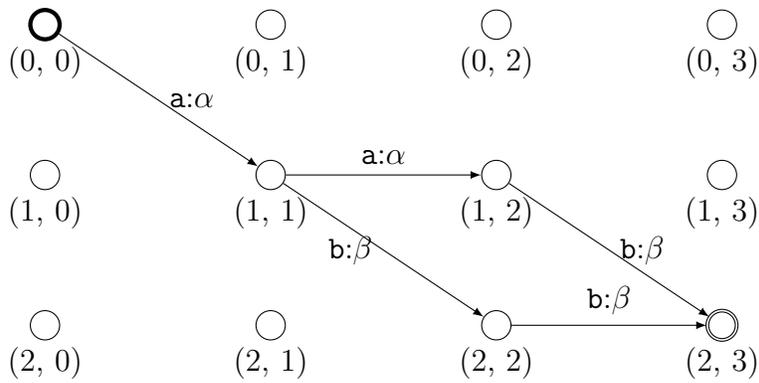

The definition of \eqref{eq:fst-comp} does not provide an explicit
structure of the composed FST.
In this thesis, we prefer another approach to composing FSTs which
we term \term{structured composition} \citep{TWGL2015}.  Formally, the structured
composition (or $\sigma$-composition) of two FSTs
$T_1 = (G_1, \Sigma_1, \Lambda_1, I_1, F_1, i_1, o_1, w_1)$
with $G_1 = (V_1, E_1, \tail_1, \head_1)$
and
$T_2 = (G_2, \Sigma_2, \Lambda_2, I_2, F_2, i_2, o_2,\allowbreak w_2)$
with $G_2 = (V_2, E_2, \tail_2, \head_2)$
is defined as 
$T_1 \circ_\sigma T_2 = (G, \Sigma, \Lambda, I,\allowbreak F, i, o, w)$
where $G = (V, E, \tail, \head)$ with the following
constraints.
\begin{align} \label{eq:graph-comp}
V & = V_1 \times V_2 &
E & = \Big\{\langle e_1, e_2 \rangle \in E_1 \times E_2 : o_1(e_1) = i_2(e_2) \Big\}
\end{align}
\begin{align}
\Sigma & = \Sigma_1 &
    i(\langle e_1, e_2 \rangle) & = i_1(e_1) \\
\Lambda & = \Lambda_2 &
    o(\langle e_1, e_2 \rangle) & = o_2(e_2) \\
I & = I_1 \times I_2 &
    \tail(\langle e_1, e_2 \rangle) & = \langle \tail_1(e_1), \tail_2(e_2) \rangle \\
F & = F_1 \times F_2 &
    \head(\langle e_1, e_2 \rangle) & = \langle \head_1(e_1), \head_2(e_2) \rangle
\end{align}
The $\sigma$-composed graph is defined over pairs of vertices and pairs of edges.
Due to the edge constraints \eqref{eq:graph-comp}, the outputs from $T_1$
have to match the inputs to $T_2$.
An example is shown in Figure \ref{fig:fst-sigma-comp}.
Note that the definition of the weight
function is left to the users.  One possible definition is to let
\begin{align}
w(\langle e_1, e_2 \rangle) = w_1(e_1) \otimes w_2(e_2).
\end{align}

As a consequence of \eqref{eq:graph-comp}, the outgoing edges
and incoming edges of a vertex in the $\sigma$-composed FST can be computed locally
with $\outedges_1(\cdot)$ and $\outedges_2(\cdot)$.  Specifically,
\begin{align} \label{eq:outedges}
\outedges(\langle v_1, v_2 \rangle)
    = \Big\{ \langle e_1, e_2 \rangle: e_1 \in \outedges_1(v_1), e_2 \in \outedges_2(v_2),
        o_1(e_1) = i_2(e_2) \Big\}.
\end{align}
Since computing $\outedges(\langle v_1, v_2 \rangle)$ only requires
access to edges connected to $v_1$ and $v_2$,
traversing the composed FST, for example with DFS (Algorithm \ref{alg:dfs}),
only needs to keep track of the vertex pairs.
The edges can be computed on the fly and do not need to be stored in memory.
In general, composing FSTs without explicitly storing the entire graph
is commonly known as on-the-fly composition or lazy composition.

Another way to compute structured composition is to view FSTs as state machines.
Take Figure \ref{fig:fst-sigma-comp} for example.
We start from vertex 0 in both $T_1$ and $T_2$.
To get to vertex 1 in $T_1$, we can either produce \texttt{A}
or produce \texttt{B} as output.  In $T_2$, to go from vertex 0 to vertex 1,
we only have the choice to consume \texttt{A}.
Therefore, in the composed FST $T_1 \circ_\sigma T_2$ we have the edge
from $(0, 0)$ to $(1, 1)$ with input \texttt{a} and output $\alpha$.
By walking through the vertices one by one, we are essentially
performing a search on both FSTs while computing \eqref{eq:outedges}
on the fly.

\section{FST-Based Speech Recognizers}
\label{sec:fst-rec}

We now have all the necessary tools to describe
a basic speech recognizer based on FSTs.
Modern speech recognizers are
based on a series of string function compositions.
Following \citep{MPR2002}, we represent the string functions as FSTs,
and the compositions are realized with structured composition.
The first FST $U$ takes acoustic frames as inputs
and produces a sequence of frame labels.
The second FST $H$ takes in frame labels and
converts them into phonemes.
The third FST $L$ takes phonemes as inputs
and produces words as outputs.
The fourth FST $G$ takes words as inputs
and produces words as outputs.
Finally, the four FSTs are $\sigma$-composed into
\begin{align}
D = U \circ_\sigma H \circ_\sigma L \circ_\sigma G.
\end{align}
The path weights in $D$ are typically negative log probabilities,
so the maximum-weight path
is the most likely path based on the probabilities.
To find the maximum-weight path,
we simply negate the weights and find the shortest path.
The tropical semiring is used to combine the negated weights.
In sum, to predict a sequence from a sequence of frames,
we create the FST $D$ and find the maximum-weight path in $D$.
Below we describe the four FSTs $U$, $H$, $L$ and $G$ in detail.

\subsection{Hidden Markov models}

\begin{figure}
\begin{center}
\begin{tikzpicture}[ver/.style={circle,draw}]

\node[ver] (y0) at (0, 0) {};
\node[ver] (y1) at (1, 0) {};
\node[ver] (y2) at (2, 0) {};
\node[ver] (y3) at (3, 0) {};

\node at (0, 0.5) {$y_1$};
\node at (1, 0.5) {$y_2$};
\node at (2, 0.5) {$y_3$};
\node at (3, 0.5) {$y_4$};

\node[ver, fill=black!50] (x0) at (0, -1) {};
\node[ver, fill=black!50] (x1) at (1, -1) {};
\node[ver, fill=black!50] (x2) at (2, -1) {};
\node[ver, fill=black!50] (x3) at (3, -1) {};

\node at (0, -1.5) {$x_1$};
\node at (1, -1.5) {$x_2$};
\node at (2, -1.5) {$x_3$};
\node at (3, -1.5) {$x_4$};

\draw[->] (y0) -- (x0);
\draw[->] (y1) -- (x1);
\draw[->] (y2) -- (x2);
\draw[->] (y3) -- (x3);

\draw[->] (y0) -- (y1);
\draw[->] (y1) -- (y2);
\draw[->] (y2) -- (y3);

\end{tikzpicture}
\caption{A hidden Markov model for $4$ frames $x_1, x_2, x_3, x_4$
    with hidden labels $y_1, y_2, y_3, y_4$.
    Note that this is a graphical model where vertices are random variables
    and edges represent conditional dependencies, not an FST.
    The observed variables are shaded, and the unobserved are not.}
\label{fig:hmm}
\end{center}
\end{figure}
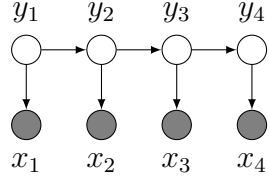

The dynamics of speech are commonly modeled by hidden Markov models (HMM).
The generative story of a hidden Markov model is as follows.
We first generate a state $y_1$ based on a prior distribution
and generate a frame $x_1$ given $y_1$.
A state $y_2$ is generated given $y_1$,
and a frame $x_2$ is generated given $y_2$.
In general, for some $t = 2, \dots, T$, the state $y_t$
is generated given $y_{t-1}$,
and $x_t$ is generated given $y_t$.
For a sequence of $T$ frames $x_1, \dots, x_T$, the probability
distribution defined by an HMM is
\begin{align}
p(x_{1:T}, y_{1:T}) = p(y_1) \prod_{t=2}^T p(y_t | y_{t-1}) \prod_{t=1}^T p(x_t | y_t),
\end{align}
where $x_{1:T}$ is a shorthand for $x_1, \dots, x_T$,
$x_t \in \mathbb{R}^d$ for some $d \in \mathbb{N}$
and $y_t \in \{1, \dots, S\}$ for some $S \in \mathbb{N}$
for $t=1, \dots, T$.
An example of a four-frame HMM is shown in Figure \ref{fig:hmm}.
Note that Figure \ref{fig:hmm} is not an FST but a graphical model
where vertices are random variables and edges represent conditional dependencies.
The learnable parameters in an HMM are $p(y_1)$, $p(y_t | y_{t-1})$, and $p(x_t | y_t)$
for $t = 1, \dots, T$.
The probabilities $p(y_t | y_{t-1})$ are commonly known as transition probabilities,
and $p(x_t | y_t)$ emission probabilities.

\begin{figure}
\begin{center}
\begin{tikzpicture}[ver/.style={circle,draw}]

\node[ver] (x0) at (0, 0) {1};
\node[ver] (x1) at (4, 0) {2};
\node[ver] (x2) at (8, 0) {3};

\draw[->] (x0) edge node[above] {\texttt{ih2}:$\epsilon$/$\log a_{12}$} (x1);
\draw[->] (x1) edge node[above] {\texttt{ih3}:$\epsilon$/$\log a_{23}$} (x2);

\draw[->] (x0) edge[loop] node[above] {\texttt{ih1}:$\epsilon$/$\log a_{11}$} (x0);
\draw[->] (x1) edge[loop] node[above] {\texttt{ih2}:$\epsilon$/$\log a_{22}$} (x1);
\draw[->] (x2) edge[loop] node[above] {\texttt{ih3}:$\epsilon$/$\log a_{33}$} (x2);

\end{tikzpicture}
\caption{A 3-state FST for the phoneme \texttt{ih}, where $\log a_{ij}$ is
    the log transition probability from state $i$ to state $j$.}
\label{fig:phone-hmm}
\end{center}
\end{figure}
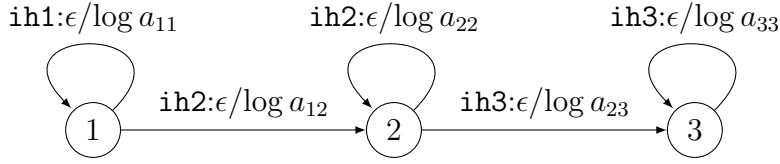

The transition probabilities $p(y_t | y_{t-1})$ can be represented
as a square matrix $A$ of size $S \times S$ in which $A_{ij} \in [0, 1]$ is
the probability of transitioning from state $i$ to state $j$
satisfying $\sum_{j=1}^S A_{ij} = 1$.
To forbid the transition from state $i$ to state $j$, we simply let $A_{ij} = 0$.
It is common to model a phoneme as a 3-state HMM,
and parameterize $A$ as
\begin{align}
\begin{pmatrix}
a_{11} & a_{12} & 0 \\
0 & a_{22} & a_{23} \\
0 & 0 & a_{33}
\end{pmatrix}
\end{align}
where we are only allowed to either stay in the current state or move
to the next state.
The allowed transitions can be conveniently represented as
an FST, where vertices are states and edges are transitions
associated with transition probabilities.
An example of the allowed state transitions is shown in Figure \ref{fig:phone-hmm}.

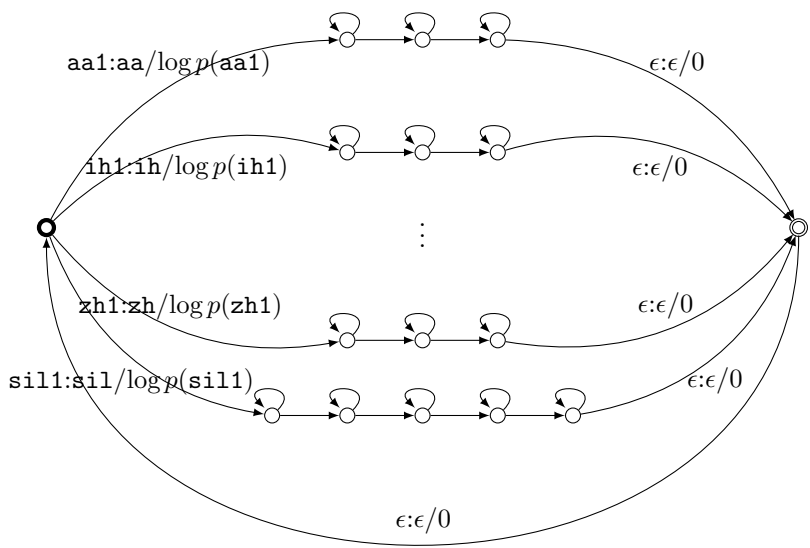
\begin{figure}
\begin{center}
\begin{tikzpicture}[ver/.style={circle,draw,inner sep=2pt}, font=\footnotesize]

\node[ver] (x1) at (1, 0) {};
\node[ver] (x2) at (2, 0) {};
\node[ver] (x3) at (3, 0) {};

\draw[->] (x1) edge (x2);
\draw[->] (x2) edge (x3);

\draw[->] (x1) edge[loop] (x1);
\draw[->] (x2) edge[loop] (x2);
\draw[->] (x3) edge[loop] (x3);

\node[ver] (y1) at (1, -1.5) {};
\node[ver] (y2) at (2, -1.5) {};
\node[ver] (y3) at (3, -1.5) {};

\draw[->] (y1) edge (y2);
\draw[->] (y2) edge (y3);

\draw[->] (y1) edge[loop] (y1);
\draw[->] (y2) edge[loop] (y2);
\draw[->] (y3) edge[loop] (y3);

\node at (2, -2.5) {$\vdots$};

\node[ver] (z1) at (1, -4) {};
\node[ver] (z2) at (2, -4) {};
\node[ver] (z3) at (3, -4) {};

\draw[->] (z1) edge (z2);
\draw[->] (z2) edge (z3);

\draw[->] (z1) edge[loop] (z1);
\draw[->] (z2) edge[loop] (z2);
\draw[->] (z3) edge[loop] (z3);

\node[ver] (w0) at (0, -5) {};
\node[ver] (w1) at (1, -5) {};
\node[ver] (w2) at (2, -5) {};
\node[ver] (w3) at (3, -5) {};
\node[ver] (w4) at (4, -5) {};

\draw[->] (w0) edge (w1);
\draw[->] (w1) edge (w2);
\draw[->] (w2) edge (w3);
\draw[->] (w3) edge (w4);

\draw[->] (w0) edge[loop] (w0);
\draw[->] (w1) edge[loop] (w1);
\draw[->] (w2) edge[loop] (w2);
\draw[->] (w3) edge[loop] (w3);
\draw[->] (w4) edge[loop] (w4);

\node[ver, ultra thick] (s) at (-3, -2.5) {};

\draw[->] (s) edge[bend left] node[above] {\texttt{aa1}:\texttt{aa}/$\log p(\texttt{aa1})$} (x1);
\draw[->] (s) edge[bend left] node[below] {\texttt{ih1}:\texttt{ih}/$\log p(\texttt{ih1})$} (y1);
\draw[->] (s) edge[bend right] node[above] {\texttt{zh1}:\texttt{zh}/$\log p(\texttt{zh1})$} (z1);
\draw[->] (s) edge[bend right] node[below] {\texttt{sil1}:\texttt{sil}/$\log p(\texttt{sil1})$} (w0);

\node[ver, double] (t) at (7, -2.5) {};

\draw[->] (x3) edge[bend left] node[above] {$\epsilon$:$\epsilon$/0} (t);
\draw[->] (y3) edge[bend left] node[below] {$\epsilon$:$\epsilon$/0} (t);
\draw[->] (z3) edge[bend right] node[above] {$\epsilon$:$\epsilon$/0} (t);
\draw[->] (w4) edge[bend right] node[below] {$\epsilon$:$\epsilon$/0} (t);

\draw[->] (t) edge[bend left=90, looseness=1.4] node[above] {$\epsilon$:$\epsilon$/0} (s) {};

\end{tikzpicture}
\caption{An FST $H$ that converts phoneme states to phonemes.}
\label{fig:h}
\end{center}
\end{figure}

To allow transitions from phonemes to phonemes, we connect the FSTs,
such as the ones in Figure \ref{fig:phone-hmm}, in parallel.
An example is shown in Figure \ref{fig:h}.
An additional start state and an additional end state are added.
A backward edge from the end state to the start state is also added
to allow a sequence of phonemes with indefinite length.
The prior distribution $p(y_1)$ entering the phoneme
is also added from the start state to each of the phoneme FSTs.
Note that when entering one of the phoneme FSTs, the edge
consumes a phoneme state and produces a phoneme,
and the phoneme FSTs only consume phoneme states
and do not produce any output.
Silences are modeled with five states rather than three,
because they are allowed to be longer than other phonemes.
This finishes the construction of the FST $H$
that converts phoneme states to phonemes.

\begin{figure}
\begin{center}
\begin{tikzpicture}[ver/.style={circle,draw}]

\node[ver] (x0) at (0, 0) {};
\node[ver] (x1) at (2, 0) {};
\node[ver] (x2) at (4, 0) {};
\node[ver] (x3) at (6, 0) {};
\node[ver] (x4) at (8, 0) {};

\draw[->] (x0) edge[bend left=60] (x1);
\draw[->] (x0) edge[bend left=45] (x1);
\draw[->] (x0) edge[bend left=30] (x1);
\draw[->] (x0) edge[bend left=15] (x1);
\draw[->] (x0) edge[bend right=60] (x1);

\draw[->] (x1) edge[bend left=60] (x2);
\draw[->] (x1) edge[bend left=45] (x2);
\draw[->] (x1) edge[bend left=30] (x2);
\draw[->] (x1) edge[bend left=15] (x2);
\draw[->] (x1) edge[bend right=60] (x2);

\draw[->] (x2) edge[bend left=60] (x3);
\draw[->] (x2) edge[bend left=45] (x3);
\draw[->] (x2) edge[bend left=30] (x3);
\draw[->] (x2) edge[bend left=15] (x3);
\draw[->] (x2) edge[bend right=60] (x3);

\draw[->] (x3) edge[bend left=60] (x4);
\draw[->] (x3) edge[bend left=45] (x4);
\draw[->] (x3) edge[bend left=30] (x4);
\draw[->] (x3) edge[bend left=15] (x4);
\draw[->] (x3) edge[bend right=60] (x4);

\node (y0) at (1, -1.5) {$\begin{bmatrix} \vdots \end{bmatrix}$};
\node (y1) at (3, -1.5) {$\begin{bmatrix} \vdots \end{bmatrix}$};
\node (y2) at (5, -1.5) {$\begin{bmatrix} \vdots \end{bmatrix}$};
\node (y3) at (7, -1.5) {$\begin{bmatrix} \vdots \end{bmatrix}$};

\node at (1, -0.2) {$\vdots$};
\node at (3, -0.2) {$\vdots$};
\node at (5, -0.2) {$\vdots$};
\node at (7, -0.2) {$\vdots$};

\node (z0) at (1, -2.25) {$x_1$};
\node (z1) at (3, -2.25) {$x_2$};
\node (z2) at (5, -2.25) {$x_3$};
\node (z3) at (7, -2.25) {$x_4$};

\node (t0) at (3, 2.5) {\begin{tabular}{c} $s$:$s$/$\log \pi_{i, s} + \log p(x_1 | \mu_{i, s}, \sigma^2_{i, s})$ \end{tabular}};

\draw[->] (t0) edge[bend right] (1, 0.7);

\end{tikzpicture}
\caption{An FST $U$ for 4 frames with GMM probabilities.
    For example, if each phoneme has 3 states, and each
    state has 64 Gaussian components, then
    $s \in \{\texttt{aa1}, \texttt{aa2}, \texttt{aa3}, \dots, \allowbreak \texttt{zh1}, \texttt{zh2}, \texttt{zh3} \}$,
    $i \in \{1, \dots, C = 64\}$.
    If there are $L$ phonemes (i.e., $S = 3L$), then there are $3L \times 64$ (i.e., $S \times C)$ edges between each pair of vertices.}
\label{fig:gmm}
\end{center}
\end{figure}
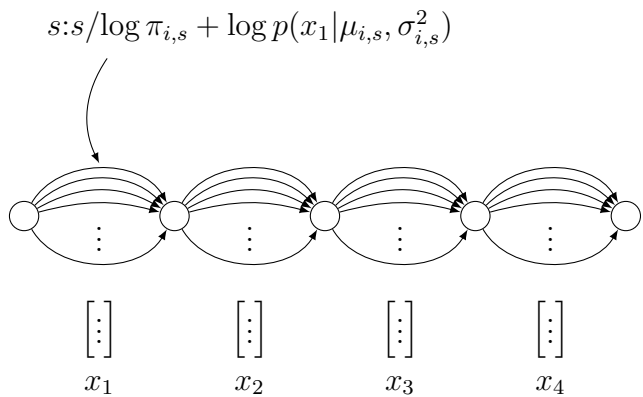

The emission probabilities $p(x_t | y_t)$ in HMM are typically modeled
by Gaussian mixture models (GMM).  Specifically,
\begin{align}
p(x_t | y_t) = \sum_{i=1}^C \pi_{i, y_t} p( x_t | \mu_{i, y_t}, \sigma^2_{i, y_t})
\end{align}
where there are $C$ components for each state in $\{1, \dots, S\}$,
each component $p(x_t | \mu_{i, y}, \sigma^2_{i, y})$ is a Gaussian distribution with mean $\mu_{i, y}$
and variance $\sigma^2_{i, y}$,
and $\pi_{i, y}$ is the probability of selecting the $i$-th component.
For a sequence of $T$ frames, the emission probability for each frame
can be independently evaluated.
Since each state has $C$ Gaussian components, there are $S \times C$
evaluations for each frame.
We can build an FST that has $S \times C$ edges for every frame.
Each edge has a weight computed from its Gaussian component
and has a phone state as the output symbol.
An example is shown in Figure \ref{fig:gmm}.
Note that the FST is constructed by following the generative story
of Gaussian mixture models, evaluating a single Gaussian component
(selected based on $\pi$) for every frame
rather than evaluating the weighted average of all Gaussian components.

By $\sigma$-composing $U$ and $H$, we construct an FST where each path in the FST
is a sample drawn from the generative story and the weight of each path
is the corresponding log probability.
To better see the connection, we can traverse the FST $U$ and $H$ synchronously
as the way we compute $U \circ_\sigma H$.
First when we traverse an edge in $U$, we select one phoneme state and one Gaussian component,
collecting the log probability and producing a phoneme state as output.
Then the FST $H$ receives the phoneme state
and waits for the next phoneme state,
collecting transition probabilities and producing a phoneme when necessary.
This completes the construction of $U$ and $H$.

\subsection{Pronunciation dictionary}

\begin{figure}
\begin{center}
\begin{tikzpicture}[ver/.style={circle,draw}]

\node[ver, ultra thick] (s) at (0, 0) {};
\node[ver, double] (t) at (12.5, 0) {};

\node[ver] (x0) at (2.5, 1) {};
\node[ver] (x1) at (5, 1) {};
\node[ver] (x2) at (7.5, 1) {};
\node[ver] (x3) at (10, 1) {};

\draw[->] (s) edge node[above] {$\epsilon$:$\epsilon$} (x0);
\draw[->] (x0) edge node[above] {\texttt{iy}:\texttt{either}} (x1);
\draw[->] (x1) edge node[above] {\texttt{dh}:$\epsilon$} (x2);
\draw[->] (x2) edge node[above] {\texttt{er}:$\epsilon$} (x3);
\draw[->] (x3) edge node[above] {$\epsilon$:$\epsilon$} (t);

\node[ver] (y0) at (2.5, 0) {};
\node[ver] (y1) at (5, 0) {};
\node[ver] (y2) at (7.5, 0) {};
\node[ver] (y3) at (10, 0) {};

\draw[->] (s) edge node[above] {$\epsilon$:$\epsilon$} (y0);
\draw[->] (y0) edge node[above] {\texttt{ay}:\texttt{either}} (y1);
\draw[->] (y1) edge node[above] {\texttt{dh}:$\epsilon$} (y2);
\draw[->] (y2) edge node[above] {\texttt{er}:$\epsilon$} (y3);
\draw[->] (y3) edge node[above] {$\epsilon$:$\epsilon$} (t);

\node[ver] (z0) at (2.5, 2) {};
\node[ver] (z1) at (5, 2) {};
\node[ver] (z2) at (7.5, 2) {};
\node[ver] (z3) at (10, 2) {};

\draw[->] (s) edge node[above] {$\epsilon$:$\epsilon$} (z0);
\draw[->] (z0) edge node[above] {\texttt{k}:\texttt{cat}} (z1);
\draw[->] (z1) edge node[above] {\texttt{ae}:$\epsilon$} (z2);
\draw[->] (z2) edge node[above] {\texttt{t}:$\epsilon$} (z3);
\draw[->] (z3) edge node[above] {$\epsilon$:$\epsilon$} (t);

\node[ver] (w0) at (2.5, -2) {};
\node[ver] (w1) at (6.25, -2) {};
\node[ver] (w2) at (10, -2) {};

\draw[->] (s) edge node[above] {$\epsilon$:$\epsilon$} (w0);
\draw[->] (w0) edge node[above] {\texttt{z}:\texttt{zoo}} (w1);
\draw[->] (w1) edge node[above] {\texttt{uw}:$\epsilon$} (w2);
\draw[->] (w2) edge node[above] {$\epsilon$:$\epsilon$} (t);

\node[ver] (v0) at (2.5, -3) {};
\node[ver] (v1) at (10, -3) {};

\draw[->] (s) edge node[above] {$\epsilon$:$\epsilon$} (v0);
\draw[->] (v0) edge node[above] {\texttt{sil}:\texttt{sil}} (v1);
\draw[->] (v1) edge node[above] {$\epsilon$:$\epsilon$} (t);

\node at (6.25, -0.75) {$\vdots$};

\draw[->] (t) edge[bend left=90,looseness=1.2] node[above] {$\epsilon$:$\epsilon$} (s);

\end{tikzpicture}
\caption{An FST $L$ representing a pronunciation dictionary.}
\label{fig:l}
\end{center}
\end{figure}
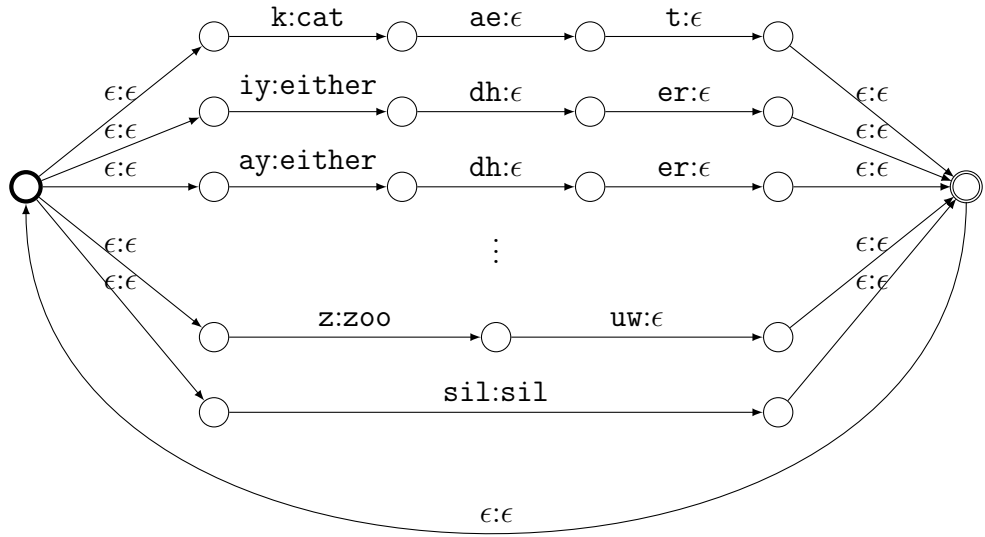

The pronunciation dictionary, or lexicon, is a mapping from a word to its pronunciation.
To construct an FST representing the lexicon, we create, for each entry in the lexicon,
a path consuming a sequence of phonemes and producing a word.
Similar to the FST $H$, we have a start vertex and an end vertex that joins
the pronunciation paths, and we also have a backward edge that allows indefinite
amount of words to be produced.
An example is shown in Figure \ref{fig:l}.  Note that the weights on the edges can
either be 0 or the probability of a chosen pronunciation.
The FST also allows a word to have multiple pronunciations.
Silences are considered as words and are included in the FST.
The size of the FST can be significantly reduced if we maintain
a prefix tree of the phonemes instead of having parallel pronunciations
for every word.

\subsection{Language models}

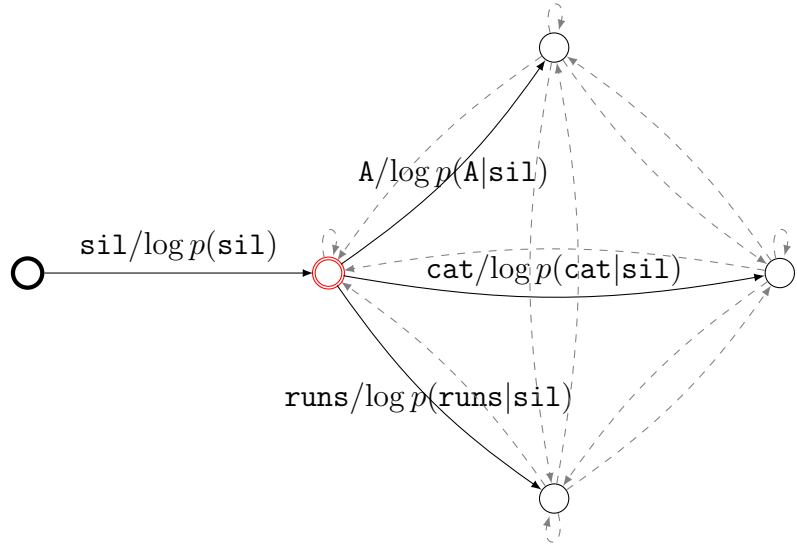
\begin{figure}
\begin{center}
\begin{tikzpicture}[ver/.style={circle,draw}]

\node[ver, ultra thick] (x0) at (-4, 0) {};

\node[ver, double, red] (x1) at (0, 0) {};
\node[ver] (x2) at (6, 0) {};
\node[ver] (x3) at (3, 3) {};
\node[ver] (x4) at (3, -3) {};

\draw[->] (x0) edge node[above] {\texttt{sil}/$\log p(\texttt{sil})$} (x1);
\draw[->] (x1) edge[bend right=10] node[above] {\texttt{cat}/$\log p(\texttt{cat}|\texttt{sil})$} (x2);
\draw[->] (x1) edge[bend right=10] node {\texttt{A}/$\log p(\texttt{A}|\texttt{sil})$} (x3);
\draw[->] (x1) edge[bend right=10] node {\texttt{runs}/$\log p(\texttt{runs}|\texttt{sil})$} (x4);

\draw[->] (x1) edge[loop above, black!50, dashed] (x1);
\draw[->] (x2) edge[loop above, black!50, dashed] (x2);
\draw[->] (x3) edge[loop above, black!50, dashed] (x3);
\draw[->] (x4) edge[loop below, black!50, dashed] (x4);

\draw[->] (x2) edge[bend right=10, black!50, dashed] (x1);
\draw[->] (x2) edge[bend right=10, black!50, dashed] (x3);
\draw[->] (x2) edge[bend right=10, black!50, dashed] (x4);

\draw[->] (x3) edge[bend right=10, black!50, dashed] (x1);
\draw[->] (x3) edge[bend right=10, black!50, dashed] (x2);
\draw[->] (x3) edge[bend right=10, black!50, dashed] (x4);

\draw[->] (x4) edge[bend right=10, black!50, dashed] (x1);
\draw[->] (x4) edge[bend right=10, black!50, dashed] (x2);
\draw[->] (x4) edge[bend right=10, black!50, dashed] (x3);

\end{tikzpicture}
\caption{An FST $G$ representing a bigram language model
    for the vocabulary \{\texttt{sil}, \texttt{A}, \texttt{cat}, \texttt{runs}\}.
    Some edges are dashed to avoid clutter.
    The output symbols are the same as the input symbols for every edge,
    hence ignored. The vertex in red is the silence state.}
\label{fig:g}
\end{center}
\end{figure}

Language models assign probabilities to word sequences.
For any $k \in \mathbb{N}$,
a word sequence $w_{1:k} = w_1, \dots, w_k$ of length $k$
has probability
\begin{align}
p(w_{1:k}) = \prod_{i=1}^k p(w_i | w_{1:i-1}).
\end{align}
As the sequence gets longer, i.e., as $i$ gets larger,
$p(w_i | w_{1:i-1})$ depends on more words,
which makes estimating the probabilities difficult.
A simple approach to remedy this problem is to introduce
the Markov assumption
\begin{align}
p(w_i | w_{1:i-1}) = p(w_i | w_{i-n+1:i-1}).
\end{align}
A language model that gives a probability for the current word
based on the previous $n-1$ words is
commonly known as an $n$-gram language model.
To construct an FST representing a language model,
we create vertices that corresponds to history
words $w_{i-n+1:i-1}$.
For each vertex, the outgoing edges produce the next
word $w_i$ with the weight $\log p(w_i | w_{i-n+1:i-1})$.
A path in this FST consumes a sequence of words
and has the probability of the word sequence
assigned by the language model.
There is a start vertex representing the state
of having no history words.
We assume the utterance starts and ends with silences,
so there is only one edge going out from the start state
expecting a silence.
An example is shown in Figure \ref{fig:g}.
Back-off language models can also be approximated with
FSTs \citep{AMR2003}.

\section{Summary}

In this chapter, we have reviewed the definition of finite-state transducers (FST)
and constructed a basic speech recognizer by $\sigma$-composing
an HMM emission FST $U$, an HMM transition FST $H$, a lexicon $L$,
and a language model $G$.
Since $H \circ_\sigma L \circ_\sigma G$ is shared across all
utterances, it is typically $\sigma$-composed and saved.
Each individual FST and the composed FST can also be determinized and minimized
for efficiency, which we do not cover.
Interested readers should refer to \citep{MPR2002} for further details.

%% file: seg.tex
\chapter{Discriminative Segmental Models}
\label{ch:seg}

The problem of prediction in general can be considered
as a search problem.
Given an input $x$, we first construct a set of possible
outputs $\mathcal{Y}(x)$, referred to as the search space.
For every output hypothesis $y$ in the search space $\mathcal{Y}(x)$,
we measure how well the output matches the input,
assigning a weight to each pair $(x, y)$.
Prediction can be considered as finding the hypothesis
$\hat{y}$ such that the weight of $(x, \hat{y})$
is larger than any other pairs. 

Extending the above paradigm, the problem of sequence prediction,
such as speech recognition
as we have seen in Chapter \ref{ch:background},
can be considered as a search problem.
Given an input sequence, the search space in this case
is a set of sequences of connected segments.
The number of such sequences is exponential in
the number of possible segments.
Fortunately, we do not need to store exponentially many such sequences,
and can represent the search space compactly as an finite-state
transducer (FST).
Consider an FST with
vertices corresponding to time points,
and edges corresponding to segments.
A path in the FST corresponds to a sequence of connected segments,
and the set of all paths defined by the FST is the search space.
The FST is weighted,
and the weights assigned to the edges are based
on how well the segments match the input.
Prediction can be considered as finding the maximum-weight path,
because the maximum-weight path, by definition, is a path
that best matches the input.

In this chapter,
we formally define segmental models,
including search spaces constructed from sequences of input vectors,
weight functions that measure how well a segment matches the acoustic signals,
and loss functions used for training segmental models.

\section{Preliminaries}

Let $\mathcal{X}$ be the input space,
the set of all sequences of real-valued vectors,
e.g., log mel filter bank features or
mel frequency cepstral coefficients (MFCCs).
Specifically, for a sequence of $T$ vectors $x = (x_1, \dots, x_T) \in \mathcal{X}$,
each $x_t \in \mathbb{R}^d$ is a $d$-dimensional vector,
also referred to as a \term{frame}, for $t \in \{1, \dots, T\}$.
Let $\mathcal{Y}$ be the output space,
the set of all label sequences,
where each label in a label sequence comes
from a label set $L$, e.g., a phoneme set in the case
of phoneme recognition.
Given any $T$ frames,
a \term{segmentation} of length $K$ is a sequence of
time points $((1 = s_1, t_1), \dots, (s_K, t_K = T))$,
where $s_k \leq t_k$ and $t_k + 1 = s_{k+1}$ for $k \in \{2, \dots, K\}$.
A \term{segment} (typically denoted $e$ in later sections) is a tuple $(\ell, s, t)$ where
$\ell \in L$ is its label, $s$ is the start time,
and $t$ is the end time.

A \term{segmental model} is a tuple $(\Theta, w)$
where $\Theta$ is a set of parameters,
and $w: \mathcal{X} \times E \to \mathbb{R}$ is a weight function
parameterized by $\Theta$ and $E$ is the set of all segment tuples $(\ell, s, t)$.
A sequence of segments forms a \term{path}.
Specifically, a path of length $K$ is a sequence of segments
$(e_1, \dots, e_K)$, where $e_k \in E$ for $k \in \{1, \dots, K\}$.
Let $\mathcal{P}$ be the set of all paths.
For any path $p$, we overload $w$ such that
$w(x, p) = \sum_{e \in p} w(x, e)$.
We will also abbreviate $w(x, e)$ and $w(x, p)$
as $w(e)$ and $w(p)$ respectively when the context
is clear.  The concrete form of the weight function
will be defined in later sections.

Given an input $x \in \mathcal{X}$,
segmental models aim to solve sequence prediction
by reducing it to finding the maximum-weight path
\begin{equation} \label{eq:inf}
\argmax_{p \in \mathcal{P}} w(x, p).
\end{equation}
The set of paths $\mathcal{P}$, also
referred to as the \term{search space},
can be compactly represented as an FST.
Once we have the search space FST, we can simply invoke
Algorithm~\ref{alg:shortest} to find the maximum-weight path.
We will describe how the search space FST is constructed
in the next section.

A segmental model can be trained by finding
a set of parameters that minimizes a loss function.
We emphasize that the model definition
is not tied to any loss function,
allowing us to study the behavior of segmental models
under different loss functions.
We define various loss functions
for training segmental models in Section \ref{sec:losses}, and
discuss how the properties of the losses
affect the training requirement.

\section{Search Space}
\label{sec:search-space}

To represent the set of paths $\mathcal{P}$
as an FST, we place a vertex at every time point
and connect vertices based on the set of segments.
Suppose we have $T$ frames.
The set of segments $E$ is an exhaustive enumeration
of tuples $(\ell, s, t)$ for all $\ell \in L$ and
$1 \leq s \leq t \leq T$.
We create a set of vertices $V = \{v_0, v_1, \dots, v_T\}$
and a time function $\tau: V \to \mathbb{N}$
such that $\tau(v_t) = t$ for $t \in \{0, 1, \dots, T\}$.
For every segment $(\ell, s, t) \in E$, we create
an edge $e$ such that $i(e) = o(e) = \ell$,
$\tail(e) = v_{s-1}$, and $\head(e) = v_t$.
In other words, for any $e \in E$, the corresponding segment
$(\ell, s, t) = (o(e), \tau(\tail(e)), \tau(\head(e)))$.
As a result, we will use $w(e)$ and $w((\ell, s, t))$ interchangeably.
We set $\Sigma = \Lambda = L$, $I = \{v_0\}$, and $F = \{v_T\}$
to complete the construction of the FST
given any $T$ frames.

\begin{figure}
\begin{center}
\begin{tikzpicture}[ver/.style={circle,draw}]
\node [ver, ultra thick] (x0) at (0, 0) {};
\node [ver] (x1) at (2, 0) {};
\node [ver] (x2) at (4, 0) {};
\node [ver] (x3) at (6, 0) {};
\node [ver] (x4) at (8, 0) {};
\node [ver, double] (x5) at (10, 0) {};

\draw[->] (x0) edge (x1);
\draw[->] (x1) edge (x2);
\draw[->] (x2) edge (x3);
\draw[->] (x3) edge (x4);
\draw[->] (x4) edge (x5);

\draw[->] (x0) edge [out=10,in=170] (x1);
\draw[->] (x1) edge [out=10,in=170] (x2);
\draw[->] (x2) edge [out=10,in=170] (x3);
\draw[->] (x3) edge [out=10,in=170] (x4);
\draw[->] (x4) edge [out=10,in=170] (x5);

\draw[->] (x0) edge [out=-10,in=190] (x1);
\draw[->] (x1) edge [out=-10,in=190] (x2);
\draw[->] (x2) edge [out=-10,in=190] (x3);
\draw[->] (x3) edge [out=-10,in=190] (x4);
\draw[->] (x4) edge [out=-10,in=190] (x5);

\draw[->] (x0) edge [out=25,in=155] (x2);
\draw[->] (x1) edge [out=25,in=155] (x3);
\draw[->] (x2) edge [out=25,in=155] (x4);
\draw[->] (x3) edge [out=25,in=155] (x5);

\draw[->] (x0) edge [out=30,in=150] (x2);
\draw[->] (x1) edge [out=30,in=150] (x3);
\draw[->] (x2) edge [out=30,in=150] (x4);
\draw[->] (x3) edge [out=30,in=150] (x5);

\draw[->] (x0) edge [out=35,in=145] (x2);
\draw[->] (x1) edge [out=35,in=145] (x3);
\draw[->] (x2) edge [out=35,in=145] (x4);
\draw[->] (x3) edge [out=35,in=145] (x5);

\draw[->] (x0) edge [out=45,in=135] (x3);
\draw[->] (x1) edge [out=45,in=135] (x4);
\draw[->] (x2) edge [out=45,in=135] (x5);

\draw[->] (x0) edge [out=50,in=130] (x3);
\draw[->] (x1) edge [out=50,in=130] (x4);
\draw[->] (x2) edge [out=50,in=130] (x5);

\draw[->] (x0) edge [out=55,in=125] (x3);
\draw[->] (x1) edge [out=55,in=125] (x4);
\draw[->] (x2) edge [out=55,in=125] (x5);

\node (y0) at (1, -0.8) {$\begin{bmatrix}\vdots\end{bmatrix}$};
\node (y1) at (3, -0.8) {$\begin{bmatrix}\vdots\end{bmatrix}$};
\node (y2) at (5, -0.8) {$\begin{bmatrix}\vdots\end{bmatrix}$};
\node (y3) at (7, -0.8) {$\begin{bmatrix}\vdots\end{bmatrix}$};
\node (y4) at (9, -0.8) {$\begin{bmatrix}\vdots\end{bmatrix}$};

\node at (1, -1.5) {$x_1$};
\node at (3, -1.5) {$x_2$};
\node at (5, -1.5) {$x_3$};
\node at (7, -1.5) {$x_4$};
\node at (9, -1.5) {$x_5$};

\end{tikzpicture}
\caption{An example search space for a five-frame input sequence
    with a label set $L$ of size three and maximum segment duration $D$ of two frames,
    i.e., $T = 5$, $|L| = 3$, and $D = 3$.
    The three edges between any two nodes are associated with the three labels.
}
\label{fig:search-space}
\end{center}
\end{figure}
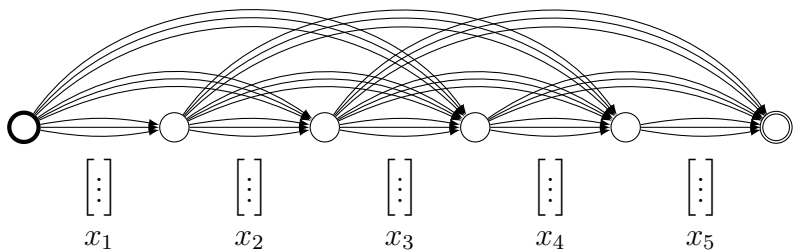

The number of segments in the graph is $O(|L|T^2)$.
To reduce the size of the search space,
a maximum duration constraint is typically imposed
while creating the search space.  Specifically,
we only create segments $(\ell, s, t)$ with $t - s + 1 \leq D$,
for some maximum duration $D$.
Adding such constraint makes the number of segments $O(|L|TD)$.
An example of a search space
is shown in Figure~\ref{fig:search-space}.

\section{Weight Functions}

Here we detail two types of weight
functions based on prior work by ourselves and others \citep{HF2012,TWGL2015,ADYJ2013,LKDSR2016}.
We will compare the two weight functions and in Chapter \ref{ch:e2e}.
The term feature function is often used in the literature
to denote the function $\phi: \mathcal{X} \times E \to \mathbb{R}^m$
for some $m$, when the weight function $w(x, e)$ is of the form
$\theta^\top \phi(x, e)$ for some parameter vector $\theta \in \mathbb{R}^m$.
In general, the weight function
need not be a dot product, but can be any \mbox{(sub-)differentiable} real-valued
function.

A weight function is typically a composition of two steps:
first the input vectors are converted into an intermediate representation,
and second the intermediate representation is converted into
segment weights.
The function in the first step that converts
input vectors to an intermediate representation
are called an \term{encoder}.
Specifically, an encoder is a function
that takes in $T$ frames $x_1, \dots, x_T$
and outputs $\tilde{T}$ feature vectors $h_1, \dots, h_{\tilde{T}}$.
The number of output vectors $\tilde{T}$ can be different from the number of
input vectors $T$ depending on the encoder architecture.
We defer the actual implementation of encoders
in the experimental sections.  Here we define weight functions based on
$h_1, \dots, h_{\tilde{T}}$.
We use $\Theta_\text{enc}$ to denote the parameters for
the encoder, and let $\Theta_\text{dec}$ be
the remaining parameters in the weight function.
Note that $\Theta = \Theta_\text{enc} \cup \Theta_\text{dec}$.

\subsection{FC weight function}
\label{sec:fc-weight}

The following weight function,
termed FC weight,
is based on a frame classifier and is similar to weight functions used
in a variety of prior work~\citep{HF2012,TWGL2015,ADYJ2013}.
The frame classifier takes
in the output $h_1, \dots, h_{\tilde{T}}$ from the encoder,
and produces
a sequence of log probability vectors over the labels
\begin{equation}
z_i = \text{logsoftmax}(W h_i + b)
\end{equation}
where $z_i \in \mathbb{R}^{|L|}$ and $W$ and $b$ are the parameters,
for $i \in \{1, \dots \tilde{T}\}$.
Based on these posterior vectors, we define several functions that summarize the posteriors over a segment:

\paragraph{frame average} The average of transformed log probabilities
\begin{equation}
w_{\text{avg}}((\ell, s, t)) = \frac{1}{t - s + 1} \sum_{i=s}^{t} (u_{i})_{\ell},
\end{equation}
where $u_i = W_\text{avg} z_i$ for $i \in \{1, \dots, \tilde{T}\}$.
\paragraph{frame samples} A sample of transformed log probabilities
\begin{equation}
w_{\text{spl-}j}((\ell, s, t)) = (u_{j})_{\ell}
\end{equation}
at time $j \in \{(t - s)/6, (t - s)/2, 5(t - s)/6\}$,
where $u_i = W_\text{spl} z_i$ for $i \in \{1, \dots, \tilde{T}\}$.
\paragraph{boundary} The average of transformed log probabilities
around the left boundary (start) of the segment
\begin{equation}
w_{\text{left}-k}((\ell, s, t)) = (u_{i-k})_{\ell}
\end{equation}
and around the right boundary (end)
\begin{equation}
w_{\text{right}-k}((\ell, s, t)) = (u'_{i+k})_{\ell}
\end{equation}
where $u_i = W_\text{left} z_i$ and
$u'_i = W_\text{right} z_i$ for $i \in \{1, \dots, \tilde{T}\}$,
and $k = 1, 2, 3$.
\paragraph{duration} The label-dependent duration weight
\begin{equation}
w_{\text{dur}}((\ell, s, t)) = d_{\ell, t - s}.
\end{equation}
\paragraph{bias} A label-dependent bias
\begin{equation}
w_{\text{bias}}((\ell, s, t)) = b'_{\ell}.
\end{equation}

\paragraph{} \hspace{0pt}
The final FC weight function is the sum
of all the above weight functions.
When the FC weight function is used,
$\Theta_\text{dec}$ is $\{W_\text{avg}, W_\text{spl},
W_\text{left}, W_\text{right}, d, b'\}$.
Note that $\{W, b\}$ are considered parameters
of the encoders.

\subsection{MLP weight function}

The MLP weight function based on a multi-layer perceptron (MLP)
is inspired by~\citep{TWGL2015,KDS2016,LKDSR2016}.
Two hidden layers
\begin{align}
z^{(1)}_{\ell, s, t} & = \text{ReLU}(W_1 [ h_s; h_t; c_\ell; d_{\lfloor\log_{1.6}(t-s)\rfloor}] + b_1) \\
z^{(2)}_{\ell, s, t} & = \tanh(W_2 z^{(1)}_{\ell, s, t} + b_2)
\end{align}
are computed directly from the outputs $h_1, \dots, h_{\tilde{T}}$
of the encoder before computing the final weight,
where $c_\ell$ is a label embedding vector for the label $\ell$,
$d_k$ is a duration embedding vector for the
duration $k$ in log scale\footnote{We use 5 different
duration embedding vectors for our experiments. The base 1.6 is chosen such that
$k \in \{0, \dots, 4\}$.}, with $k = \lfloor\log_{1.6}(t-s+1)\rfloor$,
and $\text{ReLU}(x) = \max(x, 0)$.
The final weight for the segment is defined as
\begin{align}
w((\ell, s, t)) & = \theta^\top z^{(2)}_{\ell, s, t}.
\end{align}
When the MLP weight function is used,
$\Theta_\text{dec}$ is $\{W_1, b_1, W_2, b_2, \theta\}$.
Although the MLP weight function is conceptually simple,
it is more expensive to compute
than the FC weight function.
In~\citep{KDS2016,LKDSR2016}, an LSTM
is created for each segment consuming
the outputs of the encoder, followed by
an MLP taking the output vectors of the per segment LSTM
at the segment boundary.
In order to reduce the computation,
we discard the per segment LSTM,
and use the vectors produced by the encoder at the segment boundary
as input to the MLP.

\section{Losses}
\label{sec:losses}

Recall that a path $p = ((\ell_1, s_1, t_1), \dots, (\ell_K, s_K, t_K))$
consists of a label sequence $y = (\ell_1, \dots, \ell_K)$
and a segmentation $z = ((s_1, t_1), \dots, (s_K, t_K))$.
In the following, we will use $(y, z)$ and $p$ interchangeably.
We will also denote the space of all segmentations $\mathcal{Z}$.

Training a segmental model aims to find a set of parameters $\Theta$
that minimizes the expected task loss, for example, the expected edit distance
\begin{equation}
\mathbb{E}_{(x, y) \sim \mathcal{D}}[\edit(h_\Theta(x), y)]
\end{equation}
where $h$ is the inference algorithm (Algorithm~\ref{alg:shortest})
parameterized with $\Theta$, $\edit$ computes the edit distance
between two sequences,
and the expectation is taken over samples $(x, y) \in \mathcal{X} \times \mathcal{Y}$
drawn from a distribution $\mathcal{D}$.
The expectation can be decomposed into two steps
\begin{equation}
\mathbb{E}_{x \sim \mathcal{D}(x)} \mathbb{E}_{y \sim \mathcal{D}(y|x)}[\edit(h_\Theta(x), y)],
\end{equation}
first sampling $x$ and then sampling $y$.
To obtain a good discriminator $h$, it suffices to optimize
the inner expectation
\begin{equation}
\mathbb{E}_{y \sim \mathcal{D}(y|x)}[\edit(h_\Theta(x), y)],
\end{equation}
for any $x$ drawn from $\mathcal{D}(x)$.
However, the edit distance, due to its discrete nature, is difficult to optimize,
so instead we minimize the expected loss
\begin{equation} \label{eq:exp-loss}
\mathbb{E}_{y \sim \mathcal{D}(y|x)}[\mathcal{L}(\Theta; x, y)],
\end{equation}
where $\mathcal{L}$ is a surrogate loss function.
If segmentations are considered in the loss function,
then we can minimize
\begin{equation} \label{eq:exp-seg-loss}
\mathbb{E}_{(y, z) \sim \mathcal{D}'(y, z | x)}[\mathcal{L}(\Theta; x, y, z)],
\end{equation}
where $\mathcal{D}'$ is a conditional distribution over
$\mathcal{X} \times \mathcal{Y} \times \mathcal{Z}$.

Since the distribution $\mathcal{D}$ is unknown,
we use a tranining set $S = \{(x_1, y_1), \dots, (x_n, y_n)\}$
of size $n$ to approximate the expectation and instead minimize
\begin{equation}
\frac{1}{n} \sum_{i=1}^n \mathcal{L}(\Theta; x_i, y_i).
\end{equation}
If segmentations are considered,
we use a tranining set $S = \{(x_1, y_1, z_1), \dots, (x_n, y_n, z_n)\}$
of size $n$ to approximate $\mathcal{D}'$ and minimize
\begin{equation}
\frac{1}{n} \sum_{i=1}^n \mathcal{L}(\Theta; x_i, y_i, z_i).
\end{equation}

The connection between the surrogate loss $\mathcal{L}$
and the edit distance depends on the choice of loss.
We will optimize the loss functions with first-order methods,
such as stochastic gradient descent. When listing the
function definitions along with reasons for using them,
we will also list the \mbox{(sub-)gradients} with respect to
the weight $w(e)$ for some edge $e$.
We assume the weight function $w$ is \mbox{(sub-)differentiable} and
the \mbox{(sub-)gradients} with respect to the parameters can be obtained with back-propagation.

\subsection{Hinge loss}

Given an utterance $x$ and a ground truth path $p = (y, z)$,
the hinge loss is defined as
\begin{equation}
\mathcal{L}_{\text{hinge}}(\Theta; x, p)
    = \max_{p' \in \mathcal{P}} \left[ \cost(p', p) - w(p) + w(p') \right]
\end{equation}
where $\cost$ is a user-defined cost function.
The connection between the hinge loss and the task loss is through
the cost function.
Suppose $\hat{p} = \argmax_{p \in \mathcal{P}} w(p)$ is
the maximum-weight path found by Algorithm~\ref{alg:shortest}.
The cost of the inferred path $\hat{p}$ against the ground truth
$p$ can be upper-bounded by the hinge loss:
\begin{align}
\cost(\hat{p}, p) \leq \cost(\hat{p}, p) - w(p) + w(\hat{p})
    \leq \mathcal{L}_{\text{hinge}}(\Theta; x, p).
\end{align}
When the cost function is the edit distance, minimizing
the hinge loss is minimizing an upper bound on the edit distance
of the predicted sequence.

The hinge loss itself is difficult to
optimize when the cost function is the edit distance.
In practice, the cost function is assumed to be decomposable
\begin{equation}
\cost(p', p) = \sum_{e' \in p'} \cost(e', p)
\end{equation}
to allow efficient dynamic programming.
When the cost is decomposable, the hinge loss can be written as
\begin{align}
\mathcal{L}_{\text{hinge}}(\Theta; x, p)
    & = \max_{p' \in \mathcal{P}} \left[ \sum_{e' \in p'} \cost(e', p)
        - \sum_{e \in p} w(e) + \sum_{e' \in p'} w(e') \right] \\
    & = \max_{p' \in \mathcal{P}} \sum_{e' \in p'} \left[ \cost(e', p)
        + w(e') \right] - \sum_{e \in p} w(e),
\end{align}
and the $\max$ operator in the first term
can be solved with Algorithm~\ref{alg:shortest}
by adding the costs to the weights for all segments.

A subgradient of the hinge loss with respect to $w(e)$ is
\begin{equation}
\frac{\partial\mathcal{L}_{\text{hinge}}(\Theta; x, p)}{\partial w(e)}
    = -\ind_{e \in p} + \ind_{e \in \tilde{p}}
\end{equation}
where
\begin{equation}
\tilde{p} = \argmax_{p' \in \mathcal{P}} [\cost(p', p) + w(p')],
\end{equation}
which is the path that maximizes the first term in the hinge loss,
and can be obtained with Algorithm~\ref{alg:shortest}
with the cost added.

Linear models are called support vector machines (SVM)
when trained with the hinge loss, and are referred to
as structured SVMs when used for solving structured prediction
problems, e.g., sequence prediction in our case.
A hinge loss with cost zero is also called perceptron loss \citep{LH2005}.
Segmental models trained with the hinge loss have been studied
in~\citep{ZG2013,TGL2014,TWGL2015}.

\subsection{Log loss}

Segmental models can be treated as probabilistic
models by defining probability distributions on
the set of all paths $\mathcal{P}$. Specifically, the probability
of a path $p = (y, z)$ is defined as
\begin{equation}
P(y, z | x) = P(p | x) = \frac{1}{Z(x)} \exp(w(x, p))
\end{equation}
where
\begin{equation}
Z(x) = \sum_{p' \in \mathcal{P}} \exp(w(x, p'))
\end{equation}
is the partition function.
Given an input $x$ and a ground truth path $p$, the log loss is defined as
\begin{equation}
\mathcal{L}_{\text{log}}(\Theta; x, p) = -\log P(p | x).
\end{equation}
Minimizing the log loss is equivalent to maximizing the conditional likelihood.
In addition, the conditional likelihood can be written as
\begin{align}
P(y, z | x) & = \mathbb{E}_{(y', z') \sim P(y', z' | x)}[\ind_{(y', z') = (y, z)}] \\
    & = 1 - \mathbb{E}_{(y', z') \sim P(y', z' | x)}[\ind_{(y', z') \neq (y, z)}].
\end{align}
Therefore, maximizing the conditional likelihood is equivalent to minimizing
the expected zero-one loss
\begin{equation}
\mathbb{E}_{(y', z') \sim P(y', z' | x)}[\ind_{(y', z') \neq (y, z)}],
\end{equation}
where $P(y, z | x)$ is used to approximate $\mathcal{D}'(y, z | x)$ and
the zero-one loss $\ind_{(y', z') \neq (y, z)}$ is used.
The use of the log loss is justified because the expectation above
can be seen as an approximation of \eqref{eq:exp-seg-loss}.
Segmental models trained with log loss
have been referred to as semi-Markov CRFs~\citep{SC2005}.
Minimizing log loss is equivalent to maximizing mutual information,
and is commonly referred to as the MMI criterion \citep{BBSM1986}.

Since the weight for the ground truth path $p$ can be efficiently
computed, we are left with the problem of computing the partition function $Z(x)$.
The partition function can also be computed efficiently with the following
dynamic programming algorithm.
Let $\mathcal{P}(u, v)$ be the set of paths
that start at vertex $u$ and end at vertex $v$.
For any vertex $v$, define the forward marginal as
\begin{equation}
\alpha(v) = \log \sum_{p' \in \mathcal{P}(v_0, v)} \exp(w(p')).
\end{equation}
By expanding the edges ending at $v$, we have
\begin{align}
\alpha(v) &= \log \sum_{p' \in \mathcal{P}(v_0, v)} \exp\left(\sum_{e \in p'} w(e)\right) \\
    &= \log \sum_{e \in \inedges(v)} \sum_{p' \in \mathcal{P}(v_0, \tail(e))}
        \exp\left(w(e) + \sum_{e' \in p'} w(e')\right) \\
    &= \log \sum_{e \in \inedges(v)} \exp\left(w(e) + \alpha(\tail(e))\right)
\end{align}
Similarly, the backward marginal at $v$ is defined as
\begin{equation}
\beta(v) = \log \sum_{p' \in \mathcal{P}(v, v_T)} \exp(w(p')),
\end{equation}
and has similar recursive structure.
The complete algorithm is shown in Algorithm~\ref{alg:fb}.
Once all entries in $\alpha$ and $\beta$ are computed,
the log partition function is
\begin{equation}
\log Z(x) = \alpha(v_T) = \beta(v_0).
\end{equation}
Note that we store all of the entries in log space to maintain
numerical stability.

The gradient of the log loss with respect to $w(e)$ is
\begin{align}
\frac{\partial\mathcal{L}_{\text{log}}(\Theta; x, p)}{\partial w(e)}
    & = -\ind_{e \in p} + \frac{1}{Z(x)}\sum_{p' \ni e} \exp(w(p')) \\
    & = -\ind_{e \in p} + \exp \Big[ \alpha(\tail(e)) + w(e)
        + \beta(\head(e)) - \log Z(x) \Big],
\end{align}
which can also be efficiently computed once
the marginals are computed.

Note that the form of Algorithm~\ref{alg:fb} is very similar
to that of Algorithm~\ref{alg:shortest}.
Recall that to get the standard shortest-path algorithm (Algorithm \ref{alg:shortest}),
we simply use the tropical semiring when running Algorithm~\ref{alg:gen-shortest}.
If the tropical semiring is replaced by the log semiring,
we arrive at Algorithm~\ref{alg:fb},
demonstrating the generality of FST algorithms
and semirings.

\begin{algorithm}
\caption{Computing forward and backward marginals} \label{alg:fb}
\begin{algorithmic}
\State $\alpha(v_0) = 0$
\State $\beta(v_T) = 0$
\State $\text{logadd}(a, b) = \log(\exp(a) + \exp(b))$
\For{$v = v_0, v_1, \dots, v_T$}
    \State $\alpha(v) = \text{logadd}_{e \in \inedges(v)} \Big[ \alpha(\tail(e)) + w(e) \Big]$
\EndFor
\For{$v = v_T, v_{T-1}, \dots, v_0$}
    \State $\beta(v) = \text{logadd}_{e \in \outedges(v)} \Big[ \beta(\head(e)) + w(e) \Big]$
\EndFor
\end{algorithmic}
\end{algorithm}

Log loss can be modified to include a cost function.
Speficically, we define
\begin{equation}
P^+_{p}(p' | x) = \frac{1}{Z(x, p)} \exp(w(x, p') + \cost(p', p))
\end{equation}
where $p$ is the ground-truth path and
\begin{equation}
Z(x, p) = \sum_{p'' \in \mathcal{P}} \exp(w(x, p'') + \cost(p'', p)).
\end{equation}
Unlike $P$, the distribution $P^+$ considers both the weights and the costs.
The modified log loss, also known as the boosted MMI criterion \citep{P+2008},
is defined as
\begin{align}
\mathcal{L}_{\text{bMMI}}(\Theta; x, p) = -\log P^+_{p}(p | x).
\end{align} 

\subsection{Marginal log loss}

Given an input $x$ and a label sequence $y$,
the marginal log loss is defined as
\begin{equation}
\mathcal{L}_{\text{mll}}(\Theta; x, y) = -\log P(y | x) = -\log \sum_{z \in \mathcal{Z}} P(y, z | x)
\end{equation}
where the segmentation is marginalized compared to log loss;
hence the name.
Following the same argument as for the log loss,
the marginal distribution
can be written as
\begin{equation}
P(y | x) = 1 - \mathbb{E}_{y' \sim P(y' | x)}[\ind_{y \neq y'}],
\end{equation}
and maximizing the marginal distribution is equivalent to
minimizing the expected zero-one loss
\begin{equation}
\mathbb{E}_{y' \sim P(y' | x)}[\ind_{y \neq y'}],
\end{equation}
where $P(y | x)$ is used to approximate $\mathcal{D}(y | x)$.
Note that the zero-one loss $\ind_{y \neq y'}$ only depends on the label sequence.
While the log loss has a connection to \eqref{eq:exp-seg-loss},
the marginal log loss justifies its use by directly approximating \eqref{eq:exp-loss}
with the above expected zero-one loss.

Note that both the hinge loss and the log loss depend on the ground-truth
path, or more specifically, the ground-truth segmentation.
The marginal log loss marginalizes over the segmentations, so
it only depends on the ground-truth label sequence and
not the segmentation.  The lack of reliance on the ground-truth
segmentation makes the marginal log loss attractive for
tasks such as speech recognition, because collecting
ground-truth segmentations for phonemes or words is time-consuming
and/or expensive.  In addition, the boundaries of phonemes and words
tend to be ambiguous, so it can be preferable to leave the decision
to the model.
Segmental models trained with the marginal log loss have
been referred to as segmental CRFs~\citep{ZN2009}.

To compute the marginal log loss,
we can rewrite it as
\begin{align}
\mathcal{L}_{\text{mll}}(\Theta; x, y) & = -\log \sum_{z \in \mathcal{Z}} P(y, z | x) \\
    & = -\log \sum_{z \in \mathcal{Z}} \exp(w(x, (y, z))) + \log Z(x) \\
    & = -\underbrace{\log \sum_{p': \Gamma(p') = y} \exp(w(x, p'))}_{\log Z(x, y)} + \log Z(x)
\end{align}
where $\Gamma$ extracts the label sequence from a path, i.e.,
for $p' = (y', z')$, $\Gamma(p') = y'$.
Since the partition function can be efficiently computed
from Algorithm~\ref{alg:fb}, we only need to compute
$\log Z(x, y)$. Since the term $\log Z(x, y)$ is identical to $\log Z(x)$
except that it involves a constrained 
search space considering all paths with the same label sequence $y$,
the strategy is to construct the constrained search space
with an FST and run Algorithm~\ref{alg:fb} on the FST.
Let $F$ be a chain FST that represents $y$,
with edges $\{e_1, \dots, e_{|y|}\}$,
where $i(e_k) = o(e_k) = y_k$ for all $k \in \{1, \dots, |y|\}$.
Let $G$ be the search space consisting of all paths
in $\mathcal{P}$. The term $\log Z(x, y)$
can be efficiently computed by running Algorithm~\ref{alg:fb} on
the $\sigma$-composition of $G$ and $F$, i.e., $G \circ_\sigma F$.
Note that after $\sigma$-composing $G$ and $F$,
we only allow paths in $G$ that can produce
output sequences that $F$ accepts.
Since $F$ only accepts $y$, the paths in $G \circ_\sigma F$
are all the paths in $G$ that produce $y$ as desired.
Let the forward and backward marginals computed on $G \circ_\sigma F$
be $\alpha'$ and $\beta'$.
We have $\log Z(x, y) = \alpha'(v_T) = \beta'(v_0)$.

The gradient of the marginal log loss with respect to $w(e)$ is
\begin{align}
\frac{\partial\mathcal{L}_{\text{mll}}(\Theta; x, y)}{\partial w(e)}
    & = -\frac{1}{Z(x, y)} \sum_{\substack{p' \ni e\\ \Gamma(p') = y}} \exp(w(p'))
        + \frac{1}{Z(x)}\sum_{p' \ni e} \exp(w(p')) \\
    & = - \exp \Big[ \alpha'(\tail(e)) + w(e) + \beta'(\head(e)) - \log Z(x, y) \Big] \notag\\
    & \qquad {} + \exp \Big[ \alpha(\tail(e)) + w(e) + \beta(\head(e)) - \log Z(x) \Big].
\end{align}
and can be efficiently computed once all of the marginals are computed.

\subsection{Empirical Bayes risk}

The empirical Bayes risk (EBR) \citep{SE2006} is defined as
\begin{align}
\mathcal{L}_{\text{ebr}}(\Theta; x, p) = \mathbb{E}_{p' \sim P(p'|x)}[\cost(p', p)].
\end{align}
The intuition behind EBR is that if $P$ is a good approximation for $\mathcal{D}'$,
then EBR is a good approximation for
\begin{align*}
\mathbb{E}_{p' \sim \mathcal{D}'(p' | x)}[\cost(p', p)].
\end{align*}
The goal is to find $P$ that achieves a low expected cost.
Empirical Bayes risk is also referred to as minimum phone error (MPE)
when the cost is based on phone errors,
and is referred to as minimum word error (MWE) when the cost is based on
word errors respectively \citep{PW2002}.
A boosted version of EBR can be obtained by replacing $P$ with $P^+$ \citep{MN2008}.

Since $Z(x)$ can be computed through the forward and backward marginals $\alpha$ and $\beta$,
we are left with $\sum_{p'} \exp(w(p'))\cost(p', p)$.
Similar to the forward and backward marginals, we have
\begin{align}
\alpha''(v) & = \log \sum_{p' \in \mathcal{P}(v_0, v)} \exp(w(p') + \log \cost(p', p)) \\
    & = \log \sum_{e' \in \inedges(v)} \exp(w(e') + \log \cost(e', p))
        \sum_{p' \in \mathcal{P}(v_0, \tail(e))} \exp(w(p') + \log \cost(p', p)) \\
    & = \log\sum_{e' \in \inedges(v)} \exp(w(e') + \log\cost(e', p) + \alpha''(\tail(e))) \\
\beta''(v) & = \log \sum_{p' \in \mathcal{P}(v, v_T)} \exp(w(p') + \log \cost(p', p)) \\
    & = \log \sum_{e' \in \outedges(v)} \exp(w(e') + \log\cost(e', p))
        \sum_{p' \in \mathcal{P}(\head(e), v_T)} \exp(w(p') + \log\cost(p', p)) \\
    & = \log\sum_{e' \in \outedges(v)} \exp(w(e') + \log\cost(e', p) + \beta''(\head(e)))
\end{align}
where $\cost$ is assumed to be non-negative, and the marginals
are stored in log space to maintain numerical stability.
Given the cost-augmented marginals, we have
\begin{align}
\mathbb{E}_{p' \sim P(p'|x)}[\cost(p', p)] = \exp(\alpha''(v_T) - \log Z(x))
    = \exp(\beta''(v_0) - \log Z(x)).
\end{align}

The gradient of EBR with respect to $w(e)$ of some edge $e$ is
\begin{align}
\frac{\mathcal{L}_{\text{ebr}}(\Theta; x, p)}{w(e)}
    & = \frac{\sum_{p' \ni e} \exp(w(p'))\cost(p', p)}{Z(x)}
        - \frac{\left[\sum_{p'} \exp(w(p'))\cost(p', p)\right]\left[\sum_{p' \ni e}\exp(w(p'))\right]}{(Z(x))^2} \\
    & = \exp(\alpha''(\tail(e)) + w(e) + \log \cost(e, p) + \beta''(\head(e)) - \log Z(x)) \notag\\
    & \quad{} - \mathcal{L}_{\text{ebr}}(\Theta; x, p) [\exp(\alpha(\tail(e)) + w(e) + \beta(\head(e)) - \log Z(x))]
\end{align}
It is also interesting to note that, in general,
\begin{align}
\frac{\partial}{\partial w(e)} \mathbb{E}_{p' \sim P(p'|x)}[f(p')]
    &= \mathbb{E}_{p' \sim P(p'|x)}[\mathbbm{1}_{p' \ni e} f(p')]
        - \mathbb{E}_{p' \sim P(p'|x)}[\mathbbm{1}_{p' \ni e}]
        \mathbb{E}_{p' \sim P(p'|x)}[f(p')] \\
    &= \text{Cov}(\mathbbm{1}_{p' \ni e}, f(p')),
\end{align}
for any function $f: \mathcal{P} \to \mathbb{R}$.

\subsection{Ramp loss}

The ramp loss \citep{CSWB2006} is defined as
\begin{align}
\mathcal{L}_{\text{ramp}}(\Theta; x, p) = \max_{p'}[\cost(p', p) + w(p') - \max_{p''}w(p'')].
\end{align}
It is easy to see that
\begin{align}
\mathcal{L}_{\text{hinge}}(\Theta; x, p)
    & = \max_{p'}[\cost(p', p) + w(p') - w(p)] \\
    & \geq \max_{p'}[\cost(p', p) + w(p') - \max_{p''}w(p'')]
        = \mathcal{L}_{\text{ramp}}(\Theta; x, p).
\end{align}
In addition, for $\hat{p} = \argmax_{p'} w(p')$,
\begin{align}
\cost(\hat{p}, p) &= \cost(\hat{p}, p) + w(\hat{p}) - w(\hat{p}) \\
    & \leq \max_{p'}[\cost(p', p) + w(p')] - w(\hat{p}) \\
    & = \max_{p'}[\cost(p', p) + w(p') ] - \max_{p'} w(p')
    = \mathcal{L}_{\text{ramp}}(\Theta; x, p).
\end{align}
Therefore, we have
\begin{align}
\mathcal{L}_{\text{hinge}}(\Theta; x, p) \geq \mathcal{L}_{\text{ramp}}(\Theta; x, p)
    \geq \cost(\hat{p}, p).
\end{align}
In other words, ramp loss is also an upper bound on the cost of the predicted sequence,
but is tighter than hinge loss.
However, optimizing the ramp loss is more complicated.
Ramp loss can be minimized with the concave-convex
procedure (CCCP) \citep{YR2003}.  See \citep{GS2012} for a detailed implementation
of CCCP for solving ramp loss.

\subsection{Connections between losses}

Log loss augmented with a cost function, or boosted MMI, can be
seen as a soft version of hinge loss.
By the definition of boosted MMI, where
\begin{align}
\mathcal{L}_{\text{bMMI}}(\Theta; x, p) = -\log P(p | x)
    = -w(p) + \log \sum_{p'} \exp(w(p') + \cost(p', p)),
\end{align}
we can see that the second term acts as a soft-max instead of
a max function.
To see this, note that
\begin{align}
\log (e^{x_1} + e^{x_2} + \dots + e^{x_n})
    & = \log e^{x_{\text{max}}} (e^{x_1 - x_{\text{max}}} + \dots + e^{x_n - x_{\text{max}}}) \\
    & = x_{\text{max}} + \log (e^{x_1 - x_{\text{max}}} + \dots + e^{x_n - x_{\text{max}}}),
    \label{eq:log-sum-exp}
\end{align}
where $x_{\text{max}} = \max_i x_i$.
The second term in \eqref{eq:log-sum-exp} is small when
the difference between $x_{\text{max}}$ and other $x_i$'s is large.
In fact, we can introduce an addition factor $\epsilon$
to control this effect.  Specifically,
\begin{align}
\epsilon \log (e^{x_1/\epsilon} + e^{x_2/\epsilon} + \dots + e^{x_n\epsilon})
    & = x_{\text{max}} + \epsilon \log (e^{(x_1 - x_{\text{max})/\epsilon}} + \dots + e^{(x_n - x_{\text{max}})/\epsilon}).
    \label{eq:eps-log-sum-exp}
\end{align}
When $\epsilon \to 0$, the second term vanishes and
\eqref{eq:eps-log-sum-exp} acts exactly like a max function.
We can add the additional $\epsilon$ factor to the cost-augmented log loss
and get
\begin{align}
\mathcal{L}^{\epsilon}_{\text{bMMI}}(\Theta; x, p) = -\log P(p | x)
    = -w(p) + \epsilon\log \sum_{p'} \exp\left(\frac{1}{\epsilon}(w(p') + \cost(p', p))\right).
\end{align}
As $\epsilon \to 0$, $\mathcal{L}^{\epsilon}_{\text{bMMI}} \to \mathcal{L}_{\text{hinge}}$.
This connection was first presented in \citep{HDSN2008}.
\cite{GS2010} independently proposed softmax-margin,
the equivalence of cost-augmented log loss.

As we have noted when we introduced log loss,
\begin{align}
P(p | x) = \mathbb{E}_{p' \sim P(p'|x)}[\mathbbm{1}_{p' = p}]
    = 1 - \mathbb{E}_{p' \sim P(p'|x)}[\mathbbm{1}_{p' \neq p}].
\end{align}
Therefore, minimizing log loss is equivalent to minimizing EBR with
$\cost(p', p) = \mathbbm{1}_{p' \neq p}$.

Another connection between cost-augmented log loss and EBR is the following.
Let
\begin{align}
\mathcal{L}_{\lambda, \text{bMMI}}(\Theta; x, p) = -\log P(p | x)
    = -w(p) + \log \sum_{p'} \exp(w(p') + \lambda \cost(p', p)),
\end{align}
where we introduce a scaling factor $\lambda$ for the cost function
in cost-augmented log loss.
If we take the derivative of cost-augmented log loss with respect
to $\lambda$, we have
\begin{align}
\frac{\partial}{\partial \lambda} \mathcal{L}_{\lambda, \text{bMMI}}
    &= \frac{\sum_{p'} [\exp(w(p') + \lambda \cost(p', p)) \cost(p', p)]}{
        \sum_{p'} \exp(w(p') + \lambda \cost(p', p))} \\
    &= \mathbb{E}_{p' \sim P^+_\lambda(p'|x)}[\cost(p', p)],
\end{align}
where
\begin{align}
P^+_{\lambda}(p'|x) = \frac{\exp(w(p') + \lambda\cost(p', p))}{
    \sum_{p''} \exp(w(p'') + \lambda\cost(p'', p))}
\end{align}
Therefore, the derivative of cost-augmented log loss
with respect to $\lambda$ is cost-augmented EBR,
or sometimes referred to as boosted MPE.
This connection was first presented in \citep{MWN2009}.

\section{Preliminary Experiments}

In this section\footnote{These results were published in \citep{TGL2014}},
we describe experiments comparing segmental models
trained with various loss functions and two cost functions.
Instead of using the entire search space, we follow \citep{ZN2009, Z+2010}
and use a baseline recognizer to generate sparse search spaces,
commonly known as \term{lattices}.
Segmental models are then applied on these lattices.
The weight function used in these experiments is a linear function
$w(x, e) = \theta^\top \phi(x, e)$ where $\theta$ is a parameter
vector and $\phi(x, e)$ is a feature vector for some edge $e$.
The feature function $\phi$ is task-dependent, and is described
in detail later.
Hinge loss and ramp loss are sensitive to the scale of the cost,
so the scale of the cost function is tuned.
Specifically, we introduce the parameter $\mu$
in the hinge loss
\begin{align}
\mathcal{L}_{\text{hinge}} = \max_{p'} [\mu \cdot \cost(p', p) - w(p) + w(p')],
\end{align}
and parameters $\mu_1$ and $\mu_2$ for the ramp loss
\begin{align}
\mathcal{L}_{\text{ramp}} = \max_{p'} [\mu_1 \cdot \cost(p', p) + w(p')]
    - \max_{p'}[\mu_2 \cdot \cost(p', p) + w(p')].
\end{align}
Note the additional cost term in the above ramp loss.
This version of ramp loss is a generalization of the standard ramp loss \citep{C2012, GS2012}.

\subsection{Cost functions}

One cost function was proposed in
\citep{PW2002} in the context of MPE/MWE training,
and we refer to it as \term{MPE cost}.
The cost of an edge is the duration of the non-overlapping part of a
matching ground-truth edge that gives the lowest error, where the
error is one if the label is correct and 0.5 otherwise.
Formally, for any hypothesized edge $e'$, we define 
\begin{align}
\cost_{\text{MPE}}(e', p) =
    1 - \max_{e \in p} \left[ \mathbbm{1}_{o(e) = o(e')}
    \frac{|e \cap e'|}{|e|}
    + \frac{1}{2} \mathbbm{1}_{o(e) \ne o(e')}
    \frac{|e \cap e'|}{|e|} \right],
\end{align}
and
\begin{align}
\cost_{\text{MPE}}(p', p) = \sum_{e' \in p'} \cost_{\text{MPE}}(e', p),
\end{align}
where $|e|$ denotes the length of segment $e$.

The MPE cost only penalizes false negatives and does not account for false positives.
Therefore, we propose
an alternative that we refer to as the \term{overlap cost}:
\begin{align}
\cost_{\text{overlap}}(e', p) = 1 - \mathbbm{1}_{o(e) = o(\tilde{e})}
    \frac{|e \cap \tilde{e}|}{|e \cup \tilde{e}|},
\end{align}
where $\tilde{e} = \argmax_{e \in p} |e' \cap e|$.
This cost function finds
the most overlapping edge in the ground truth and 
considers any part of the union of the two edges that is not overlapping to be in error.
The cost for the whole path is again
\begin{align}
\cost_{\text{overlap}}(p', p) = \sum_{e' \in p'} \cost_{\text{overlap}}(e', p).
\end{align}

\subsection{Results}
\label{sec:seg-result}

We study the various losses and cost functions on two tasks.  One is a
standard speech recognition task, namely TIMIT \citep{G+1993} phonetic recognition.
The second is a sign language recognition task from video, in
particular recognition of fingerspelled letter sequences in American
Sign Language (ASL).  Both are tasks on which there is prior work using
semi-Markov CRFs~\citep{Z2012,KSL2013}, and both are small enough (in terms of data
set size and decoding search space) to run many empirical comparisons
in a reasonable amount of time.  For the ASL task, we use the data and
experimental setup of~\citep{KSL2013}: We obtain baseline lattices
using their tandem HMM-based system, and we use the
same set of segmental feature functions.  We use forced
alignments of the ground-truth transcriptions for training.
We train all models from
all-zero weights and optimize with Rprop \citep{RB1993} for 20 epochs.
We use $L_1$ and $L_2$ regularization, with parameters tuned over the
grid $\{0,10^{-6}, 10^{-5}, \dots, 0.1, 1\}^2$.  For hinge and ramp loss,
we use the standard forms without tuning the cost weights (i.e., $\mu=1$,
$\mu_1 = 0$, and $\mu_2 = 1$).

\begin{table}
\begin{center}
\caption{\label{tbl:asl-loss-comp}
    Letter error rates (\%) for a baseline tandem HMM and segmental models trained with various losses
    and cost functions on the ASL data set.}
\begin{tabular}{l|ll|llll|l}

           &            &            & Andy   & Drucie & Rita    & Robin & Avg \\
\hline
\multicolumn{3}{l|}{Tandem HMM} & 13.8   & 7.10   & 26.1    & 11.5 & 14.6  \\
\hline
\hline
\multirow{8}{*}{segmental models}
        &            & log loss   & \textbf{9.9}    & \textbf{6.7}    & 23.9    & \textbf{10.5} & 12.8 \\
\cline{2-8}
        & MPE cost   & hinge loss & 10.5   & \textbf{6.7}    & 23.5    & 10.9 & 12.9 \\
        &            & EBR        & 13.8   & 7.2    & 27.6    & 12.6 & 15.3 \\
        &            & ramp       & 13.8   & 7.2    & 28.6    & 12.7 & 15.6 \\
\cline{2-8}
        & overlap cost
                     & hinge loss & 10.3   & 6.9    & 21.9    & \textbf{10.5} & 12.4 \\ 
        &            & EBR        & 10.4   & 6.8    & 23.5    & 11.6 & 13.1 \\
        &            & ramp       & 12.0   & 6.8    & 25.2    & 11.7 & 13.9 \\
        &            & tuned ramp & 10.1   & 6.8    & \textbf{21.5}    & 10.6 & \textbf{12.3}
\end{tabular}
\end{center}
\end{table}

\begin{table}
\begin{center}
\caption{\label{tbl:timit-loss-comp}
    Phone error rates (\%) for a baseline HMM and segmental models trained with various losses
    and cost functions on the TIMIT data set.}
\begin{tabular}{l|ll|l}

           &            &            & test \\
\hline
\multicolumn{3}{l|}{Tandem HMM} & 30.7 \\
\hline
\hline
\multirow{8}{*}{segmental models}
        &            & log loss   & \textbf{30.1} \\
\cline{2-4}
        & MPE cost   & hinge loss & 30.2 \\
        &            & EBR        & 30.7 \\
        &            & ramp       & 30.2 \\
\cline{2-4}
        & overlap cost
                     & hinge loss & \textbf{30.1} \\
        &            & EBR        & \textbf{30.1} \\
        &            & ramp       & 30.3 
\end{tabular}
\end{center}
\end{table}

For TIMIT, we use the standard 3696-utterance training set and
192-utterance core test set, plus a random 192 utterances from the
full test set (excluding the core test set) as a development set.
We collapse the 61 phones in the phone set to 48 for training,
and further collapse them to 39 for evaluation \citep{LH1989}.

We use lattices generated by a baseline monophone HMM system with
39-dimensional MFCCs.  The resulting lattices have an average density
(average number of hypothesized edges per ground truth edge) of 60.1.
The oracle phone error rate is 6.3\% for the development set and 7.0\%
for the core test set.  We use oracle paths
(paths with minimum phone error) from the lattices as ground truth for training.
We implement segmental models with various feature functions.
The base features are the acoustic
and language model score from the baseline recognizer, and a bias (a feature that is always one).
We also include a set of features based on spectro-temporal receptive
fields implemented as follows.  We begin with 40-dimensional log mel
filter bank features.  For each segment, we divide it evenly into thirds
in both time and frequency, resulting in nine patches for each segment.
For each patch, we have a $3 \times 13$ receptive field of all ones, and
convolve it with the patch.  The resulting $3 \times 13 \times 9$ numbers
are lexicalized to form the final features for the segmental model.
Specifically, for any feature vector $v$, we refer to $v \otimes \mathbf{1}_\ell$
as the lexicalized feature vector, where $\mathbf{1}_\ell$ is a one-hot vector
for the label $\ell$.
We optimize the loss functions using AdaGrad \citep{DHS2011}, using step
size 0.1 for 10 epochs.  $L_1$ and $L_2$ regularization parameters are
tuned over the grid $\{0, 0.001, 0.1, 1\} \times \{0, 0.1, 1, 100\}$.

The results for ASL recognition, averaged over four signers, are shown
in Table~\ref{tbl:asl-loss-comp}.  The evaluation metric
is the letter error rate, which is the percentage of letters that are
substituted, inserted, or deleted.  The results for TIMIT are shown
in Table~\ref{tbl:timit-loss-comp}.  We observe three
consistent conclusions:
\begin{itemize}
\item Segmental models perform significantly better than the baseline HMM.
\item Across losses, overlap cost is better than MPE cost.
\item Hinge loss with overlap cost is the best performer,
    but this is only by a small margin, and
    log loss is very competitive even without using an 
    explicit cost function.
\item Non-convex losses (ramp and EBR) are difficult to optimize
    and therefore achieve inconsistent results.
    We suspect a warm start might be able to remedy this.
\end{itemize}

For the ASL task, we tuned on a development set the cost weights
for ramp loss over the grid $\{-100,-10,-1,-0.1,-0.01\} \times
\{0.01,0.1,1,10,100\}$, using overlap cost.  The test result of tuned
ramp slightly improves over hinge loss, confirming that if ramp is tuned
carefully, it is able to outperform hinge.  However, even though tuned
ramp loss achieves very good results, considering the time spent tuning
$\mu_1$ and $\mu_2$, we still favor hinge and log loss.

We also conducted experiments to determine how the results are affected by different levels of noise in the feature functions, using simulated phone detector-based features.
Similarly to~\citep{Z+2010}, we define a detection event as a (time, phone label) pair,
and a feature function that is
an indicator of whether a phone detection event occurs in
the time span of the edge.
If we set a high weight for the phone event that occurs
in an edge with the same phone label, then we
can exactly recover the oracle path.
This allows us to conduct a series of simulated experiments
with different amounts of noise added to the oracle phone events,
or gold events. 
For all experiments below, we use the same TIMIT setting
except that we only use the acoustic and language model score
with the simulated phone detector features,
with no regularizer and one epoch of AdaGrad.
The ramp cost weights are set to $\mu_1=0, \mu_2=1$. The cost weight for hinge is tuned
over $\{0.01, 0.1, 1, 10, 100\}$ and results are only shown for the best-performing value.

The first set of experiments randomly perturbs the
correct phone label of each event to an incorrect label
with the corruption probability shown on the $x$-axis; the event times are not perturbed. 
The second set of experiments perturbs the time
for each event but not the label.
We add Gaussian noise with mean
set to the time at which the event occurs and with several
standard deviations shown on the $x$-axis.  
For the third and fourth set of experiments, we randomly
include an edge in the lattice as a false positive event,
or randomly delete an event from the gold events.

The conclusions are consistent with our previous observation, namely,
that hinge is the consistent winner but only by a very small margin, that
log loss is very competitive, that non-convex losses are hard to optimize,
and that overlap cost is better than MPE cost.  As a byproduct, we note
that we could achieve 17.7\% given a phone detector with
any of the following characteristics: up to 50\% phone error rate but
perfect time information, up to 5-frame time perturbations (in standard
deviation) but perfect labels, 1.8 false positives per gold edge, or 20\%
false negatives.

\begin{figure}
\begin{center}
\includegraphics[width=3.1in]{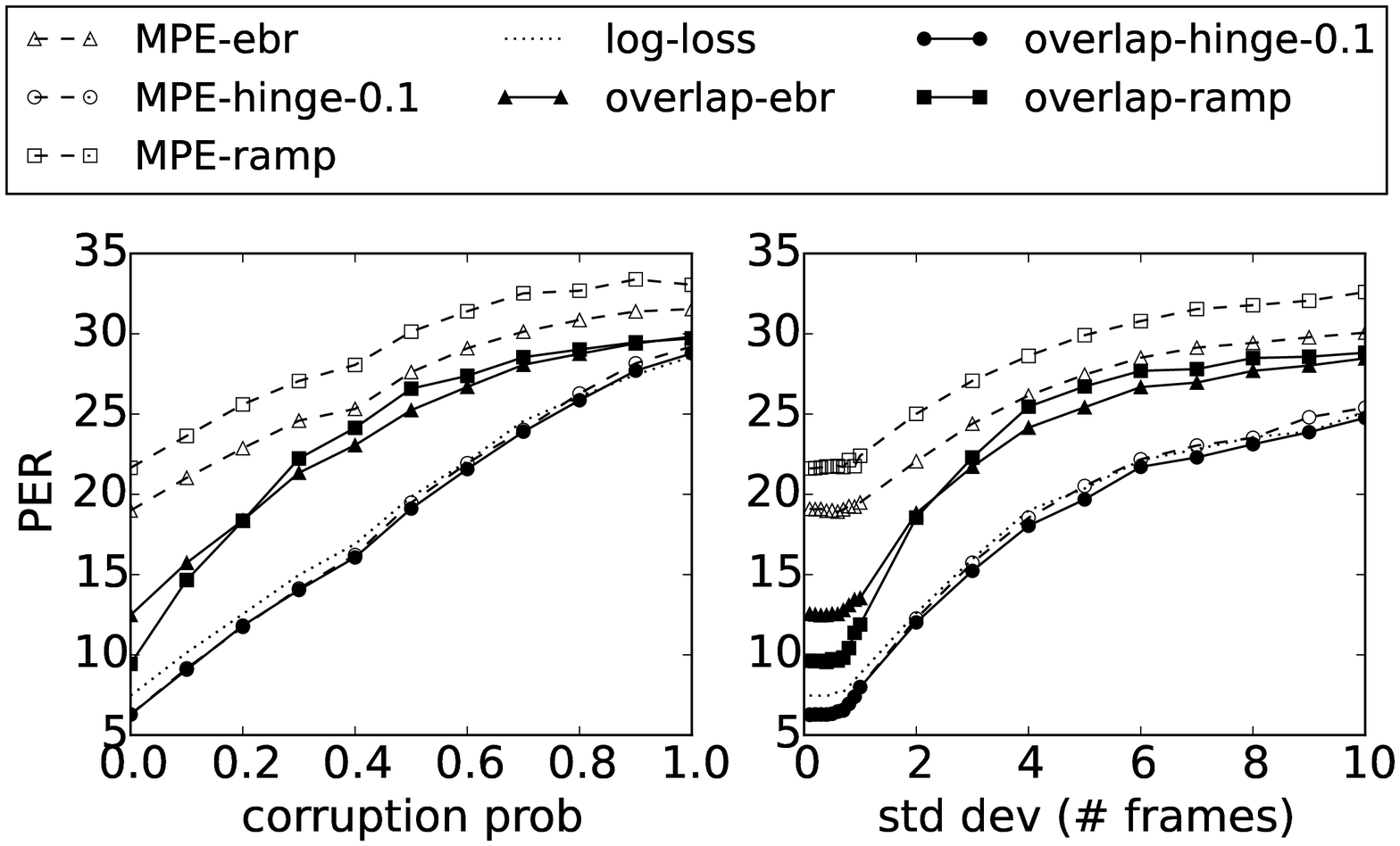}
\includegraphics[width=3.1in]{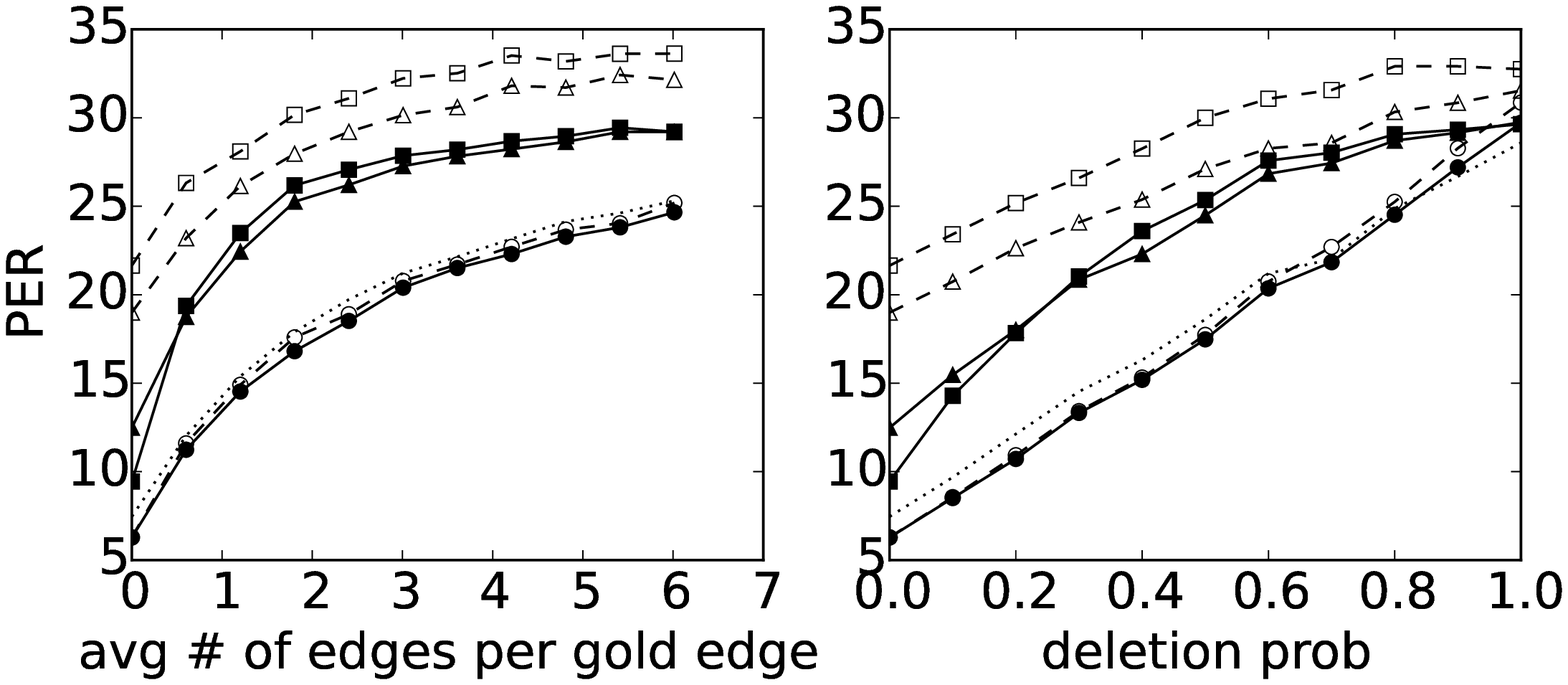}
\caption{\label{fig:gt-noise}
    Lattice rescoring with various noise added to a perfect
    phone detector.
    \emph{From Left to Right}: Perturbing phone labels.  Perturbing time.
    Adding false positive events.  Adding false negative events.
}
\end{center}
\end{figure}

\section{Summary}

In this chapter, we have described a formal framework for discriminative segmental models.
The definition is not tied to any weight function or any loss function,
encompassing a large family of segmental models.
We have described two weight functions, one based on frame classifiers, and one
based on segmental recurrent neural networks.
We have also described various loss functions, and have drawn connections between surrogate
losses and the true loss that we want to minimize.
We have shown how \mbox{(sub-)gradients} of loss functions can
be efficiently computed with FST algorithms.
The flexibility to use different loss functions
allows us to train segmental models in different
training settings.
Preliminary results have shown that segmental models outperform HMMs by a large margin 
and that overlap cost is better than MPE cost across loss functions.
In terms of loss functions, non-convex losses are more difficult to optimize.
Hinge loss is the best performer while log loss is very competitive.

Our final task loss is the edit distance,
but none of the surrogate losses
can be optimized when the cost function is the edit distance.
There are two difficulties for optimizing the edit distance: the edit distance is discrete,
and it involves a minimization over all alignments to the ground truth.
Using a cost function that considers all alignments to the ground truth
without the minimization has been explored in \citep{HMSN2005, VG2015, S2017}.
Other approaches for directly minimizing task losses \citep{KM2011,KK2015}
are also suitable for minimizing the edit distance.

%% file: dsc.tex
\chapter{Discriminative Segmental Cascades}

As we have seen in Chapter \ref{ch:seg}, search spaces
defined by segmental models are dense and redundant,
in the sense that many segments
that have the same label
differ only by a few frames (or even just a single frame)
and that many segments with the same start time and
end time differ only in the labels.
In other words, most of the random segments
do not look like the ground truths.

Decoding in multiple passes is an approach
that exploits dense search spaces.
Since the search spaces are dense,
a simple weight function is typically
enough to significantly reduce the size of search spaces.
After the first pass, we may use more complicated weight functions
to make predictions or to further reduce the search spaces.
Reducing the size of search spaces is
commonly known as \term{pruning}.

Pruning has to be done carefully,
because it might hurt the final performance if the ground truth segments are pruned.
To measure how pruning affects the search spaces and the quality of predictions,
we measure the lowest achievable task loss, the \term{oracle error rate},
of the search spaces after pruning.
A simple experiment on the phonetic recognition task shows that
the oracle error rate, i.e., the best edit distance we can achieve, on average stays at zero
even if we remove 50\% of segments at random.
This result shows that the search spaces are indeed dense and redundant.

Since the pruned search space is smaller than the original space,
inference and learning using the pruned search space are faster
in the second pass.
As a result, we can afford to use computationally
expensive weight functions for better prediction in the second pass.
In general, with the same weight function, multi-pass systems
have the potential to decode faster than single-pass systems.
In other words, under the same decoding time budget, multi-pass systems can use
computationally more expensive weight functions
than single-pass systems to obtain better performance.


Since pruning plays an important role,
in the following sections, we present and compare
several pruning strategies and their consequences.
We then develop a multi-pass system,
trading between the size of
the search space and the computational complexity of
weight functions.

\section{Pruning}

In this section, we describe several pruning approaches
to reduce search spaces.
Note that pruning is a form of inference,
and we assume a pruning approach has access to search spaces
defined by an FST, and a weight function $w : E \to \mathbb{R}$
parameterized by $\Theta_{\text{prn}}$.
We discuss the choice of $w$ and how $\Theta_{\text{prn}}$ is obtained
in the experimental sections.

\subsection{Greedy pruning}

A very naive approach, referred to as
greedy pruning, is to prune the edges branching out
from a vertex based on their edge weights.
Specifically, the set of edges branching out from a vertex $v$ is pruned
based on a threshold $\tau$.  Let
\begin{align}
S_v = \{e \in \outedges(v): w(e) \geq \tau\}
\end{align}
be the set of vertices that survive pruning.
We collect the edges $\bigcup_{v \in V} S_v$ to form
the pruned graph.
There are two nice properties about this approach.
First, the pruning procedure is embarrassingly parallelizable.
Second, if we let
\begin{align}
\tau = \lambda \min_{e \in \outedges(v)} w(e)
    + (1 - \lambda) \max_{e \in \outedges(v)} w(e)
\end{align}
where $0 \leq \lambda \leq 1$ is a parameter that
controls the amount of pruning, the graph is guaranteed to be connected.
To see this, note that
there is at least one edge surviving for every vertex.
In other words, from the start vertex,
there is at least one edge for us to traverse
for every vertex, and we eventually arrive
at the final vertex.

\subsection{Beam pruning}

Beam pruning, or more generally beam search \citep{L1976},
is a widely used search and pruning method.
The motivation behind beam search is to constrain a search algorithm,
such as the shortest-path algorithm (Algorithm \ref{alg:shortest}),
with a fixed memory budget.
Due to the memory budget, we cannot afford to remember
all the paths we have traversed,
so we prune the paths that are less likely to have high weights.

Specifically, the beam search algorithm performs the following steps while
visiting vertices in topological order.
We keep an approximate shortest distance $\hat{d}(u)$ to every
vertex $u$.
Suppose vertex $v$ is the current vertex being visited.
A set of surviving edges
\begin{align}
S_v = \{e \in \inedges(v): \hat{d}(\tail(e)) \geq \tau\}
\end{align}
is computed based on a threshold $\tau$, and then
the approximate shortest distance for $v$ is updated by setting
\begin{align}
\hat{d}(v) = \max_{e \in S_v} w(e) + \hat{d}(\tail(e)).
\end{align}
Note that this equation is identical to the shortest-path recursion,
except that the set of incoming edges $\inedges(v)$ is pruned before updating.
To reconstruct the graph after beam pruning, we simply collect the
surviving edges $\bigcup_{v \in V} S_v$.
We can also let
\begin{align}
\tau = \lambda \min_{e \in \inedges(v)} \hat{d}(\tail(e))
    + (1 - \lambda) \max_{e \in \inedges(v)} \hat{d}(\tail(e))
\end{align}
where $0 \leq \lambda \leq 1$ is a parameter that
controls the amount of pruning.

Note that beam pruning is very similar to greedy pruning.
For greedy pruning, edges are pruned based on edge weights,
while for beam pruning, edges are pruned based on estimates of shortest
distances.

\subsection{Max-marginal pruning}

It is obvious that if an edge is pruned, paths going through the edge
are no longer in the search space.
Whenever we prune an edge, we discard the maximum-weight
path that passes through the edge.
Obviously if the maximum-weight path passing through
the edge survives pruning, the edge survives pruning.
The \term{max-marginal} for an edge is defined as
the weight of the maximum-weight path passing through
the edge.

Max-marginal pruning was first proposed by \cite{SO1999} and later
rediscovered by \cite{WST2012}.
Formally, the max-marginal of an edge $e$ is defined as
\begin{equation}
\gamma(e) = \max_{p \ni e} w(p),
\end{equation}
i.e., the weight of the maximum-weight path that passes through $e$.
The pruning procedure is simple: choose a threshold $\tau$
and discard edge $e$ if $\gamma(e) < \tau$.
One way to set the threshold \citep{WST2012} is
\begin{equation}
\tau_\lambda = \lambda \max_{e \in E} \gamma(e) + (1 - \lambda) \frac{1}{|E|}\sum_{e \in E} \gamma(e),
\end{equation}
where $E$ is the set of segments and $0 \leq \lambda \leq 1$.
There are two nice properties about max-marginal pruning
with the above threshold:
There is at least one path surviving;
all paths with weights higher than $\tau_\lambda$ survive the pruning.
The first property is true because paths that achieve the maximum weight
always survive.  The second property can be proved by contradiction.
Suppose path $p$ has weight $s > \tau_\lambda$ but is pruned.
Then there exists an edge $e$ in $p$ such that $\gamma(e) < \tau_\lambda$.
On the other hand, since $p$ passes through $e$, $\gamma(e)$
is at least $s$, i.e., $\gamma(e) > s > \tau_\lambda$, a contradiction. 
Note that we can only guarantee that paths with weights larger
than $\tau_\lambda$ survive pruning.  It does not imply
that all paths that survive pruning have weights larger than $\tau_\lambda$.

To calculate $\gamma(e)$ for each $e \in E$, note that
\begin{align}
\gamma(e) = \max_{p \ni e} w(p)
    = \Big[ \max_{p_1 \in \mathcal{P}(s, \tail(e))} w(p_1) \Big]
        + w(e)
        + \Big[ \max_{p_2 \in \mathcal{P}(\head(e), t)} w(p_2) \Big],
\end{align}
where $\mathcal{P}(u, v)$ is the set of paths that start at vertex $u$
and end at vertex $v$.
We can construct an algorithm like the shortest-path algorithm
to calculate the first and third terms as follows.
\begin{align}
d_1(v) & = \max_{p_1 \in \mathcal{P}(s, v)} w(p_1)
    = \max_{e \in \inedges(v)} [ d_1(\tail(e)) + w(e) ] \\
d_2(v) & = \max_{p_2 \in \mathcal{P}(v, t)} w(p_2)
    = \max_{e \in \outedges(v)} [ d_2(\head(e)) + w(e) ]
\end{align}
As shown above, the runtime of max-marginal pruning is
the same as that of the shortest-path algorithm (Algorithm \ref{alg:shortest}).

Max-marginal pruning can also be applied to vertices.
Let $\gamma(v) = \max_{p \ni v} w(p)$ be the max-marginal
of the vertex $v$.  We can also obtain a similar recursion
\begin{align}
\gamma(v) = \max_{p \ni v} w(p)
    = \Big[ \max_{p_1 \in \mathcal{P}(s, v)} w(p_1) \Big]
        + \Big[ \max_{p_2 \in \mathcal{P}(v, t)} w(p_2) \Big]
    = d_1(v) + d_2(v),
\end{align}
and set a threshold based on the convex combination
of $\max_{v \in V} \gamma(v)$ and $\frac{1}{|V|} \sum_{v \in V} \gamma(v)$.
Similar guarantees also hold for max-marginal vertex pruning.

\section{Discriminative Segmental Cascades}

Recall that the runtime for finding the shortest path is $O(|E|C)$
where $E$ is the set of segments in the search space and
$C$ is the computational cost of evaluating the weight function
on a single segment.
Suppose we have a fixed time budget.
Since the pruned search space is smaller than the original space,
we can afford to use a more computationally expensive weight
function on the pruned space for better prediction.
Since additional pruning can be applied to the search space,
we end up with a system having multiple rounds of pruning
with increasingly expensive weight functions.
Prediction is then done on the pruned space from the final round.

The above approach is also commonly referred to as \term{rescoring},
because hypotheses are given new scores (weights) after pruning.
Models are also named after the pass in which they are used.
For example, a first-pass model is one that searches
over the entire search space;
a second-pass model is a model that rescores the pruned search space
produced by a first-pass model.
Based on rescoring, \cite{WST2012} proposed
structured prediction cascades, a type of multi-pass system
with multiple rounds of pruning.
The complete search space is denoted $H_1$.
At round $k$, we have hypothesis space $H_k$,
train a pruning model $\Theta_k$ on $H_k$,
and prune $H_k$ to get $H_k'$.
The pruned space $H_k'$ is expanded to $H_{k+1}$ by
considering more context for every segment, such as label $n$-grams.
The process can be repeated for any number of rounds.

\begin{figure}
    \begin{center}
    \begin{tikzpicture}[ver/.style={draw,circle}]
    \node [ver, ultra thick] (x1) at (0, 0) {};
    \node [ver] (x2) at (1, 0) {};
    \node [ver] (x3) at (2, 0) {};
    \node [ver] (x4) at (3, 0) {};
    \node [ver, double] (x5) at (4, 0) {};
    
    \draw[->,black!50] (x1) to [out=10,in=170] (x2);
    \draw[->,black!50] (x1) to [] (x2);
    \draw[->,black!50] (x1) to [out=-10,in=190] (x2);
    
    \draw[->,black!50] (x2) to [out=10,in=170] (x3);
    \draw[->,black!50] (x2) to [] (x3);
    \draw[->,black!50] (x2) to [out=-10,in=190] (x3);
    
    \draw[->,black!50] (x3) to [out=10,in=170] (x4);
    \draw[->,black!50] (x3) to [] (x4);
    \draw[->,black!50] (x3) to [out=-10,in=190] (x4);
    
    \draw[->,black!50] (x4) to [out=10,in=170] (x5);
    \draw[->,black!50] (x4) to [] (x5);
    \draw[->,black!50] (x4) to [out=-10,in=190] (x5);
    
    \draw[->,black!50] (x1) to [out=30,in=150] (x3);
    \draw[->,black!50] (x1) to [out=25,in=155] (x3);
    \draw[->,black!50] (x1) to [out=20,in=160] (x3);
    
    \draw[->,black!50] (x2) to [out=30,in=150] (x4);
    \draw[->,black!50] (x2) to [out=25,in=155] (x4);
    \draw[->,black!50] (x2) to [out=20,in=160] (x4);
    
    \draw[->,black!50] (x3) to [out=30,in=150] (x5);
    \draw[->,black!50] (x3) to [out=25,in=155] (x5);
    \draw[->,black!50] (x3) to [out=20,in=160] (x5);
    
    \draw[->,black!50] (x1) to [out=45,in=135] (x4);
    \draw[->,black!50] (x1) to [out=40,in=140] (x4);
    \draw[->,black!50] (x1) to [out=35,in=145] (x4);
    
    \draw[->,black!50] (x2) to [out=45,in=135] (x5);
    \draw[->,black!50] (x2) to [out=40,in=140] (x5);
    \draw[->,black!50] (x2) to [out=35,in=145] (x5);
    
    \node [ver,dashed] (y0) at (7, 0) {};
    \node [ver, ultra thick] (y1) at (8, 0) {};
    \node [ver] (y2) at (9, 0) {};
    \node [ver] (y3) at (10, 0) {};
    \node [ver] (y4) at (11, 0) {};
    \node [ver, double] (y5) at (12, 0) {};

    \draw[->,dashed,black!50] (y0) to node [black,above] {$\epsilon$} (y1);
    
    \draw[->,black!50] (y1) to [out=20,in=160] node [black,above] {a} (y2);
    \draw[->,black!50] (y1) to [out=-20,in=200] node [black,below] {b} (y2);
    
    \draw[->,black!50] (y4) to [] node [black,below] {c} (y5);
    \draw[->,black!50] (y2) to [out=35,in=145] node [black,above] {b} (y4);
    \draw[->,black!50] (y2) to node [black,below] {a} (y3);
    \draw[->,black!50] (y3) to node [black,below] {a} (y4);
    
    \node [ver, ultra thick] (z1) at (3.5, -3.5) {};
    \node [ver] (z21) at (4.5, -3.5) {};
    \node [ver] (z22) at (4.5, -2.5) {};
    \node [ver] (z3) at (5.5, -3.5) {};
    \node [ver] (z41) at (6.5, -3.5) {};
    \node [ver] (z42) at (6.5, -2.5) {};
    \node [ver, double] (z5) at (7.5, -3.5) {};
    
    \draw[->,black!50] (z1) to node [black,below] {$\epsilon$a} (z21);
    \draw[->,black!50] (z1) to node [black,left] {$\epsilon$b} (z22);
    
    \draw[->,black!50] (z21) to node [black,below] {aa} (z3);
    \draw[->,black!50] (z21) to node [black,right] {ab} (z42);
    \draw[->,black!50] (z22) to node [black,left] {ba} (z3);
    \draw[->,black!50] (z22) to node [black,above] {bb} (z42);
    
    \draw[->,black!50] (z3) to node [black,below] {aa} (z41);
    \draw[->,black!50] (z41) to node [black,below] {ac} (z5);
    \draw[->,black!50] (z42) to node [black,right] {bc} (z5);
    
    \node at (2, -1) {first-pass hypothesis space $H_1$};
    \node at (9, -1) {pruned hypothesis space $H_1'$};
    \node at (5.5, -4.5) {second-pass hypothesis space $H_2 = {H_1'}^\uparrow$};
    
    \end{tikzpicture}
    \end{center}
    \caption{An example segmental cascade. The search space $H_1$
        is the complete space.  The search space $H_1'$
        is pruned from $H_1$.  The search space $H_2$
        is the self-expansion of $H_1'$.
        The additional vertex and edge added to $H_1'$ before
        self-expansion are shown in dashed lines. \label{fig:cascade}}
\end{figure}
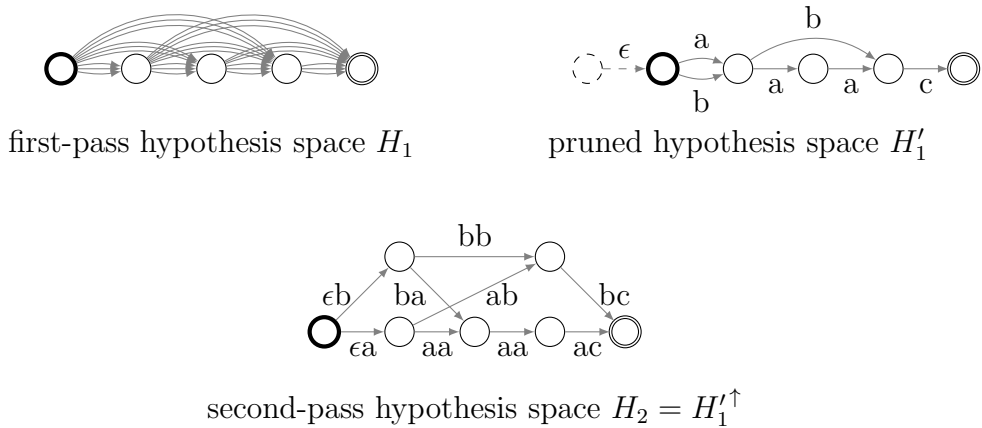

Inspired by \citep{WST2012}, we propose \term{discriminative segmental
cascades}, a multi-pass system consisting of segmental models
with increasingly expensive weight functions.
In our cascades, all search spaces are represented as FSTs.
We denote the complete search space $H_1$.
At the $k$-th pass, based on the search space $H_k$,
we train a segmental model $\Theta_k$, and
use it to prune the search space to get $H_k'$.
In our framework, we can decide whether to expand
the search space to include more context.
Since the search spaces are represented
as FSTs, we propose a new FST operation, termed \term{self-expansion},
to efficiently consider neighboring segments for every segment.
If we decide to self-expand, then $H_{k+1} = {H_k'}^\uparrow$,
where ${H_k'}^\uparrow$ is the self-expansion of $H_k'$.
Otherwise, $H_{k+1} = H_k'$.
Suppose we decide to self-expand.  Then the weight function
for the subsequent pass $w_{k+1}$ can make use of a
wider context than the weight function of the previous pass.
Including a wider context is one possible approach to increase
the expressiveness of the weight function.
Other approaches include using a deeper or more complex neural network,
or extracting more sophisticated features such as tracking formants or pitches,
both of which can be done without self-expansion.
The process can be repeated many times, forming a segmental cascade.
An example of a segmental cascade is shown in Figure \ref{fig:cascade}.

Segmental models in each of these passes can be trained with
the losses described in Section \ref{sec:losses}.
\cite{WST2012} proposed to train the pruning models with
the filtering loss, measuring how well the ground truths are
retained after pruning.  We decide to train the segmental models
in the cascade with losses in Section \ref{sec:losses}
that directly measure the final performance,
making sure that we have a model for prediction after each pass.
This approach also makes it easier to monitor when to stop stacking
the cascade.

\subsection{Self-expansion}
\label{sec:self-expansion}

Given a search space represented as an FST
$H$, we can expand the context of a segment by considering
its neighbors.
Suppose we want to incorporate the previous segment
given the current segment.
We create a new FST $H^\uparrow$ that remembers and
keeps track of the previous segment (edge) traversed in $H$.
Vertices in $H^\uparrow$ correspond to the possible
previous segments in $H$.
Edges in $H^\uparrow$ are
of the form $\langle e_1, e_2 \rangle$,
where $e_1$ is the edge just traversed and $e_2$ is the edge
to be traversed.
We refer to $H^\uparrow$ as the \term{self-expansion} of $H$.
Formally, let $H = ((V, E, \tail, \head), \Sigma, \Lambda, I, F, i, o)$.
The self-expansion of $H$
is defined as $H^\uparrow = ((V^\uparrow, E^\uparrow, \tail^\uparrow, \head^\uparrow),
\Sigma^\uparrow, \Lambda^\uparrow, I^\uparrow, F^\uparrow, i^\uparrow, o^\uparrow)$
where
\begin{align}
V^\uparrow & = E \cup \{ e_0 \} & E^\uparrow & = \{\langle e_1, e_2 \rangle : \head(e_1) = \tail(e_2) \}
\end{align}
\begin{align}
I^\uparrow & = \{ e_0 \} & \tail^\uparrow(\langle e_1, e_2 \rangle) & = e_1 \\
F^\uparrow & = \{ e \in E : \head(e) \in F \} & \head^\uparrow(\langle e_1, e_2 \rangle) & = e_2 \\
\Sigma^\uparrow & = \Sigma & i^\uparrow(\langle e_1, e_2 \rangle) & = i(e_2) \\
\Lambda^\uparrow & = \Lambda & o^\uparrow(\langle e_1, e_2 \rangle) & = o(e_2)
\end{align}
A vertex $v_0$ and an $e_0$ is created in $H$ before self-expansion,
where $i(e_0) = o(e_0) = \epsilon$, $\tail(e_0) = v_0$, and $\head(e_0) = v_1$
assuming we only have a single start state $v_1$, i.e., $I = \{v_1\}$.
The edge $e_0$ is created as a sentinel, corresponding to a vertex in $H^\uparrow$
for not having traversed any edge.

The weight function $w^\uparrow: E^\uparrow \to \mathbb{R}$
is task-dependent, so we do not define it explicitly here.
However, since $w^\uparrow$ is defined on $E^\uparrow$,
the weight function after taking a edge $\langle e_1, e_2 \rangle \in E^\uparrow$
can make use of all the information about $e_1$ and $e_2$ to compute
the weight, including the start times, end times, and labels.
We also observe that
\begin{align}
\outedges^\uparrow(e) & = \{ \langle e, e' \rangle : e' \in \outedges(\head(e)) \} \\
\inedges^\uparrow(e) & = \{ \langle e', e \rangle : e' \in \inedges(\tail(e)) \}
\end{align}
for $e \in V^\uparrow = E$.  Therefore, similarly to $\sigma$-composition,
self-expansion can be computed lazily (on the fly) while traversing $H$.

\section{Experiments}
\label{sec:dsc-exp}

We conduct experiments on the TIMIT data set
in the same setting as in Section \ref{sec:seg-result},
except that here we follow \citep{GMH2013} and use 40-dimensional log mel filter bank outputs
instead of 39-dimensional MFCCs.
The data set is phonetically transcribed, so we have the option of training
the encoders with frame-wise cross entropy based on
the ground-truth frame labels.

\subsection{A segmental baseline}

In the following experiments, we explore
convolutional neural network (CNN) encoders.
The CNN encoders are inspired by \citep{SZ2015} in that they are
deep convolutional neural networks with small $3 \times 3$ and $5 \times 5$ filters.
For frame classification, the input to the network is a window of
15 frames of 40-dimensional log mel
filter outputs centered on the current frame. The network has five convolutional layers,
with 64 filters of size $5 \times 5$ for the first layer and 128, 128, 256, and 256 filters
of size $3 \times 3$ for layers two to five respectively,
each of which is followed by a rectified linear unit
(ReLU) activation \citep{NH2010}, with max pooling layers after the first
and the third ReLU layers.
The output of the final ReLU layer
is concatenated with a window of 15 frames of 39-dimensional MFCCs centered 
on the current frame, and the resulting vector is passed
through three fully connected ReLU layers with 4096 units each.
The network is trained with SGD for 35 epochs with
step size 0.001, momentum 0.95, weight decay 0.005, and a
mini-batch size of 100 frame predictions.
Dropout \citep{S+2014} are applied to the concatenation layer with rate 20\%
and to the fully connected layers with rate 50\%.
This classifier was tuned on the development
set and achieves a 22.3\% frame error rate (after collapsing
to 39 phone labels) on the development set
and 23.0\% on the test set.

After confirming that the CNN encoder is able to perform
well on frame classification, we use the CNN encoder to generate
frame posterior probabilities,
and train segmental models with the FC weight function
and a maximum duration of 30 frames.
Hinge loss is optimized with AdaGrad \citep{DHS2011} with step size 0.01
and a mini-batch size of 1 utterance for 70 epochs.
No explicit regularizer is used except early stopping
on the development set.
We use a scaled overlap cost
for our experiments:
\begin{align}
\cost(p', p) &= \sum_{e' \in p} \cost(e', p) \\
\cost(e', p) &= |e' \cup \tilde{e}| - |e' \cap \tilde{e}| \ind_{o(e') = o(\tilde{e})}
\end{align}
where $p'$ is the hypothesized path, $p$ is the ground-truth path,
$|e' \cup e|$ denotes the union of the two intervals
defined by $e$ and $e'$,
$|e' \cap e|$ denotes the intersection of the two
intervals, $o(e)$ is the output label of $e$,
and $\tilde{e} = \argmax_{e \in p} |e' \cap e|$.
In words, $\tilde{e}$ is the edge that overlaps the most with $e'$.
The cost is non-negative and is only zero if $e' = e$,
and it can be seen as an estimate of the number of incorrectly predicted frames.

The results are shown in Table \ref{tbl:1st-pass}.
Our first-pass segmental model is on par with
\cite{ADYJ2013}, where a CNN encoder is also used,
and it is also on par with a baseline HMM-DNN hybrid
system that we trained with Kaldi using the standard recipe for TIMIT \citep{P2011}.

\begin{table}
\begin{center}
\caption{Phoneme error rates (\%) on TIMIT
    comparing a HMM-DNN hybrid system to
    discriminative segmental models.}
\label{tbl:1st-pass}
\renewcommand{\arraystretch}{1.5}
\begin{tabular}{p{6cm}ll}
              & dev      & test \\
\hline
HMM-DNN       &          & 21.4 \\
\hline
deep segmental neural network \newline
\citep{ADYJ2013}
              &          & 21.8 \\
our segmental model
              & 22.2     & 21.7
\end{tabular}
\end{center}
\end{table}

\subsection{Pruning comparison}

To construct a second-pass segmental model,
we first evaluate pruning approaches based
on oracle error rates and lattice densities.
The \term{oracle error rate} is the lowest achievable
phone error rate of a given lattice.
The \term{lattice density} is defined as the total number of
edges divided by the number of ground-truth edges,
i.e., the number of hypothesized edges per ground-truth edge.
In general, a dense lattice has a higher chance to contain
the ground-truth path than a sparse lattice.
In fact, the entire search space is a dense lattice
that always contains the ground-truth path.
A good pruner is expected to produce sparse lattices
while maintaining a low oracle error rate.

\begin{figure}
\begin{center}
\includegraphics[height=6cm]{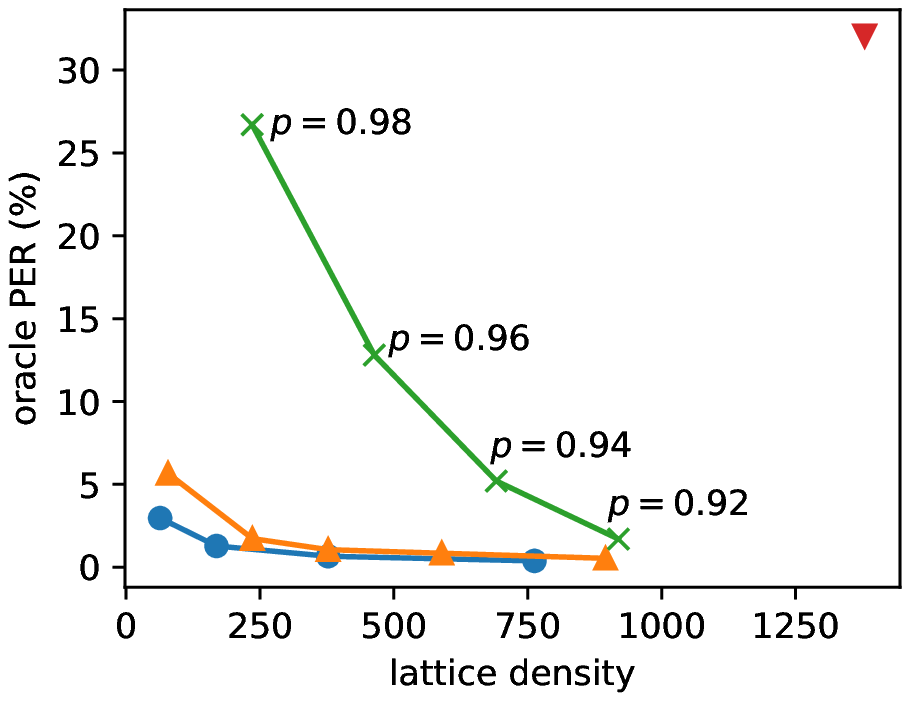}
\includegraphics[height=6cm]{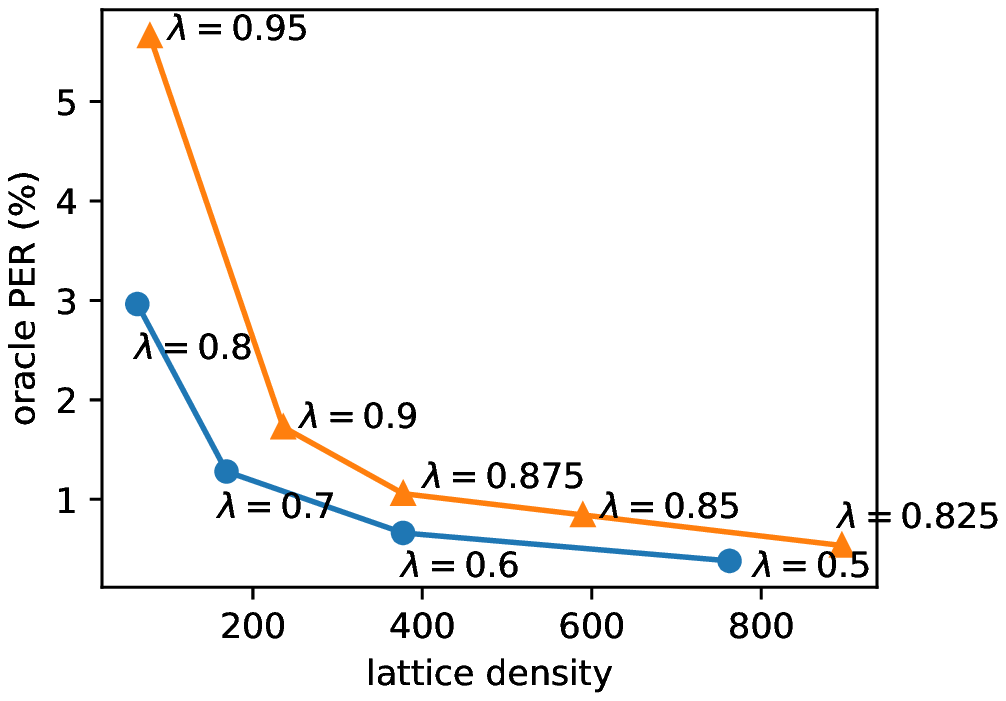}

\begin{tikzpicture}[font=\strut]
\draw[color=c0,thick] (0.0, 0) -- (0.6, 0);
\path[fill=c0] (0.3, 0) circle (0.1);
\node at (0.6, 0) [right] {max-marginal pruning};
\end{tikzpicture}
\begin{tikzpicture}[font=\strut]
\draw[color=c1,thick] (0.0, 0) -- (0.6, 0);
\fill[color=c1,thick] (0.3, 0.16) -- (0.164, -0.08) -- (0.436, -0.08) -- cycle;
\node at (0.6, 0) [right] {greedy pruning};
\end{tikzpicture}
\begin{tikzpicture}[font=\strut]
\draw[color=c2,thick] (0.0, 0) -- (0.6, 0);
\draw[color=c2,thick] (0.2, 0.1) -- (0.4, -0.1);
\draw[color=c2,thick] (0.2, -0.1) -- (0.4, 0.1);
\node at (0.6, 0) [right] {random pruning};
\end{tikzpicture}
\begin{tikzpicture}[font=\strut]
\draw[color=c3,thick] (0.0, 0) -- (0.6, 0);
\fill[color=c3,thick] (0.3, -0.16) -- (0.164, 0.08) -- (0.436, 0.08) -- cycle;
\node at (0.6, 0) [right] {beam pruning};
\end{tikzpicture}
\end{center}
\caption{\emph{Left}: Oracle error rates vs lattice densities for various pruning
    approaches. \emph{Right}: A re-scaled version of the left figure focusing on
    greedy pruning and max-marginal pruning.}
\label{fig:cnn-max-marginal-pruning}
\end{figure}

We compare max-marginal pruning with $\lambda \in \{0.5, 0.6, 0.7, 0.8\}$,
greedy pruning with $\lambda \in \{0.95, 0.9, 0.875, 0.85, 0.825\}$,
random pruning with probability $\{0.92, 0.94, 0.96, 0.98\}$,
and beam pruning with $\lambda=0.9$.
The results are shown in Figure \ref{fig:cnn-max-marginal-pruning}.
For reference, the average density of the complete spaces is 11587.67.
We observe that search spaces of segmental models are robust to
missing edges---the oracle error rate is only 1.7\% when we randomly drop
92\% of the edges.
We also observe that max-marginal pruning
produces significantly sparser lattices than greedy pruning and random pruning
while maintaining low oracle error rates.
Beam pruning performs the worst.  Since greedy pruning works well
by just comparing edges with the same tail vertex (i.e., with the same start time),
we suspect the bad performance
of beam pruning is due to comparing distances at different time points.

\subsection{Improving prediction}

Based on the above pruning comparison, we decide
to use the lattices generated by max-marginal pruning
with $\lambda = 0.8$.
Let $H_1$ be the entire search space, and $H_1'$
be the pruned search space.
To train a segmental model with a wider segmental context,
we first self-expand $H_1'$ into $H_2 = {H_1'}^\uparrow$
and introduce the following weight functions.
Note that the weight functions can now depend on two
segments instead of one.

\paragraph{lattice weight}
Instead of re-learning all of the weight functions
in the first-pass model, we reuse them in the second pass
by having
\begin{align}
w_{\text{lat}}(\langle (\ell_1, s_1, t_1), (\ell_2, s_2, t_2) \rangle)
    = \alpha \cdot w((\ell_2, s_2, t_2))
\end{align}
from the first level of the cascade, where
$\alpha$ is a learnable parameter.
\paragraph{bigram LM weight} We use the weight
\begin{align}
w_{\text{LM}}(\langle (\ell_1, s_1, t_1), (\ell_2, s_2, t_2) \rangle)
    = \beta \cdot \log p(\ell_2 | \ell_1)
\end{align}
to include the log probability of the bigram $\ell_1 \ell_2$,
where $\beta$ is a learnable parameter.
\paragraph{second-order boundary weight} We define
\begin{align}
w_{\text{left}-k}(\langle (\ell_1, s_1, t_1), (\ell_2, s_2, t_2) \rangle)
    & = u_{s_2 - k, \ell_1, \ell_2} \\
w_{\text{right}-k}(\langle (\ell_1, s_1, t_1), (\ell_2, s_2, t_2) \rangle)
    & = u_{t_2 + k, \ell_1, \ell_2}'
\end{align}
similarly to the first-order boundary weights,
where $u_{t, \ell, \ell'} = \theta_{\ell, \ell'}^\top h_t$ and
$u_{t, \ell, \ell'}' = {\theta'_{\ell, \ell'}}^\top h_t$
for $t = 1, \dots, \tilde{T}$
and some learnable parameters $\theta, \theta'$.

\paragraph{} \hspace{0pt}
Since the search space is sparse, we can afford to compute
segment features based on a neural network for every segment.
We choose a similar CNN architecture to the one for frame
classification.  We pre-train the CNN with a segment classification task.
Here the features at the input layer are the log mel filter outputs
from a 15-frame window around the segment's central frame.
Instead of concatenation with 15-frame MFCCs,
we concatenate with a segmental feature vector consisting of
the average MFCCs of three sub-segments
in the ratio of 3-4-3, plus two four-frame averages at both
boundaries and length indicators for length 0 to 20 (similar
to the segmental feature vectors of \citep{HG1998,CM1999}).  This CNN is
trained on the ground-truth segments in the training set.
Finally, we build an ensemble of such networks with different
random seeds and a majority vote.  This ensemble classifier
has a 15.0\% segment classification error on the test set.
A comparison of our CNN
to other segment classifiers is shown in Table \ref{tbl:segment-classification}.

\begin{table}
\begin{center}
\caption{TIMIT segment classification error rates (\%).}
\label{tbl:segment-classification} 
\renewcommand{\arraystretch}{1.5}
\begin{tabular}{p{8cm}l}
             & test \\
\hline
Gaussian mixture model (GMM) \newline
\citep{CM1999} 
             & 26.3 \\
SVM \newline
\citep{CM1999}
             & 22.4 \\

Hierarchical GMM \newline
\citep{H1998}
             & 21.0 \\
Discriminative hierarchical GMM \newline
\citep{CG2007}
             & 16.8 \\
SVM with deep scattering spectrum \newline
\citep{AM2014}
             & 15.9 \\
\hline
our CNN ensemble \newline
\citep{TWGL2015}
             & 15.0
\end{tabular}
\end{center}
\end{table}

Directly running the CNN segment classifier
for every edge in the lattice is, however,
still too time-consuming.   We instead
compress the best-performing (single) CNN into a shallow
fully connected network with one hidden layer of 512 ReLUs by training
it to predict the log probability outputs of the deep network,
as proposed by \cite{BC2014}.  We then use the log probability outputs
of the shallow network.  We refer to the result as segment NN weights.

\paragraph{segment NN weight} Let
\begin{align}
w_{\text{seg}}(\langle (\ell_1, s_1, t_1), (\ell_2, s_2, t_2) \rangle)
    = (u_{s_2, t_2})_{\ell_2}
\end{align}
where $u_{s, t} = \theta^\top z_{s, t}$, $z_{s, t}$ is a log probability
vector of the labels obtained by passing the segment $x_s, \dots, x_t$
to the CNN segment classifier, and
$\theta$ is some learnable parameter vector.

\paragraph{} \hspace{0pt}
To add more context information, we use the same CNN architecture
and training setup to learn a bi-phone frame classifier,
but with an added 256-unit bottleneck linear layer before
the softmax.  Each frame is labeled with its segment label
and one additional label from a neighboring segment.  If the
current frame is in the first half of the segment, the additional
label is the previous phone; if it is in the second half, then
the additional label is the next phone.  The learned bottleneck
layer outputs are used to define weights (although they do not
correspond to log probabilities) with averaging and sampling
as for the uni-phone case. We refer to the resulting weights as
bi-phone NN bottleneck weights.

We use the sum of the lattice weight, the
bigram LM weight, second-order boundary weights,
segment NN weights, bi-phone NN bottleneck weights,
length indicators, and lexicalized bias as our final weight function for the
second level of the cascade. Hinge loss is minimized with
AdaGrad for 20 epochs with step size
0.01.  Again, no explicit regularizer is used except early
stopping based on the PER on the development set.
Results with these additional weights are shown in Table \ref{tbl:2nd-pass}.
Adding the second-order weights, bigram LM, and the
above NN weights gives a 1.8\% absolute improvement over
our best first-pass system, demonstrating the value of
including such expensive features with wider context.

\begin{table}
\begin{center}
\caption{Phoneme error rates (\%)
    with discriminative segmental cascades.
    The search space for the first pass is denoted $H_1$,
    and the search space for the second pass ${H_1'}^\uparrow$
    is denoted $H_2$, where $H_1$ is pruned to get $H_1'$.}
\label{tbl:2nd-pass}
\begin{tabular}{lll}
              & dev    & test    \\
\hline
1$^{\text{st}}$ pass ($H_1$)
              & 22.2   & 21.7    \\
\hline
2$^{\text{nd}}$ pass ($H_2$)     \\
+ bigram LM
              & 19.8   &         \\
+ 2nd-order boundary features
              & 19.2   &         \\
+ segment NN 
              & 18.9   &         \\
+ bi-phone NN bottleneck
              & 18.8   & 19.9
\end{tabular}
\end{center}
\end{table}

\section{Improving Decoding and Training Speed}
\label{sec:fast-dsc}

We have shown that segmental cascades can be used
to incorporate computationally expensive weight functions
while maintaining efficiency by shifting
the computationally expensive weights to later passes.
In this section, instead of incorporating additional weights,
we shift weights to later passes to improve decoding speed.
Due to their recent success \citep{GMH2013},
we also explore long short-term memory (LSTM) networks \citep{HS1997}
as encoders instead of CNN encoders.

\subsection{LSTM Encoders}

In this section, we describe
long short-term memory (LSTM) networks.
We start by defining a single-layer LSTM.
Suppose the input vectors are $x_1, \dots, x_T$.
First we apply a linear transformation to the inputs to get
\begin{align}
x_t' = \begin{pmatrix} W_{xu} \\ W_{xi} \\ W_{xf} \\ W_{xo} \end{pmatrix} x_t,
\end{align}
which can be done for $t=1, \dots, T$ in parallel.
Next, assume $h_0 = c_0 = 0$.
Suppose we have already computed $h_t$ and $c_t$,
and we want to compute $h_{t+1}$ and $c_{t+1}$.
We apply a linear transformation to get the candidates
$u'$, $i'$, $f'$, and $o'$
for the cell update vectors $u$ and the gates $i$, $f$ and $o$.
\begin{align}
\begin{pmatrix}
u' \\ i' \\ f' \\ o'
\end{pmatrix}
= x_t' + \begin{pmatrix} W_{hu} \\ W_{hi} \\ W_{hf} \\ W_{ho} \end{pmatrix} h_t
    + \begin{pmatrix} 0 \\ W_{ci} \\ W_{cf} \\ 0 \end{pmatrix} c_t.
\end{align}
The update vectors $u$, the input gate $i$, and the forget gate $f$
are then computed from the candidates by applying nonlinear transformations.
\begin{align}
u & = \tanh(u') \\
i & = \text{logistic}(i') \\
f & = \text{logistic}(f')
\end{align}
The new cell vector $c_{t+1}$ is computed with the update vector gated by the input gate
and the previous cell.
\begin{align}
c_{t+1} & = i \odot u + f \odot c_t
\end{align}
Finally, the output gate $o$ is computed from the updated cell $c_{t+1}$,
and the new hidden vector $h_{t+1}$ is updated with the cell gated
by the output gate.
\begin{align}
o & = \text{logistic}(o' + W_{co} c_{t+1}) \\
h_{t+1} & = o \odot \tanh(c_{t+1})
\end{align}
The final results after running an LSTM on $x_1, \dots, x_T$
are $h_1, \dots, h_T$ and $c_1, \dots, c_T$.
We use $h_{1:T} = \text{LSTM}(x_{1:T})$ to
denote the process above.

Since LSTMs have an assigned direction, it is natural to
run two LSTMs in opposite directions and combine the hidden vectors.
Formally,
\begin{align}
h^f_{1:T} & = \text{LSTM}(x_{1:T}) \\
h^b_{T:1} & = \text{LSTM}(x_{T:1}) \\
h_t & = W_f h^f_t + W_b h^b_t
\end{align}
We use $h_{1:T} = \text{BiLSTM}(x_{1:T})$
as a shorthand for the bidirectional LSTMs.
Note that the parameters of the two LSTMs are
not shared.
Another way to increase the complexity of the acoustic encoder
is to stack them on top of each other.
For example, if we stack 3 layers of bidirectional LSTMs,
we get
\begin{align}
h^{(1)}_{1:T} &= \text{BLSTM}(x_{1:T}) \\
h^{(2)}_{1:T} &= \text{BLSTM}(h^{(1)}_{1:T}) \\
h^{(3)}_{1:T} &= \text{BLSTM}(h^{(2)}_{1:T})
\end{align}
We use $h_{1:T} = \text{DBLSTM}^3(x_{1:T})$ to
denote the 3-layer bidirectional LSTMs
(where ``D'' refers to ``deep'').
Again the parameters of the LSTMs
in each layer and each direction
are not shared.

\subsection{Experiments}

As in Section \ref{sec:dsc-exp},
we first explore LSTMs on a frame classification task,
Specifically, the log probability vector of a frame at time $t$
is $z_t = \text{logsoftmax}(W h_t + b)$
where $h_t$ is the hidden vector at time $t$ produced by an LSTM,
and $W$ and $b$ are the parameters.
We then explore segmental cascades with 
weight functions based on log probabilities
produced by LSTMs.
Experiments are conducted on the TIMIT data set.
We compute 41-dimensional log mel filter bank outputs
including energy, and concatenate with the delta's and double delta's
to form the final 123-dimensional acoustic feature vectors.
Instead of using a 192-utterance development set,
we follow the Kaldi recipe \citep{P2011} and use 400 utterances
from the complete test set (disjoint from the core test set)
as the validation set.  We report numbers on the core test set,
the same test set as in Section \ref{sec:seg-result}.

\subsubsection{LSTM frame classification}

We build a
frame classifier by stacking three layers of bidirectional LSTMs. The
cell and hidden vectors have 256 units. We train the frame
classifier with frame-wise cross entropy and optimize with AdaGrad \citep{DHS2011}
with mini-batch size of 1 utterance and step size 0.01 for 30 epochs. We
choose the best-performing model on the development set (early
stopping).

\begin{figure}
\begin{center}
\includegraphics[width=7cm]{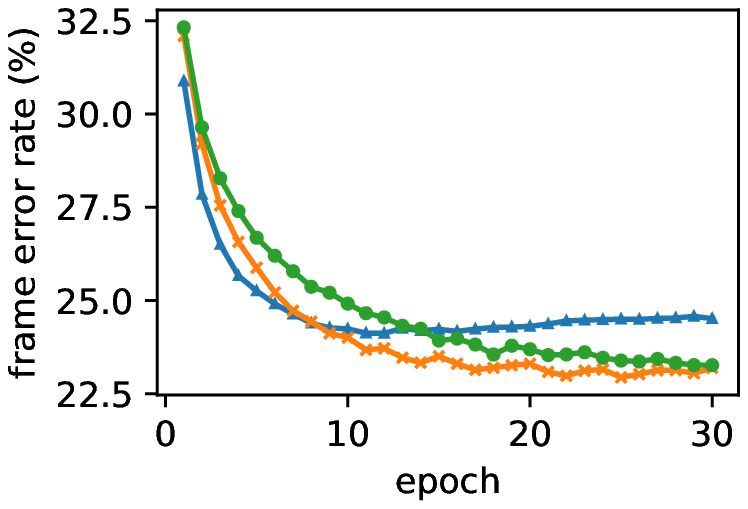}
\includegraphics[width=7cm]{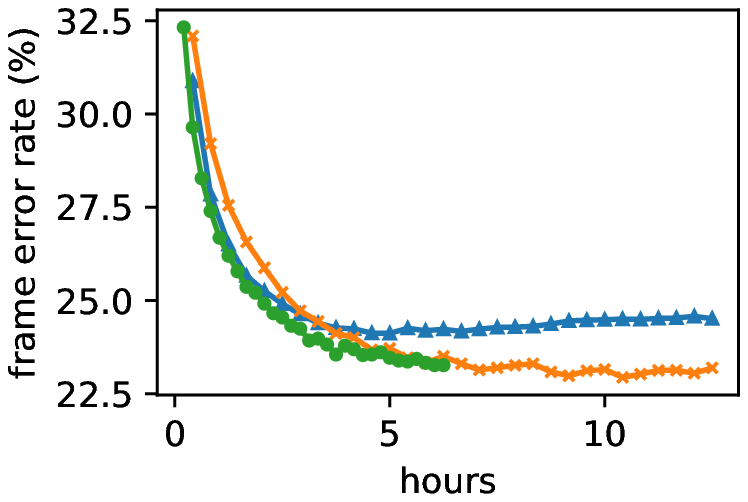}

\begin{tikzpicture}[font=\strut]
\node [right] at (0.2, 0) {vanilla};
\draw[blue] (-0.2, 0) -- (0, 0);
\draw[blue] (0, 0) -- (0.2, 0);
\path[fill=c0] (-0.07, -0.04) -- (0.07, -0.04)
    -- (0, 0.08) -- cycle;
\end{tikzpicture}
\begin{tikzpicture}[font=\strut]
\node [right] at (0.2, 0) {dropout};
\draw[c1] (-0.2, 0) -- (0, 0);
\draw[c1] (0, 0) -- (0.2, 0);
\draw[c1] (-0.07, 0.07) -- (0.07, -0.07);
\draw[c1] (-0.07, -0.07) -- (0.07, 0.07);
\end{tikzpicture}
\begin{tikzpicture}[font=\strut]
\node [right] at (0.2, 0) {subsampling+dropout};
\draw[c2] (-0.2, 0) -- (0, 0);
\draw[c2] (0, 0) -- (0.2, 0);
\path[fill=c2,radius=0.06] (0, 0) circle;
\end{tikzpicture}
\caption{Development set frame error rates vs.\ number of epochs (left)
    and vs.\ training hours (right).}
\label{fig:lstm-frame}
\end{center}
\end{figure}

Following \citep{VDH2013, MGM2015}, we consider dropping half of the
frames for any given utterance in order to save time on feeding
forward.
Specifically, we only use $x_2, x_4, \dots, x_{T - 2}, x_T$
to feed forward through the deep LSTM and generate $z_2, z_4, \dots,
z_{T - 2}, z_T$ (assuming $T$ is even, without loss of generality).
We then copy each even-indexed output to its previous frame, i.e., $z_{i-1} = z_i$
for $i = 2, 4, \dots, T-2, T$.
During training, the cross entropy is calculated over all frames
and propagated back.  Specifically,
let $E_i$ be the cross entropy at frame $i$, and $E = \sum_{i=1}^T E_i$.
The gradient of $z_i$ is the sum of the gradients from the current frame
and the copied frame.
Dropping even-indexed frames is similar to dropping odd-indexed frames,
except the outputs are copied from $z_i$ to $z_{i+1}$ for $i = 1, 3, \dots, T-1$.
When training LSTMs with subsampling, we alternate
between dropping even- and odd-numbered frames every other epoch.
The results are shown in Figure~\ref{fig:lstm-frame}.

We observe that with subsampling the model converges
more slowly than without, in terms of number of epochs.
However, by the end of epoch 30, there is almost no loss
in frame error rates when we drop half of the frames.
Considering the more important measure of training time rather than number of epochs,
the LSTMs with frame subsampling
converge twice as fast as those without subsampling.
For the remaining experiments, we use the log posteriors at each frame
of the subsampled LSTM outputs as the
inputs to
the segmental models.

\subsubsection{Segmental cascade experiments}

Our baseline system, denoted $R$, is a
first-pass segmental model with the FC weight function (Section \ref{sec:fc-weight}).
The baseline system is trained by optimizing hinge loss 
with AdaGrad using mini-batch size 1 utterance, step size 0.1,
and early stopping for 50 epochs.

We decide to shift the FC weight function to the second pass,
and train a segmental model, denoted $A_1$,
with just the label posterior and a label-independent bias.
Specifically, the label posterior weight is defined as
\begin{align}
w((\ell, s, t)) = \sum_{i=s}^t (z_i)_\ell,
\end{align}
where $z_i = \text{logsoftmax}(W h_i + b)$ is the log probability vector
at time $i$ based on the vector $h_i$ produced by the LSTM.
This reduces the number of features from 24528 to two.
We use hinge loss optimized with AdaGrad with mini-batch size 1 and step size 1.
Since we only have two features, learning
converges very quickly, in only three epochs.  We take the model from
the third epoch to produce lattices for subsequent passes in the 
cascade.

Lattices are generated with max-marginal pruning
with $\lambda = 0.85$.  The resulting lattices
have an average oracle error rates of 1.4\% and
average density of 213.02.
We train a second-pass
model, denoted $A_2$, with the FC weight function
except that we add a ``lattice score'' feature corresponding to
the segment score given by the two-feature system. 
Hinge loss is optimized with AdaGrad with mini-batch size 1, step size 0.1,
and early stopping for 20 epochs.

\begin{figure}
\begin{center}
\begin{minipage}{8.5cm}
\includegraphics[width=8cm]{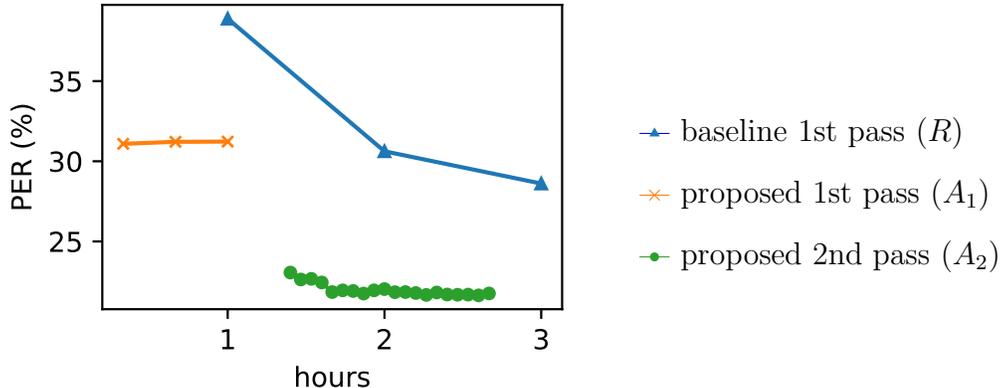}
\end{minipage}
\begin{minipage}{5cm}
\begin{tikzpicture}[font=\strut]
\node [right] at (0.2, 0) {baseline 1st pass ($R$)};
\draw[blue] (-0.2, 0) -- (0, 0);
\draw[blue] (0, 0) -- (0.2, 0);
\path[fill=c0] (-0.07, -0.04) -- (0.07, -0.04)
    -- (0, 0.08) -- cycle;
\end{tikzpicture}
\begin{tikzpicture}[font=\strut]
\node [right] at (0.2, 0) {proposed 1st pass ($A_1$)};
\draw[c1] (-0.2, 0) -- (0, 0);
\draw[c1] (0, 0) -- (0.2, 0);
\draw[c1] (-0.07, 0.07) -- (0.07, -0.07);
\draw[c1] (-0.07, -0.07) -- (0.07, 0.07);
\end{tikzpicture}
\begin{tikzpicture}[font=\strut]
\node [right] at (0.2, 0) {proposed 2nd pass ($A_2$)};
\draw[c2] (-0.2, 0) -- (0, 0);
\draw[c2] (0, 0) -- (0.2, 0);
\path[fill=c2,radius=0.06] (0, 0) circle;
\end{tikzpicture}
\end{minipage}
\caption{Learning curve of the proposed two-pass system compared with
    the baseline system.
    The time gap between the first pass and the second pass is the
    time spent on pruning.
}
\label{fig:fast-dsc-learning}
\end{center}
\end{figure}

The learning curve comparing $R$ and $A_1$ followed by $A_2$
is shown in Figure~\ref{fig:fast-dsc-learning}.
We observe that the learning time per epoch of the two-feature system $A_1$
is only one-third of the baseline system $R$.
We also observe that training of $A_2$ converges faster
than training $R$, despite the fact that they use almost identical
feature functions.
The baseline system achieves the best result at epoch 49.
In contrast, the two-pass system is done
before the baseline even finishes the third epoch.

Following Section \ref{sec:dsc-exp}, given the first-pass baseline,
we apply max-marginal pruning to produce lattices for the second-pass
baseline with $\lambda=0.8$.  The second-pass baseline features
are the lattice score from the first-pass baseline, a bigram
language model score, first-order length indicators,
and a bias.
Hinge loss is optimized with AdaGrad with mini batch size 1,
step size 0.01, and early stopping for 20 epochs.
For the proposed system, we produce lattices with
max-marginal pruning and $\lambda=0.3$ for the third-pass system.
We use the same set of features and hyper-parameters
as the second-pass baseline for the third pass.

Phone error rates of all
passes are shown in Table~\ref{tbl:fast-dsc-per}.
First, if we compare the one-pass baseline
with the proposed two-pass system,
our system is close to the baseline.
Second, we observe a healthy improvement by just adding
the bigram language model score to the second-pass baseline. 
The improvement for our third-pass system is small
but brings our final performance to within 0.4\% of the baseline second pass.

\begin{table}
\begin{center}
\caption{Phone error rates (\%) of proposed and baseline systems.}
\label{tbl:fast-dsc-per}
\begin{tabular}{ll|lll}
         &       & 1st pass & 2nd pass & 3rd pass \\
\hline
baseline & dev   & 21.9     & 21.0     &      \\
         & test  & 24.0     & 23.0     &      \\
\hline
proposed & dev   & 33.6     & 21.5     & 21.3 \\
         & test  &          & 23.7     & 23.4 \\
\end{tabular}
\end{center}
\end{table}

Next we report on the speedups in training
and decoding obtained with our proposed approach.
Table~\ref{tbl:fast-dsc-decode} shows the real-time factors
for decoding with the baseline and proposed systems.
In terms of decoding time alone, we achieve
a factor of 2.4 speedup compared to the baseline.
If the time of feeding forward LSTMs is included, then
our proposed system is two times faster
than the baseline.

\begin{table}
\begin{center}
\caption{Real-time factors for decoding.}
\label{tbl:fast-dsc-decode}
\begin{tabular}{l|llll|ll}
           & 1st pass & 2nd pass & 3rd pass & total decoding & feeding forward & total overall \\
\hline
baseline   & 0.33     & 0.01     &          & 0.34           & 0.33            & 0.67          \\
proposed   & 0.11     & 0.02     & 0.01     & 0.14           & 0.17            & 0.31
\end{tabular}
\end{center}
\end{table}

Table~\ref{tbl:fast-dsc-train} shows the times needed
to train a system to get to the performance in Table~\ref{tbl:fast-dsc-per}.
The speedup mostly comes from the fast convergence
of the first pass.
In terms of training the segmental models alone,
we achieve an 18.0-fold speedup.
If the time to train the LSTMs is included, then
we obtain a 3.4-fold speedup
compared to the baseline.

To summarize some of the above results:  With a combination of the first-pass two-feature
system and edge pruning, we prune 95\% of
the segments in the first-pass hypothesis space,
leading to significant speedup in both decoding and training.
The feed-forward time for our LSTMs is halved through frame subsampling.
In the end, with a single four-core CPU, we achieve
0.31 times real-time decoding
including feeding forward,
which is 2.2 times faster than the baseline, 
and 32.5 hours in total to obtain our final model
including LSTM training, which is
3.4 times faster than the baseline.
Excluding the LSTMs, the segmental model decoding alone is 2.4 times faster than the baseline,
and training the segmental models alone is 18 times faster than the baseline.

\begin{table}
\begin{center}
\caption{Hours for training the system.}
\label{tbl:fast-dsc-train}
\begin{tabular}{l|llll|ll}
           & 1st pass & 2nd pass & 3rd pass & total decoding & feeding forward & total overall \\
\hline
baseline   & 49.5     & 0.6      &          & 50.1           & 59.4            & 109.5  \\
proposed   & 1.0      & 1.2      & 0.6      & 2.8            & 29.7            & 32.5
\end{tabular}
\end{center}
\end{table}

\section{Summary}

In this chapter, we have described discriminative segmental cascades,
a multi-pass system for segmental models.
Max-marginal pruning lies at the heart of segmental cascades,
producing sparse lattices while having low oracle error rates.
After pruning, it becomes feasible to
incorporate computationally expensive weight functions,
such as higher-order LMs, and weight functions based on
neural networks.
We demonstrate that incorporating these weights significantly
improves the performance of segmental models.
We also show that decoding and training speed can be significantly improved
with segmental cascades by shifting weights to later
passes without sacrificing accuracy.

Though we have shown improvement with segmental cascades,
how the density and structure of search spaces affect
the training of segmental models is still not fully known.
Task-dependent priors, such as phoneme duration,
are not fully explored and can potentially be integrated
into the pruning procedure.
The fact that beam search works well for frame-based
models but fails miserably for segmental models is
also worth investigating.
Converting other more sophisticated search algorithms,
such as $\text{A}^*$ search \citep{KHGLMO1993},
into pruning algorithms is a possibility that
might show us signs as to why beam search
fails for segmental models.

%% file: e2e.tex
\chapter{End-to-End Training Approaches}
\label{ch:e2e}

In the previous chapter, we have seen how a sequence
recognizer can be trained in multiple stages:
a frame classifier is first trained with frame-wise
cross entropy, and a segmental model is trained with a sequence loss,
such as the ones defined in Section \ref{sec:losses},
based on the classifier's outputs.
In other words, the frame classifier is trained independently from
the segmental model, and is held fixed when the segmental
model is being trained.  A natural question to ask is
whether we can further improve the sequence loss by updating the
frame classifier while holding the segmental model fixed.
A more general question would be whether we can
improve the sequence loss by
updating the frame classifier and the segmental model
simultaneously.

Optimizing all of the parameters simultaneously against a single loss function
is commonly referred to as end-to-end training.
Since the loss function with respect to all the parameters
in a neural network is non-convex,
end-to-end training might be more difficult than training in multiple stages.
On the other hand, since end-to-end training is jointly optimizing
all parameters, it might achieve a lower loss, leading to better
performance.


In this chapter, we formally define these training settings,
including multi-stage training and end-to-end training,
Different training approaches and losses
have different training requirements.
We discuss the pros and cons of these training approaches,
and conduct experiments to see empirically how well
they compare to each other.

\section{Training Settings}

Recall that the set of parameters $\Theta$ consists of
$\Theta_\text{enc}$ and $\Theta_\text{dec}$,
where $\Theta_\text{enc}$ is the set of parameters
in the encoder and $\Theta_\text{dec}$ are
the rest of the parameters in the weight function.

The training approach we have seen in the previous
chapter is referred to as \term{multi-stage training}.
Specifically, the encoder is trained
in the first stage
with an encoder-specific loss
\begin{align}
\mathcal{L}_\text{enc}(\Theta_\text{enc}),
\end{align}
for example, the frame-wise cross entropy.
Let $\hat{\Theta}_\text{enc} = \argmin_{\Theta_\text{enc}}
\mathcal{L}_\text{enc}(\Theta_\text{enc})$.
In the second stage, we solve the optimization problem
\begin{align} \label{eq:2nd-stage}
\min_{\Theta_\text{dec}} \mathcal{L}_\text{seq}(\hat{\Theta}_\text{enc}, \Theta_\text{dec}),
\end{align}
where $\mathcal{L}$ is a sequence loss, such as hinge loss,
log loss, or marginal log loss, defined in Section \ref{sec:losses}.
Note that in the second stage the parameters of the encoder are held fixed.
Typically, after the first stage, we can feed the inputs
in the entire data set forward, and reuse the vectors while optimizing
the sequence loss to save computation.
Another benefit of optimizing \eqref{eq:2nd-stage} with
$\hat{\Theta}_\text{enc}$ fixed is that the loss function $\mathcal{L}$
might be convex in $\Theta_\text{dec}$, for example,
when hinge loss or log loss is used and the weight function is linear
in $\Theta_{\text{dec}}$.
The characteristics of convex functions are well-understood \citep{BV2004},
and many algorithms, other than first-order methods,
have been developed for optimizing convex functions \citep{BNO2003}.

In contrast to multi-stage training,
we refer to solving
\begin{align} \label{eq:e2e}
\min_{\Theta_\text{enc}, \Theta_\text{dec}}
   \mathcal{L}_\text{seq}(\Theta_\text{enc}, \Theta_\text{dec})
\end{align}
as \term{end-to-end training},
where all parameters $\Theta_\text{enc}$ and $\Theta_\text{dec}$
are optimized jointly.
Since all parameters are optimized jointly, the loss function is
in general non-convex.
Though end-to-end training might be harder to solve,
it might lead to a lower loss value.
Typically, the solution found by multi-stage training is
a reasonably good solution for the sequence loss.
As a result, a safe strategy, which we refer to as \term{end-to-end fine-tuning},
may be to use the solution from multi-stage
training as an initialization for solving \eqref{eq:e2e}.
One major difference between multi-stage training
and end-to-end training is their resource requirements.
Multi-stage training requires extra labels for optimizing
the encoder loss $\mathcal{L}_\text{enc}$,
while end-to-end training requires only whatever the sequence
loss $\mathcal{L}_\text{seq}$ requires.


\section{Experiments}
\label{sec:e2e-exp}

For experiments, we use the same phoneme recognition setting
on the TIMIT data set as in Section \ref{sec:fast-dsc}.
In the sections below, we first compare the performance
of CNNs and LSTMs,
and then compare two training settings for training
segmental models, multi-stage training followed by
end-to-end fine-tuning and end-to-end training
from random initialization.
We also explore various weight functions
paired with various loss functions in the context
of end-to-end training from random initialization.

\subsection{Multi-stage training}

Recall that the multi-stage training approach
demonstrated in Section \ref{sec:dsc-exp}
consists of two stages: first training a frame classifier,
and second training a segmental model based on the output
probabilities of the frame classifier.
Instead of a CNN encoder, a 3-layer bidirectional LSTM with hidden vector size of 250
is trained for frame classification.
Parameters are initialized according to \citep{GB2010}.
Vanilla SGD is used
to minimize frame-wise cross entropy for 20 epochs with
step size 0.1 and a mini-batch size of 1 utterance.
Gradients are clipped to norm 5 if the norm is above 5.
The best model (measured by frame error rates on the development set)
is chosen among the 20 epochs, and trained for another
20 epochs with step size 0.75 decayed by 0.75 after every epoch.
For phoneme recognition, we use segmental models with the FC weight function,
the same setting as in Section \ref{sec:dsc-exp}.
Hinge loss with the overlap cost are minimized with RMSProp \citep{MH2017} for 20 epochs
with step size ${10}^{-4}$ and decay 0.9,

The frame error rates (FER) and phoneme error rates (PER)
compared to the CNN in Section \ref{sec:dsc-exp}
are shown in Table \ref{tbl:cnn-vs-lstm}.
The LSTM performs better than the CNN for both tasks.
We suspect that the LSTM's superior performance
is due to the training approaches:
the LSTM is trained on entire utterances,
while the CNN is trained on batches of 15-frame windows.
This effect was also observed in \citep{JVH2014}.
For the rest of the experiments, we use 3-layer
LSTMs as our encoders.

\begin{table}
\begin{center}
\caption{Frame error rates (FER) and phoneme error rates (PER)
    on the development set
    comparing the CNN and the LSTM encoder.}
\label{tbl:cnn-vs-lstm}
\begin{tabular}{lll}
           & FER (\%)   & PER (\%)   \\
\hline
CNN        & 22.3       & 22.2  \\
LSTM       & 17.8       & 19.7
\end{tabular}
\end{center}
\end{table}

Next, we compare segmental models
trained with different losses, including
hinge loss, log loss, and marginal log loss.
We use the FC weight function and a maximum segment duration of 30 frames.
All losses are minimized with RMSProp for 20 epochs with step size ${10}^{-4}$
and decay 0.9. Results are shown in Table \ref{tbl:2s-ft}.
No explicit regularizers are used except early stopping on
the development set.  All losses achieve similar results with
log loss slightly behind.  Note that although
marginal log loss does not require manual alignments,
the frame-wise cross entropy used to train the frame classifier still does.

After two-stage training, we use the trained segmental models
and frame classifiers as an initialization for end-to-end training.
Each loss is optimized with vanilla SGD
for 20 epochs with step size 0.1 decayed by 0.75 after each epoch.
Gradients are clipped with norm 5.  Dropout is added to the LSTMs
with rate 0.2, and no other regularizers are used except early stopping on
the development set.  Results are shown in Table \ref{tbl:2s-ft}.
End-to-end fine-tuning is able to improve the error rates
for all losses with marginal log loss being the best performer.

\begin{table}
\caption{Phoneme error rates (\%) comparing different losses
    with two-stage training (2s) followed by end-to-end fine-tuning (ft).}
\label{tbl:2s-ft}
\begin{center}
\begin{tabular}{p{2cm}l|ll|ll}
weight
       &          & \multicolumn{2}{l|}{2s} & \multicolumn{2}{l}{+fine-tuning} \\
function
       & loss     & dev    & test   & dev      & test    \\
\hline
FC     & hinge    & 19.7   &        & 18.2     &         \\
       & log      & 21.0   &        & 17.3     &         \\
       & MLL      & 21.8   &        & 16.9     & 19.5
\end{tabular}
\end{center}
\end{table}

After end-to-end fine-tuning, the encoders are not constrained
to perform well on frame-wise cross entropy.  We evaluate
the encoders on the frame classification task to see
how much the encoders deviate from the initialization,
and whether the intermediate representations trained
for frame classification are still preserved.
Results are shown in Table \ref{tbl:frame-after-ft}.
The encoders fine-tuned with hinge loss and log loss
deviate less compared to the one fine-tuned with marginal log loss.
We suspect this is because hinge loss and log loss use
the ground-truth alignments in training while marginal
log loss does not.

\begin{table}
\caption{Frame error rates (\%) of LSTM encoders
    after end-to-end fine-tuning with different losses .}
\label{tbl:frame-after-ft}
\begin{center}
\begin{tabular}{ll|ll|ll}
       &          & FER    \\
\hline
\multicolumn{2}{l|}{frame-wise cross entropy}
                  & 17.8   \\
\hline
FC     & hinge    & 18.4   \\
       & log      & 18.7   \\
       & MLL      & 26.8
\end{tabular}
\end{center}
\end{table}

\subsection{End-to-end training}

After observing the success of end-to-end fine-tuning,
we conduct experiments to see whether
it is possible to train segmental models
end to end from random initialization.
We use the FC weight function and a maximum duration of 30 frames,
the same setting as in the multi-stage experiments. 
We initialize the parameters according to \citep{GB2010}.
Each loss is optimized with vanilla SGD for 20 epochs
with step size 0.1 and gradient clipping with norm 5.
The best performing model, chosen in the first 20 epochs,
is trained for another 20 epochs with vanilla SGD step size 0.75
decayed by 0.75 after each epoch.
Dropout of rate 0.2 is used throughout the training process.
No other regularizers are used except early stopping on
the development set.
Results are shown in Table \ref{tbl:e2e-fc}.

There is no significant difference in terms of performance between
models trained end to end from random initialization
and ones trained in multiple stages followed
by fine-tuning.
This shows that it is possible to train segmental models
end to end from random initialization,
and that models trained in multiple stages
can serve as a good initialization for end-to-end training.

\begin{table}
\caption{Phoneme error rates (\%) on the development set
    comparing multi-stage training
    followed by end-to-end fine-tuning (2s+ft) and end-to-end training
    from random initialization (e2e).}
\label{tbl:e2e-fc}
\begin{center}
\begin{tabular}{p{2cm}l|ll}
weight \newline function
       & loss   & 2s+ft  & e2e    \\
\hline
FC     & hinge  & 18.2   & 18.4   \\
       & log    & 17.3   & 17.4   \\
       & MLL    & 16.9   & 16.7   \\
\end{tabular}
\end{center}
\end{table}

\subsection{Weight function comparison}

\begin{figure}
\begin{center}
\includegraphics[width=8cm]{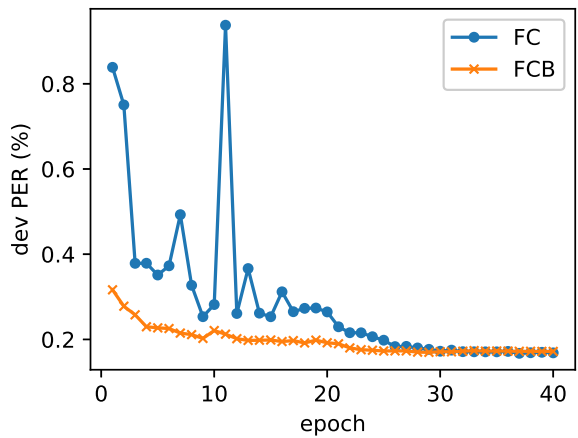}
\caption{Learning curves of segmental models trained end to end
    with the FC weight function and FCB weight function
    with marginal log loss.}
\label{fig:fc-fcb}
\end{center}
\end{figure}

Next, we consider various weight functions for end-to-end
training from random initialization.
Note that the initialization scheme in \citep{GB2010}
is not suitable the for log-soft-max operation used after
the LSTMs for producing log probabilities,
because the variances before and after
the log-soft-max operation are very different.
Based on this intuition, we
define a new weight function, called
\term{FC bottleneck (FCB)} weight function,
removing the log-soft-max layer when using the FC weight function.
More precisely, suppose $h_1, \dots, h_T$ 
is the sequence of vectors produced by an LSTM.
Whereas the FC weight function
uses the average, samples, and boundaries of
$z_i = \text{logsoftmax}(W h_i + b)$ for $i = 1, \dots, T$,
the FCB weight function uses
the average, samples, and boundaries of $h_i$'s
directly.
The learning curves of segmental models trained end to end
with FC weight function and the FCB weight function
are shown in Figure \ref{fig:fc-fcb}.
We observe significantly faster convergence with the FCB
weight function than with the FC weight function.

We do not compare the MLP weight function to the FC and
FCB weight functions in the current setting,
because it is too time-consuming to train segmental models
with the MLP weight function on the entire search space.
To train segmental models with the MLP weight function,
we follow \citep{LKDSR2016} and reduce the time resolution by a factor of four
using pyramid LSTMs.
Specifically, for an input sequence $x_{1:T}$,
\begin{align}
h^{(1)}_{1:T} &= \text{BLSTM}(x_{1:T}) \\
z^{(1)}_{1:\lfloor T/2 \rfloor} &= \text{subsample}(h^{(1)}_{1:T}) \\
h^{(2)}_{1:\lfloor T/2 \rfloor} &= \text{BLSTM}(z^{(1)}_{1:\lfloor T/2 \rfloor}) \\
z^{(2)}_{1:\lfloor T/4 \rfloor} &= \text{subsample}(h^{(2)}_{1:\lfloor T/2 \rfloor}) \\
h^{(3)}_{1:\lfloor T/4 \rfloor} &= \text{BLSTM}(z^{(2)}_{1:\lfloor T/4 \rfloor})
\end{align}
where
\begin{align}
h_2, h_4, \dots, h_{2(k-1)}, h_{2k} = \text{subsample}(h_1, h_2, \dots, h_{2k})
\end{align}
if we drop a frame for every two frames.
The final output sequence $h^{(3)}$ is then used
to compute edge weights.
An encoder with several layers of LSTMs interleaved with subsampling 
is also known as a \term{pyramid LSTM}
and has been used in other prior work, e.g., \citep{CJLV2016}.
In addition, we set the maximum duration to 8 frames,
effectively $4 \times 8 = 32$ frames on
the original time scale.
The reduced time resolution and maximum duration
result in search spaces 16 times smaller than the original ones,
making it feasible to compute the MLP weight function
on all edges.

With 3-layer pyramid LSTMs, we compare segmental models
trained with the three weight functions and three losses.
All models are trained with the same step size scheduling
as in the previous experiments.
Results with and without pyramid LSTMs are both shown in Table \ref{tbl:e2e}.
First, we observe that using pyramid LSTMs leads
to a degradation in performance for segmental models
with the FC and FCB weight functions.
However, using pyramid LSTMs has less impact on segmental models
trained with marginal log loss compared to the ones
trained with hinge loss and log loss.
We suspect this is because hinge loss and log loss rely
on the ground-truth alignments for training while marginal log loss does not.
Second, we observe that while segmental models with the
FC and FCB weight functions are able to achieve low hinge loss values,
the ones with the MLP weight
function completely fail to do so.
Finally, the best result on the development set
obtained by using the FC weight function
is slightly worse than using the MLP weight function.

\begin{table}
\caption{Phoneme error rates (\%) comparing different losses
    for end-to-end training from random initialization.}
\label{tbl:e2e}
\begin{center}
\begin{tabular}{p{2cm}l|ll|ll}
weight \newline function
       & loss     & \multicolumn{2}{p{2cm}|}{regular \newline LSTM}    & \multicolumn{2}{p{2cm}}{pyramid \newline LSTM} \\
\hline
\hline
       &          & dev    & test   & dev      & test    \\
\hline
FC     & hinge    & 18.4   &        & 23.4     &         \\
       & log      & 17.4   &        & 22.4     &         \\
       & MLL      & 16.7   & 19.6   & 17.9     &         \\
\hline
FCB    & hinge    & 18.6   &        & 24.2     &         \\
       & log      & 17.8   &        & 23.0     &         \\
       & MLL      & 17.0   &        & 17.7     &         \\
\hline
MLP    & hinge    &        &        & 67.7     &         \\
       & log      &        &        & 22.4     &         \\
       & MLL      &        &        & 17.1     & 19.2
\end{tabular}
\end{center}
\end{table}


Segmental models trained with marginal log loss
do not require ground-truth
alignments during training, making annotating data sets
easier, cheaper, and less time-consuming.
However, it is harder to diagnose errors made by models
trained end to end because the intermediate representations
are not interpretable.  Models trained in multiple stages
are easier to diagnose by looking at the performance
of proxy tasks, such as frame classification in our case.

\subsection{Training time comparison}

We compare the time to train segmental models end to end.
Times are measured on a 3GHz 4-core CPU.
Multithreading is only used for matrix operations;
all FST algorithms are implemented single-threaded.
Results are shown in Table \ref{tbl:runtime}.
Reducing the time resolution with pyramid LSTMs
significantly improves the runtime for all losses.
The MLP weight function is about 2.5 times
slower than the FC weight function.

\begin{table}
\caption{Average number of minutes for one epoch of end-to-end training on TIMIT.}
\label{tbl:runtime}
\begin{center}
\begin{tabular}{p{2cm}l|p{1.5cm}p{1.5cm}}
weight \newline function
       & loss     & regular \newline LSTM    & pyramid \newline LSTM  \\
\hline
FC     & hinge    & 119    & 38       \\ 
       & log      & 177    & 44       \\ 
       & MLL      & 260    & 47       \\ 
\hline                           
MLP    & hinge    &        & 83       \\ 
       & log      &        & 110      \\ 
       & MLL      &        & 116 
\end{tabular}
\end{center}
\end{table}

\subsection{Segmentation Analysis}

We are interested in how well
segmental models recover the phone boundaries
in the end-to-end setting when
manual alignments are not used during training.
The task, commonly known as phonetic segmentation,
is to align the phonetic transcription
to the acoustic frames.
Unlike sequence prediction, the phonetic transcription
is known.
We measure the error rates
of the boundaries under some tolerance level.
Specifically, suppose the predicted boundary is at time $\hat{b}$
and the ground truth is at time $b$.  The predicted boundary $\hat{b}$
is considered correct if the absolute distance
$t = |\hat{b} - b| \leq \tau$ for some $\tau$.
The error rates under various thresholds
for the segmental model with
the FC weight function trained with marginal log loss
are shown in Table \ref{tbl:align}.
Models proposed in \citep{KSSC2007,RSHB2012,YRLSMW2013},
shown in row 1--3 in Table \ref{tbl:align}, are
specifically trained for phonetic segmentation.
Though the alignment results are behind models
trained specifically to align,
the segmental model trained with marginal log loss
is not supervised with any ground-truth alignments,
and its performance is sufficient for many tasks.

\begin{table}
\caption{Boundary error rates (\%) of phonetic segmentation at different tolerance levels
    on the TIMIT core test set (except the last row)
    for segmental models trained end to end with marginal log loss
    compared to past results.}
\label{tbl:align}
\begin{center}
\begin{tabular}{llllll}
    & $t \leq 0\text{ms}$ & $t \leq 10\text{ms}$
    & $t \leq 20\text{ms}$ & $t \leq 30\text{ms}$ & $t \leq 40\text{ms}$ \\
\hline
\citep{KSSC2007}    &        & 20.3\%  & 7.9\%  & 3.8\%  & 1.9\% \\
\citep{RSHB2012}    &        & 19.5\%  & 6.3\%  & 2.4\%  & 1.0\% \\
\citep{YRLSMW2013}  & 22.6\% & 6.1\%   & 2.6\%  & 1.2\%  & 0.6\% \\
\hline
\hline
seg FC & 25.0\% & 10.0\%  & 5.1\%  & 3.1\%  & 2.1\% \\
\hline
seg FC on train  & 24.2\% & 9.6\%  & 5.0\%  & 3.1\%  & 2.1\% \\
\end{tabular}
\end{center}
\end{table}

We analyze the errors made by the segmental model.
The top 30 most errorful boundary types sorted by error rates
are shown in Table \ref{tbl:align-err-rate}.
We observe that most of the boundaries in Table \ref{tbl:align-err-rate}
involve silences, vowels, and semi-vowels.
The ones involving silences often appears at the start or the end of utterances.
Boundaries between two vowels and between a semi-vowel and a vowel
are known to be ambiguous \citep{SM1975,U1975}.
We see few alignment errors between consonants and vowels.

The top 30 most errorful boundary types sorted by error counts
are shown in Table \ref{tbl:align-err-count}.
We are also interested in boundary types that have hight error counts but
not necessary have high error rates.
Compared to Table \ref{tbl:align-err-rate}, we see
a very different pattern. Most of the errors in Table \ref{tbl:align-err-count}
involve silences (including voiced and unvoiced closures).
For $0\text{ms}$ and $10\text{ms}$, the segmental model
errs at the boundaries between closures and the bursts
of stop consonants, such as /b/, /d/, /g/, /p/, /t/, and /g/.
It also errs at boundaries involving liquids, such as /r/ and /l/,
across all tolerance levels.
Similarly to semi-vowels, the boundaries of liquids are also
known to be ambiguous \citep{SM1975,U1975}.
The errors for higher tolerance levels, such as 30ms and 40ms,
mostly involve silences that appear at the start
or end of utterances.
We notice that 10.5\% of the silences in the training set are longer than 30 frames.
We suspect the 30-frame maximum duration constraint is too restrictive for silences.

Some phoneme boundaries are inherently ambiguous.
In other words, there are cases where segments do not have
clear start times and end times.
We argue that segmental models are still
well-motivated and suitable
for these tasks, such as phonetic recognition.
First, the search space are stochastic
when the path weights are interpreted as probabilities.
In fact, lattices, when first proposed by \cite{SM1975}, were used to account for
phoneme boundary ambiguity.
Training segmental models with marginal log loss also takes
the ambiguity into account by marginalizing over
all segmentations.
For decoding, ideally we want to find
the label sequence that has the maximum
marginal weight
\begin{align}
\argmax_{y} \sum_{p \in \Gamma(y)} w(p), \label{eq:decode-marginal}
\end{align}
where $\Gamma(y)$ is the set of paths with the label sequence $y$.
However, the above is not tractable
and we typically approximate it by finding the most probable path
\begin{align}
\argmax_{p \in \mathcal{P}} w(p).
\end{align}
Another approach to approximate \eqref{eq:decode-marginal}
is to use beam search.
In summary, segmental models are capable of handling
ambiguous segment boundaries if the training and decoding
take the ambiguity into account.

\section{Summary}

We have compared different training approaches for segmental models,
including multi-stage training and end-to-end training.
End-to-end training improves over multi-stage
training, and end-to-end training from random initialization
is on par with end-to-end fine-tuning.
Multi-stage training is useful for diagnosing
end-to-end training, because every model found
by multi-stage training is a valid model for
end-to-end training objectives.
We have shown that segmental models can be trained with marginal log loss
from random initialization,
without requiring manual alignments for training.
As a byproduct, segmental models trained with marginal log loss
perform reasonably well on phonetic segmentation.
Segmentation errors made by
end-to-end segmental models are aligned with the studies
in acoustic phonetics.

\begin{table}
\begin{center}
\caption{Top 30 boundaries and its preceding and following phonemes
    (with at least five occurrences) in the training set
    sorted according to alignment error rates of
    the segmental model trained with marginal log loss.}
\label{tbl:align-err-rate}
\begin{tabular}{ll|ll|ll|ll|ll}
\multicolumn{2}{l|}{$t \leq 0\text{ms}$}
    & \multicolumn{2}{l|}{$t \leq 10\text{ms}$}
    & \multicolumn{2}{l|}{$t \leq 20\text{ms}$}
    & \multicolumn{2}{l|}{$t \leq 30\text{ms}$}
    & \multicolumn{2}{l}{$t \leq 40\text{ms}$} \\
\hline
en & w    &  ix & ae  &  ix & ax  &  uw & uw  &  ix & ae  \\ 
sil & ae  &  ix & eh  &  ih & eh  &  ax & eh  &  ng & ey  \\
dx & aw   &  ih & eh  &  ao & aw  &  ix & ae  &  uw & uw  \\
sil & y   &  sil & aw &  ax & eh  &  ax & aa  &  ao & aw  \\
ao & w    &  ao & aw  &  ax & aw  &  eh & y   &  eh & y   \\
ix & ae   &  y & iy   &  eh & y   &  ih & eh  &  ih & iy  \\
oy & ae   &  ah & aa  &  uw & uw  &  ix & eh  &  hh & sil \\
s & y     &  ax & eh  &  ix & eh  &  hh & sil &  ix & eh  \\
oy & ao   &  ao & ey  &  ax & aa  &  ng & ey  &  ih & eh  \\
jh & aa   &  ax & aw  &  ao & eh  &  sil & aa &  ih & sil \\
\hline
sh & ay   &  ng & ey  &  sil & ao &  ao & eh  &  w & oy   \\
jh & uh   &  ax & ay  &  cl & ow  &  ow & ow  &  ax & ay  \\
f & w     &  ao & ow  &  ih & iy  &  cl & ao  &  vcl & ah \\
ix & eh   &  eh & y   &  sil & ay &  ao & aw  &  ow & ow  \\
z & ay    &  cl & ah  &  ix & ih  &  ah & aa  &  cl & ao  \\
el & ay   &  uw & uw  &  y & iy   &  ih & iy  &  ao & eh  \\
ih & eh   &  ey & ey  &  aw & ay  &  ax & ow  &  ix & ih  \\
z & y     &  aa & ax  &  ax & ay  &  ih & sil &  sil & aa \\
sil & aw  &  aw & oy  &  ao & ow  &  ix & ih  &  sil & ao \\
ax & ow   &  w & en   &  sil & aa &  w & oy   &  er & aw  \\
\hline
jh & ao   &  sil & y  &  sil & w  &  ax & aw  &  sil & ay \\
n & oy    &  sil & w  &  ow & ow  &  aw & ay  &  oy & ao  \\
ng & eh   &  sil & ow &  cl & ao  &  ax & ay  &  ow & ey  \\
l & en    &  sil & ay &  ix & ow  &  vcl & ah &  ax & ow  \\
epi & dh  &  sil & ah &  er & sil &  ey & ey  &  ay & ey  \\
aw & hh   &  sil & aa &  sil & ah &  sil & ao &  ax & aa  \\
s & r     &  sil & r  &  hh & sil &  sil & ow &  y & iy   \\
m & aw    &  sil & ao &  ng & ey  &  ix & iy  &  ax & eh  \\
sh & ah   &  sil & dh &  ao & ih  &  sil & ay &  ao & ey  \\
aa & z    &  ix & ih  &  iy & iy  &  oy & ae  &  ax & aw
\end{tabular}
\end{center}
\end{table}

\begin{table}
\begin{center}
\caption{Top 30 boundaries and its preceeding and following phonemes
    (with at least five occurrences) in the training set
    sorted according to alignment error counts of
    the segmental model trained with marginal log loss.}
\label{tbl:align-err-count}
\begin{tabular}{ll|ll|ll|ll|ll}
\multicolumn{2}{l|}{$t \leq 0\text{ms}$}
    & \multicolumn{2}{l|}{$t \leq 10\text{ms}$}
    & \multicolumn{2}{l|}{$t \leq 20\text{ms}$}
    & \multicolumn{2}{l|}{$t \leq 30\text{ms}$}
    & \multicolumn{2}{l}{$t \leq 40\text{ms}$} \\
\hline
vcl & b  &  vcl & b  &  sil & dh  &  sil & dh  &  iy & sil \\
vcl & d  &  sil & dh &  sil & w   &  sil & w   &  er & sil \\
cl & k   &  vcl & d  &  er & sil  &  er & sil  &  sil & dh \\
cl & t   &  vcl & g  &  iy & sil  &  iy & sil  &  n & sil  \\
cl & p   &  sil & w  &  n & sil   &  n & sil   &  sil & w  \\
s & cl   &  t & r    &  t & r     &  s & sil   &  s & sil  \\
ix & n   &  er & sil &  k & l     &  cl & sil  &  cl & sil \\
vcl & g  &  iy & sil &  sil & hh  &  m & sil   &  ng & sil \\
sil & dh &  ao & r   &  ao & r    &  ng & sil  &  m & sil  \\
ix & cl  &  p & r    &  s & sil   &  sil & hh  &  z & sil  \\
\hline
vcl & jh &  sil & hh &  p & r     &  sil & r   &  m & sil  \\
n & cl   &  l & iy   &  m & sil   &  ao & r    &  z & sil  \\
ao & r   &  aa & r   &  p & l     &  z & sil   &  sil & aa \\
n & vcl  &  n & sil  &  vcl & b   &  ao & l    &  sil & ay \\
ix & z   &  y & uw   &  ng & sil  &  sil & aa  &  sil & hh \\
ix & vcl &  p & l    &  cl & sil  &  sil & ae  &  d & sil  \\
s & sil  &  k & l    &  aa & r    &  sil & y   &  vcl & sil\\
aa & r   &  r & iy   &  sil & m   &  vcl & sil &  l & sil  \\
l & iy   &  s & sil  &  y & uw    &  el & sil  &  ao & l   \\
r & iy   &  n & vcl  &  k & r     &  sil & m   &  en & sil \\
\hline
iy & cl  &  sil & b  &  ao & l    &  sil & ay  &  sil & ao \\
y & uw   &  ih & z   &  l & iy    &  d & sil   &  sil & ae \\
ax & cl  &  k & r    &  sil & r   &  k & l     &  t & sil  \\
dx & ix  &  vcl & jh &  r & ay    &  y & uw    &  sil & ah \\
t & r    &  ix & z   &  sil & b   &  l & sil   &  ao & r   \\
er & cl  &  n & cl   &  sil & n   &  aa & r    &  aa & r   \\
ix & s   &  m & sil  &  r & aa    &  t & sil   &  ix & sil \\
p & r    &  z & sil  &  sil & ae  &  en & sil  &  sil & y  \\
cl & ch  &  sil & m  &  z & sil   &  sil & ah  &  el & sil \\
dh & ax  &  k & w    &  k & w     &  sil & ax  &  sil & r  
\end{tabular}
\end{center}
\end{table}

%% file: related.tex
\chapter{Historical Overview of Segmental Models}

Automatic speech recognition (ASR) has been posed as a
graph search problem since the 1970s \citep{J1976}.
A search space, represented as a graph, is first constructed;
edges in the graph are assigned weights based on the input
and the edges; to predict, we simply
run the shortest-path algorithm on the graph.
The graph search paradigm has been popularized by
the use of hidden Markov models (HMM) \citep{R1989,J1998},
where the shortest-path algorithm is known as the Viterbi algorithm.
In the probabilistic setting, finding the shortest path can
be seen as finding the maximum a posteriori solution,
justifying the graph search paradigm.  The probabilistic view has spawned
many algorithms,
such as segmental k-means \citep{JR1990} and expectation maximization
\citep{RM1985},
for estimating parameters in the weight function.

Segmental models follow the same graph search paradigm
while having a different type of search spaces and
different weight functions.
Many variants of segmental models were proposed in the past, such
as stochastic segment models \citep{OR1989,ODK1996}, semi-Markov HMMs \citep{RC1987},
segmental HMMs \citep{R1993,GY1993}, semi-Markov conditional random fields (CRF) \citep{SC2005},
and segmental CRFs \citep{ZN2009}.  However, the type of search spaces
and the shortest-path algorithm stay the same.
In this chapter, we review these variants from
the graph search point of view.

The idea of using segmental features for speech recognition can
be traced back to the 1970s \citep{WMMZ1975}.  The definition of
segmental models was not explicit.  Most of the
studies still followed the graph search paradigm,
though the graph structures were typically constructed
from a lexicon with a small vocabulary,
and the weights on the edges were typically
estimated with heuristics or a small amount of data
\citep{KB1985,CSPBPS1983}.

In the following sections, we categorize variants of
segmental models as either generative or discriminative.
This categorization also aligns well with
their rough chronological order.

\section{Generative Segmental Models}

\term{Hidden semi-Markov Models} are arguably
the first segmental models applied to speech recognition \citep{L1986,RC1987}.
Given a sequence of observations $x = (x_1, \dots, x_T)$ of length $T$,
let $y = (\ell_1, \dots, \ell_K)$ be the label sequence
and $z = ((s_1, t_1), \dots,\allowbreak (s_K, t_K))$ be the segmentation,
where $\ell_1, \dots, \ell_K \in L$
for some discrete label set $L$,
and $s_1 = 1$, $t_k = T$, $s_k \leq t_k$, $t_{k-1} + 1 = s_k$,
for $k = 2, \dots, n$.
Let $e_k = (\ell_k, s_k, t_k)$ for $k = 1, \dots, K$.
The probability of $x_{1:T}$ defined by hidden semi-Markov models is
\begin{align}
p(x, y, z) = p(x_{1:T}, e_{1:K})
    = p(e_1) \prod_{k=2}^K p(e_k | e_{k-1}) \prod_{k=1}^K p(x_{s_k:t_k} | e_k).
\end{align}
The generative story is straightforward: the segments are generated one by one, following
a Markov chain, and each segment $e = (\ell, s, t)$ generates $t - s + 1$ observations.
Hidden Markov models are special cases of hidden semi-Markov models
with the constraints $K = T$ and $s_k = t_k$ for $k = 1, \dots, K$.

There are many ways to define and parameterize
$p(e_k | e_{k-1})$ and $p(x_{s_k:t_k} | e_k)$.
The most common assumptions are
\begin{align}
p(e_k | e_{k-1}) & = p(\ell_k | \ell_{k-1}) \\
p(x_{s_k:t_k} | e_k) & = p(x_{s_k:t_k} | t-s+1, \ell_k) p(t-s+1 | \ell_k)
\end{align}
where $p(\ell_k | \ell_{k-1})$ is commonly known as the transition probability,
$p(x_{s_k:t_k} | t-s+1, \ell_k)$ the emission probability,
and $p(t-s+1 | \ell_k)$ the duration probability.
The observations within a segment are typically assumed to be independent, i.e.,
\begin{align}
p(x_{s_k:t_k} | t-s+1, \ell_k) =
    \prod_{j=s_k}^{t_k} \mathcal{N}(x_j; \mu_{\ell_k}, \sigma_{\ell_k}^2),
\end{align}
where $\mu_\ell$ and $\sigma_\ell^2$ are the mean and variance of the Gaussian distribution
for the label $\ell$.
The single Gaussian case can be easily extended to a mixture of Gaussians.
Continuously variable duration HMMs \citep{L1986},
stochastic segment models \citep{OR1989} and segmental HMMs \citep{R1993,GY1993}
are all hidden semi-Markov models with the above assumptions.
The difference among them lies
in how the emission probability $p(x_{s_k:t_k} | t-s+1, \ell_k)$ is defined.
The subtle differences are summarized in \citep{GY1993-2,ODK1996}.
The duration probability is typically defined by a Poisson distribution \citep{RM1985}
or Gamma distribution \citep{L1986}.  See \citep{ODK1996} and the citations therein 
for other options.

Decoding with hidden semi-Markov models
is done by solving
\begin{align}
\argmax_{y, z} p(y, z | x)
    & = \argmax_{y, z} \log p(y, z | x) = \argmax_{y, z} \log \frac{p(x, y, z)}{p(x)} \\
    & = \argmax_{y, z} \log p(x, y, z),
\end{align}
where $y$ is the label sequence and $z$ is the segmentation,
and is equivalent to finding the maximum-weight path with
\begin{align}
w((\ell, s, t)) = \log p(t-s+1 | \ell)
    + \sum_{j=s}^t \log \mathcal{N}(x_j; \mu_\ell, \sigma_\ell^2).
\end{align}
The term $p(\ell' | \ell)$ is ignored here for simplicity, but can be
included once the search space is self-expanded (Section \ref{sec:self-expansion}).
Training hidden semi-Markov models can be done by maximizing the
likelihood of the training set.  The likelihood can be maximized
using gradient-based methods or expectation maximization (EM) \citep{RM1985}.
See \citep{GY1993-2} for a detailed explanation of the EM algorithm
for estimating the parameters of hidden semi-Markov models.

Originally motivated by using rich segmental features \citep{ZGPS1989},
the SUMMIT system was defined as a generative model \citep{GCM1996}.
However, \cite{GCM1996} introduced the notion of anti-phones
and later \cite{CG1997} introduced near-misses,
training the system with a discriminative touch.
It held the state-of-the-art result for
speaker-independent phoneme recognition on TIMIT \citep{H1998}
until the rise of deep neural networks \citep{MDH2009}.

\section{Discriminative Segmental Models}

Since maximum mutual information was introduced
as a training criterion for HMMs \citep{BBSM1986},
ASR studies have gradually
shifted from generative to discriminative modeling.

Parallel to the development of segmental models in the ASR
community, \cite{SC2005} proposed \term{semi-Markov CRFs}
for named entity recognition.
Using the same notation as in the previous section,
the probability of a label sequence $y = (\ell_1, \dots, \ell_n)$
and a segmentation $z = ((s_1, t_1), \dots, (s_K, t_K))$
given an observation sequence $x = (x_1, \dots, x_T)$ is defined as
\begin{align}
p(y, z | x) = \frac{1}{Z(x)} \exp\left( \sum_{k=1}^K \theta^\top \phi(x, e_k) \right)
\end{align}
where $Z(x) = \sum_{y', z'} \exp(\sum_{e \in (y', z')} \theta^\top \phi(x, e))$
and $e \in (y', z')$ is a shorthand for enumerating
$(\ell_1, s_1, t_1),\allowbreak \dots,\allowbreak (\ell_{K'}, s_{K'}, t_{K'})$
for $y' = (\ell_1, \dots, \ell_{K'})$ and $z' = ((s_1, t_1), \dots, (s_{K'}, y_{K'}))$
of length $K'$.
The function $\phi$, called the feature function,
extracts feature vectors that are intended to correlate well with
$y$ and $z$.
The parameter vector $\theta$ can be learned by maximizing the
likelihood of the training set.

Decoding with semi-Markov CRFs is done by solving
\begin{align}
\argmax_{y, z} \log p(y, z | x) = \argmax_{y, z} \sum_{e \in (y, z)} \theta^\top \phi(x, e),
\end{align}
and is equivalent
to finding the maximum-weight path if we
let $w(x, e) = \theta^\top \phi(x, e)$.
Similarly, training is done by maximizing the conditional likelihood
of the data, and is equivalent to minimizing
the negative log likelihood, or log loss.

Heavily influenced by \citep{SC2005},
\cite{ZN2009} proposed \term{segmental CRFs}.
The difference between segmental CRFs and semi-Markov CRFs lies
in the training loss.
Instead of optimizing the conditional likelihood $p(y, z | x)$,
segmental CRFs optimize the marginal likelihood
\begin{align}
p(y | x) = \sum_{z} p(y, z | x).
\end{align}
The connection between the marginal likelihood and marginal log loss
is clear once we take the log of the probability distribution.

As we defined in Chapter \ref{ch:seg}, we distinguish between
segmental models that consider the entire search space and ones
that do not.  The former are called first-pass segmental models.
Whether a segmental model is first-pass or not is independent
of its definition.  For example, semi-Markov CRFs
were used as first-pass segmental models in \citep{SC2005};
segmental CRFs were first used as a second-pass model
in \citep{ZN2009}, and were later used as a first-pass
model in \citep{Z2012}.

In Table~\ref{tbl:seg}, we provide a set of highlights of results
in the development of segmental models on the TIMIT data set.
\cite{Z2012} was the first to explore
discriminative segmental models that search over sequences
and segmentations exhaustively,
and did not use neural networks.
\cite{HF2012} first used (shallow) neural network-based
frame classifiers to define weight functions,
and later extended the idea to deep neural networks in~\citep{H2015}.
\cite{ADYJ2013} were the first to use
deep convolutional neural networks for the weight functions,
and were the first to train segmental models end to end.
We have compared different
losses and training strategies for segmental models, first in a rescoring framework~\citep{TGL2014} 
and then in first-pass segmental models~\citep{TWGL2016}.
We also introduced segment-level classifiers and segmental cascades
for incorporating them (and other expensive features)
into segmental weight functions~\citep{TWGL2015}.
\cite{LKDSR2016} introduced an LSTM-based weight function for every segment,
and were also the first to use pyramid LSTMs to speed up inference for segmental models.

\begin{table}
\begin{center}
\caption{TIMIT PERs (\%) for various segmental models
    compared with HMMs and the current state of the art.
    The acoustic features are speaker-independent (\textnormal{spk indep}) or
    speaker-adapted with mean and variance normalization (\textnormal{mvn})
    or maximum likelihood linear regression (\textnormal{fMLLR})~\citep{P2011}.
    Some results were obtained with MFCCs and some with log filter bank features.
}
\label{tbl:seg}
\renewcommand{\arraystretch}{1.5}
\begin{tabular}{p{8cm}lll}
              & spk indep   & +mvn        & +fMLLR \\
\hline
HMM-DNN \newline \citep{P2011}
              & 21.4        &             & 18.3 \\
HMM-CNN \newline \citep{T2015}
              & 16.5 \\
\hline
SUMMIT \newline \citep{HG1998,G2003}
              & 24.4 \\
segmental CRF (SCRF) \newline
\citep{Z2012}
              & 33.1 \\
SCRF + shallow NN \newline
\citep{HF2012}
              & 26.5 \\
SCRF + DNN \newline
\citep{H2015}
              &             &             & 19.1 \\
deep segmental NN \newline
\citep{ADYJ2013}
              & 21.9 \\
segmental cascades \newline
\citep{TWGL2015}
              & 19.9 \\
segmental RNN (SRNN) \newline
\citep{LKDSR2016}
              &             & 18.9        & 17.3 \\
two-stage + end-to-end training \newline
\citep{TWGL2016}
              & 19.7 \\
SRNN + multitask
\newline \citep{LKDS2017, T+2017}
              &             & 18.5        & 17.5 \\
\hline
\hline
Additional results in this work \\
\hspace{0.2cm} two-stage + end-to-end seg FC
              & 19.6 \\
\hspace{0.2cm} end-to-end seg FC
              & 19.6 \\
\hspace{0.2cm} end-to-end seg MLP
              & 19.2
\end{tabular}
\end{center}
\end{table}

\section{Summary}

In this chapter, we have reviewed variants of segmental models
categorized as either generative or discriminative
in rough chronological order.
We review hidden semi-Markov models,
a broad class of generative segmental models
subsuming stochastic segment models and segmental HMMs.
We then review semi-Markov CRFs,
the discriminative counterpart of hidden semi-Markov models.
We have established connections between these special cases
and the general segmental models defined in Chapter \ref{ch:seg}.

%% file: unified.tex
\chapter{A Unified Framework for Graph Search-Based Models}

Beyond segmental models,
this chapter reviews other modern models trained end to end
within the graph search paradigm.
We review connectionist temporal classification (CTC) \citep{GFGS2006},
a popular approach for training frame-based LSTMs end to end.
Drawing on the connection between CTC and marginal log loss,
we propose a framework, consisting of search spaces (represented
as FSTs), weight functions, and training losses,
that can encompass many end-to-end models, such as hidden Markov models
trained with lattice-free maximum mutual information \citep{P+2016},
and LSTMs trained with CTC, as special cases.

We compare end-to-end segmental models and end-to-end frame-based
models, including one-state HMMs and two-state HMMs with LSTM encoders,
and LSTMs trained with CTC.  Having these end-to-end models
within the unified framework allows us to see the effect of each
component while holding the other components fixed.

\section{Connectionist Temporal Classification}

In this section, we review connectionist temporal classification (CTC)
\citep{GFGS2006}.  While being conceptually simple, LSTMs trained with CTC were the state
of art for phonetic recognition in 2013 \citep{GMH2013},
and have achieved competitive results on large-vocabular speech recognition
\citep{MGM2015, MXJN2015, ZYDS2017}.

Consider a sequence of acoustic vectors $x_1, \dots, x_T$ and
its corresponding labels $y_1, \dots, y_n$,
where $y_i \in L$ for $i = 1, \dots, n$ and some label set $L$.
We assume $n < T$ because a phoneme is typically more than a frame long.
Suppose there exists a function $\mathcal{P}$ that maps $y_1, \dots, y_n$
to a path $a_1, \dots, a_T$, where $a_t \in L'$ for some other label set $L'$.
Minimizing $p(a_{1:T} | x_{1:T})$
can be done by simply minimizing the frame-wise cross entropy
\begin{align}
p(a_{1:T} | x_{1:T}) = \prod_{t=1}^T p(a_t | x_t).
\end{align}
\cite{GFGS2006} proposed a mapping $\mathcal{P}$ as follows.
Given a sequence $a_1, \dots, a_T$ where $a_t \in L \cup \{\varnothing\}$
for $t = 1, \dots, T$ and $\varnothing$ is the blank symbol.
Let $\mathcal{B}$ be a function that first replaces
duplicate contiguous labels into a single label,
and second removes all the blank symbols.
For example, 
\begin{align}
\text{\texttt{k} \texttt{ae} \texttt{t}} = \mathcal{B}(\text{%
    $\varnothing$ \texttt{k} \texttt{k} $\varnothing$ \texttt{ae} \texttt{ae} \texttt{t}
    \texttt{t} $\varnothing$}).
\end{align}
\cite{GFGS2006} used
\begin{align}
\mathcal{P}(y) = \mathcal{B}^{-1}(y) = \{a = (a_1, \dots, a_T) :
    a_t \in L \cup \{\varnothing\}, \mathcal{B}(a) = y\}
\end{align}
to map a label sequence to a sequence of the same length as
the input sequence.
However, this function $\mathcal{P}$ returns a set of sequences,
so the objective is modified to
\begin{align}
p(y_{1:n} | x_{1:T}) = \sum_{a_{1:T} \in \mathcal{B}^{-1}(y_{1:n})} p(a_{1:T} | x_{1:T})
    = \sum_{a_{1:T} \in \mathcal{B}^{-1}(y_{1:n})} \prod_{t=1}^T p(a_t | x_t),
\end{align}
marginalizing over all the possible paths.

Given an input sequence $x$, predicting a label sequence is done by finding
\begin{align}
\argmax_{y} p(y | x) = \argmax_{y} \sum_{a \in \mathcal{B}^{-1}(y)} p(a | x).
\end{align}
However, there are currently no algorithms that can solve the above efficiently,
so it is typically approximated by finding the greedy best path
\begin{align}
\argmax_{y} p(y | x) = \argmax_{y} \max_{a \in \mathcal{B}^{-1}(y)} p(a | x)
    = \mathcal{B}(\argmax_{a} p(a|x)).
\label{eq:ctc-pred}
\end{align}
In other words, we simply find the label that achieves the highest probability
at every time point and remove the duplicates and blank symbols to produce
the final decoded sequence.

\subsection{Connection to the marginal log loss}

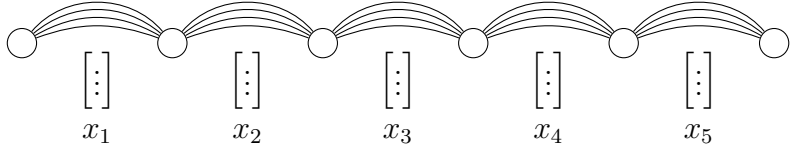
\begin{figure}
\begin{center}
\begin{tikzpicture}[ver/.style={circle,draw}]
\node [ver] (x0) at (0, 0) {};
\node [ver] (x1) at (2, 0) {};
\node [ver] (x2) at (4, 0) {};
\node [ver] (x3) at (6, 0) {};
\node [ver] (x4) at (8, 0) {};
\node [ver] (x5) at (10, 0) {};

\draw (x0) edge [out=20,in=160] (x1);
\draw (x1) edge [out=20,in=160] (x2);
\draw (x2) edge [out=20,in=160] (x3);
\draw (x3) edge [out=20,in=160] (x4);
\draw (x4) edge [out=20,in=160] (x5);

\draw (x0) edge [out=30,in=150] (x1);
\draw (x1) edge [out=30,in=150] (x2);
\draw (x2) edge [out=30,in=150] (x3);
\draw (x3) edge [out=30,in=150] (x4);
\draw (x4) edge [out=30,in=150] (x5);

\draw (x0) edge [out=40,in=140] (x1);
\draw (x1) edge [out=40,in=140] (x2);
\draw (x2) edge [out=40,in=140] (x3);
\draw (x3) edge [out=40,in=140] (x4);
\draw (x4) edge [out=40,in=140] (x5);

\draw (x0) edge [out=50,in=130] (x1);
\draw (x1) edge [out=50,in=130] (x2);
\draw (x2) edge [out=50,in=130] (x3);
\draw (x3) edge [out=50,in=130] (x4);
\draw (x4) edge [out=50,in=130] (x5);

\node (y0) at (1, -0.5) {$\begin{bmatrix}\vdots\end{bmatrix}$};
\node (y1) at (3, -0.5) {$\begin{bmatrix}\vdots\end{bmatrix}$};
\node (y2) at (5, -0.5) {$\begin{bmatrix}\vdots\end{bmatrix}$};
\node (y3) at (7, -0.5) {$\begin{bmatrix}\vdots\end{bmatrix}$};
\node (y4) at (9, -0.5) {$\begin{bmatrix}\vdots\end{bmatrix}$};

\node at (1, -1.2) {$x_1$};
\node at (3, -1.2) {$x_2$};
\node at (5, -1.2) {$x_3$};
\node at (7, -1.2) {$x_4$};
\node at (9, -1.2) {$x_5$};

\end{tikzpicture}
\caption{An example of the CTC search space for a five-frame utterance
    with a label set of size three (plus one blank).
} 
\label{fig:ctc}
\end{center}
\end{figure}

As a result of \eqref{eq:ctc-pred},
the search space of CTC has an edge
for every label in the label set (including the blank label)
at every time step.  Specifically, the search space $G$ includes
the edges $\{e_{\ell, t}: \ell \in L, t \in \{1, \dots, T\}\}$
with $v_{t-1} = \tail(e_{\ell, t})$ and $v_t = \head(e_{\ell, t})$.
An example is shown in Figure~\ref{fig:ctc}.
The weight of an edge $e_{\ell, t}$ is the log probability
of label $\ell$ at time $t$.  By construction, the decision
made at every time point is independent of the decision
at other time points conditioned on $x_{1:T}$.  In addition, since the probabilities
at every time point sum to one, the partition function $Z(x)$
of the search space, i.e., the sum of all the path probabilities,
is always $1$.

Recall that the marginal log loss is defined as
\begin{align}
\mathcal{L}_{\text{mll}} = -\log \sum_{p \in \mathcal{P}_{G \circ_\sigma F_y}} \exp(w(p))
    + \log \sum_{p \in \mathcal{P}_{G}} \exp(w(p))
\end{align}
where $\mathcal{P}_K$ is the set of paths in the FST $K$
and $F_y$ is the constraint FST constructed from the ground truth label sequence $y$.
Since $Z(x) = 1$ for CTC, the second term is zero.
By the definition of $\mathcal{B}$,
we construct the constraint FST $F_y$ such that it consists of the sequences of one or more labels
with zero or more blanks in between labels.
For example, for the label sequence ``\texttt{k} \texttt{ae} \texttt{t},''
the constraint FST is the regular expression
$\varnothing^*\texttt{k}^+\varnothing^*\texttt{ae}^+\varnothing^*\texttt{t}^+\varnothing^*$.
We have $\mathcal{B}^{-1}(y) = G \circ_\sigma F_y$.
In other words, the sum of path probabilities in $G \circ_\sigma F_y$ exactly matches
the CTC objective.
Because the first term matches the CTC objective and the second term is zero,
marginal log loss becomes the CTC objective for this type of search spaces
and the log probability weight function.

\section{Other Recent End-to-End Models}

Most mainstream end-to-end speech recognition models
can be broadly categorized as either frame-based models or
encoder-decoder models.
Frame-based LSTMs trained with CTC, HMMs, and some newer approaches like
the auto-segmentation criterion (ASG)~\citep{CPS2016}
fall under the first category, because these models
emit one symbol for every frame. 
Falling under the second category, encoder-decoder
models proposed by~\citep{CBSCB2015,BCSBB2016,CJLV2016}
generate labels one at a time while
conditioning on the input and the labels generated
in the past, without an explicit alignment between labels and frames.
Since frame-based models follow the same
graph search framework as segmental models, we will
focus on discussing the connection between these and segmental models.

Recall that marginal log loss requires a search space $G$
and a constraint FST $F$ to limit the search space to ground-truth labels.
To compute marginal log loss, we first compute the marginals
on $G$ for computing the partition function $Z(x)$,
and then compute the marginals on the $\sigma$-composed
FST $G \circ_\sigma F$ for computing $Z(x, y)$.
CTC, HMMs whtn trained with lattice-free MMI \citep{P+2016},
and ASG can all be seen as special cases
of this framework.

Comparing CTC to HMMs, the search space is different depending
on the HMM topology.
For example, two-state HMMs are used in~\citep{P+2016}.
Since the transition probabilities and emission probabilities
are all locally normalized, the partition function $Z(x)$
is always $1$.  The constraint FST representing the ground-truth labels
consists simply of sequences of repeating labels.
For example, for the label sequence ``\texttt{k} \texttt{ae} \texttt{t},''
the constraint FST for one-state HMMs is the regular expression $\texttt{k}^+\texttt{ae}^+\texttt{t}^+$.
For two-state HMMs, the constraint FST is the regular expression
$\texttt{k}_1\texttt{k}_2^*\texttt{ae}_1\texttt{ae}_2^*\texttt{t}_1\texttt{t}_2^*$.
With the above construction, marginal log loss applied to HMMs is equivalent
to lattice-free MMI~\citep{P+2016}.

For ASG, the search space is equivalent to that of one-state HMMs.
Instead of assuming conditional independence as in CTC,
ASG includes transition probabilities between states.
The constraint FST is identical to that of HMMs, with repeated
ground-truth labels. However, in ASG
the weights on the edges are not locally normalized, so
the partition function $Z(x)$ is not always $1$ and has to be computed.
With the above search space construction, marginal log loss
becomes ASG.

Another approach similar to CTC proposed in~\citep{G2012}
is called RNN transducers.  The search space of
an RNN transducer is the set of alignments from the speech
signal to all possible label sequences,
so the search space grows exponentially in the number of labels.
The weight function of a path in this approach relies
on an RNN, and is not decomposable as a sum of weights of the edges.
RNN transducers are trained with marginal log loss.
By the independence assumption imposed in~\citep{G2012},
the partition function $Z(x)$ is still $1$,
so we do not need to marginalize over the exponentially
large space.  During decoding, however, we still have to search
over the exponentially large space with, for example, beam search.

In view of this framework, even when using the same loss function,
i.e., marginal log loss,
segmental models and frame-based models differ in their search space,
weight functions, and how the search space is constrained
by the ground-truth labels during training.
A summary of special cases is shown in Table \ref{tbl:e2e-models}.

\begin{table}
\caption{An example instantiation of the components used in marginal log loss
    with the ground-truth sequence ``\texttt{k} \texttt{ae} \texttt{t}''
    and $T$ input frames
    for segmental models and various end-to-end models.
    The search space $L^T$ consists of sequences of $T$ labels.
    The label sets $L_1$ and $L_2$ contain labels in $L$ with subscript 1 and 2 respectively.
    The search space is denoted $G$ and the constraint FST is denoted $F_y$ in the table.}
\label{tbl:e2e-models}
\begin{center}
\begin{tikzpicture}[ver/.style={circle, inner sep=2pt}]
\node at (-1.5, 1.5) {$y$};
\node at (3, 1.5) {\texttt{k} \texttt{ae} \texttt{t}};

\node[ver] (y0) at (-0.5, 1.5) {};
\node[ver] (y1) at (0.5, 1.5) {};
\node[ver] (y2) at (1.5, 1.5) {};
\node[ver] (y3) at (2.5, 1.5) {};
\node[ver] (y4) at (3.5, 1.5) {};
\node[ver] (y5) at (4.5, 1.5) {};
\node[ver] (y6) at (5.5, 1.5) {};
\node[ver] (y7) at (6.5, 1.5) {};
\end{tikzpicture}

\vspace{1cm}

\begin{tikzpicture}[ver/.style={circle, draw, inner sep=2pt}]

\node at (-1.5, 1.5) {$L^T$};

\node[ver] (y0) at (-0.5, 1.5) {};
\node[ver] (y1) at (0.5, 1.5) {};
\node[ver] (y2) at (1.5, 1.5) {};
\node[ver] (y3) at (2.5, 1.5) {};
\node[ver] (y4) at (3.5, 1.5) {};
\node[ver] (y5) at (4.5, 1.5) {};
\node[ver] (y6) at (5.5, 1.5) {};
\node[ver] (y7) at (6.5, 1.5) {};

\draw (y0) edge[in=170, out=10] (y1);
\draw (y0) edge[in=160, out=20] (y1);
\draw (y0) edge[in=150, out=30] (y1);

\draw (y1) edge[in=170, out=10] (y2);
\draw (y1) edge[in=160, out=20] (y2);
\draw (y1) edge[in=150, out=30] (y2);

\draw (y2) edge[in=170, out=10] (y3);
\draw (y2) edge[in=160, out=20] (y3);
\draw (y2) edge[in=150, out=30] (y3);

\draw (y3) edge[in=170, out=10] (y4);
\draw (y3) edge[in=160, out=20] (y4);
\draw (y3) edge[in=150, out=30] (y4);

\draw (y4) edge[in=170, out=10] (y5);
\draw (y4) edge[in=160, out=20] (y5);
\draw (y4) edge[in=150, out=30] (y5);

\draw (y5) edge[in=170, out=10] (y6);
\draw (y5) edge[in=160, out=20] (y6);
\draw (y5) edge[in=150, out=30] (y6);

\draw (y6) edge[in=170, out=10] (y7);
\draw (y6) edge[in=160, out=20] (y7);
\draw (y6) edge[in=150, out=30] (y7);

\end{tikzpicture}

\vspace{1cm}

\begin{tikzpicture}[ver/.style={circle, draw, inner sep=2pt}]

\node at (-1.5, 1.5) {$H$};

\node[ver] (y0) at (-0.5, 1.5) {};
\node[ver] (y1) at (0.5, 1.5) {};
\node[ver] (y2) at (1.5, 1.5) {};
\node[ver] (y3) at (2.5, 1.5) {};
\node[ver] (y4) at (3.5, 1.5) {};
\node[ver] (y5) at (4.5, 1.5) {};
\node[ver] (y6) at (5.5, 1.5) {};
\node[ver] (y7) at (6.5, 1.5) {};

\draw (y0) edge[in=170, out=10] (y1);
\draw (y0) edge[in=165, out=15] (y1);
\draw (y0) edge[in=160, out=20] (y1);

\draw (y1) edge[in=170, out=10] (y2);
\draw (y1) edge[in=165, out=15] (y2);
\draw (y1) edge[in=160, out=20] (y2);

\draw (y2) edge[in=170, out=10] (y3);
\draw (y2) edge[in=165, out=15] (y3);
\draw (y2) edge[in=160, out=20] (y3);

\draw (y3) edge[in=170, out=10] (y4);
\draw (y3) edge[in=165, out=15] (y4);
\draw (y3) edge[in=160, out=20] (y4);

\draw (y4) edge[in=170, out=10] (y5);
\draw (y4) edge[in=165, out=15] (y5);
\draw (y4) edge[in=160, out=20] (y5);

\draw (y5) edge[in=170, out=10] (y6);
\draw (y5) edge[in=165, out=15] (y6);
\draw (y5) edge[in=160, out=20] (y6);

\draw (y6) edge[in=170, out=10] (y7);
\draw (y6) edge[in=165, out=15] (y7);
\draw (y6) edge[in=160, out=20] (y7);

\draw (y0) edge[in=155, out=25] (y2);
\draw (y0) edge[in=150, out=30] (y2);
\draw (y0) edge[in=145, out=35] (y2);

\draw (y1) edge[in=155, out=25] (y3);
\draw (y1) edge[in=150, out=30] (y3);
\draw (y1) edge[in=145, out=35] (y3);

\draw (y2) edge[in=155, out=25] (y4);
\draw (y2) edge[in=150, out=30] (y4);
\draw (y2) edge[in=145, out=35] (y4);

\draw (y3) edge[in=155, out=25] (y5);
\draw (y3) edge[in=150, out=30] (y5);
\draw (y3) edge[in=145, out=35] (y5);

\draw (y4) edge[in=155, out=25] (y6);
\draw (y4) edge[in=150, out=30] (y6);
\draw (y4) edge[in=145, out=35] (y6);

\draw (y5) edge[in=155, out=25] (y7);
\draw (y5) edge[in=150, out=30] (y7);
\draw (y5) edge[in=145, out=35] (y7);

\draw (y0) edge[in=140, out=40] (y3);
\draw (y0) edge[in=135, out=45] (y3);
\draw (y0) edge[in=130, out=50] (y3);

\draw (y1) edge[in=140, out=40] (y4);
\draw (y1) edge[in=135, out=45] (y4);
\draw (y1) edge[in=130, out=50] (y4);

\draw (y2) edge[in=140, out=40] (y5);
\draw (y2) edge[in=135, out=45] (y5);
\draw (y2) edge[in=130, out=50] (y5);

\draw (y3) edge[in=140, out=40] (y6);
\draw (y3) edge[in=135, out=45] (y6);
\draw (y3) edge[in=130, out=50] (y6);

\draw (y4) edge[in=140, out=40] (y7);
\draw (y4) edge[in=135, out=45] (y7);
\draw (y4) edge[in=130, out=50] (y7);

\end{tikzpicture}

\vspace{1cm}

\renewcommand{\arraystretch}{1.5}
\begin{tabular}{p{5cm}llp{3cm}}
                   & $G$       & $F_y$     & weight \\
\hline
segmental models   & $H$       & \texttt{k} \texttt{ae} \texttt{t} & FC, FCB, MLP \\
1-state HMMs
    & $L^T$     & $\texttt{k}^+\texttt{ae}^+\texttt{t}^+$
    & posteriors \\
2-state HMMs \newline \citep{P+2016}
    & $(L_1 \cup L_2)^T$
    & $\texttt{k}_1\texttt{k}_2^*\texttt{ae}_1\texttt{ae}_2^*\texttt{t}_1\texttt{t}_2^*$
    & posteriors \newline +transition \\
CTC \newline \citep{GFGS2006} 
    & $(L \cup \{\varnothing\})^T$
    & $\varnothing^*\texttt{k}^+\varnothing^*\texttt{ae}^+\varnothing^*\texttt{t}^+\varnothing^*$
    & posteriors \\
ASG \newline \citep{CPS2016}
    & $(L \times L)^T$
    & $\texttt{k}^+\texttt{ae}^+\texttt{t}^+$
    & posteriors \newline +transition \\
Gram-CTC \newline \citep{LZLS2017}
    & $(L^n \cup \{\varnothing\})^{T-n+1}$
    & $n$-grams of \texttt{k} \texttt{ae} \texttt{t}
    & posteriors
\end{tabular}
\end{center}
\end{table}

\section{TIMIT Experiments}

We compare various frame-based models trained end to end on the same
phonetic recognition task in the same setting as in Section \ref{sec:e2e-exp}.
We use 3-layer 256-unit biderectional LSTMs paired with
CTC, one-state HMM, and two-state HMMs.
We do not use transition probabilities for two-state HMMs.
We optimize marginal log loss with the corresponding weight functions
and the corresponding constraint FSTs
for CTC, one-state HMM, and two-state HMMs.
We run vanilla SGD with step size 0.1 and gradient clipping of norm 5
for 20 epochs.  Starting from the best performing model of the first
20 epochs, we run vanilla SGD for another 20 epochs with step size 0.75
decayed by 0.75 after each epoch.
The dropout rate is 0.2, and no other explicit regularizer is
used except early stopping on the development set.
The same experiments are repeated with pyramid LSTMs.
Results comparing with segmental models are shown in Table \ref{tbl:seg-vs-ctc}.

When standard LSTMs are used,
we fail to minimize the training loss for one-state HMMs and two-state HMMs
The only difference between CTC and one-state HMMs is the use of blank symbols,
suggesting that blank symbols play an important role in end-to-end frame-based models.
In particular, one-state HMMs and two-state HMMs explicitly marginalize over
segmentations in training and are thus sensitive to time, while LSTMs trained with CTC
are not.
Besides, the tail states in $L_2$ can be seen as label-dependent blank symbols.
Since two-state HMMs fail to achieve a low training loss,
having label-dependent blanks is not helpful in this case.

When pyramid LSTMs are used,
all variants of frame-based models achieve low PERs on the development set
and improve over ones with standard LSTMs.
Since the time resolution is reduced by four with pyramid LSTMs,
we hypothesize that using the pyramid LSTMs introduces a bias similar
to a minimum duration constraint, which helps
end-to-end training for frame-based models.

\begin{figure}
\begin{center}
\includegraphics[width=9cm]{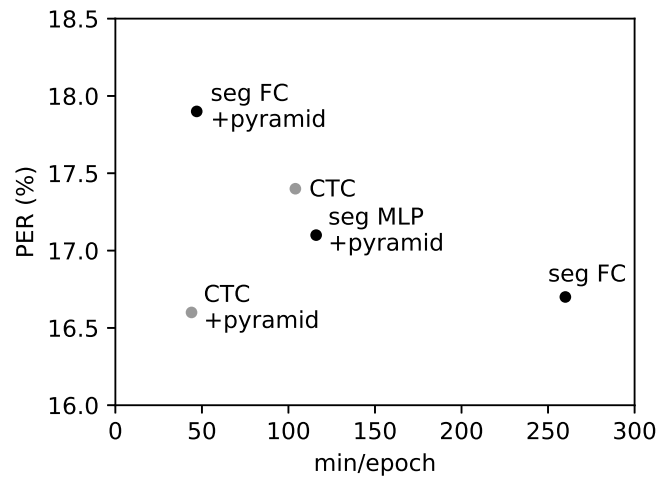}
\caption{Training minutes per epoch vs.\ PER (\%) for segmental models
    and LSTMs trained with CTC.}
\label{fig:seg-vs-ctc-train}
\end{center}
\end{figure}

The number of minutes per epoch for training segmental models and LSTMs trained with CTC
is shown in Table \ref{tbl:seg-vs-ctc-train}, and
a scatter plot showing number of minutes per epoch vs.\ PER is
shown in Figure \ref{fig:seg-vs-ctc-train}.
Pyramid LSTMs trained with CTC are the best performer while
also being the most efficient to train.
The real-time factors for decoding are shown in Table \ref{tbl:seg-vs-ctc-decode}.
Due to smaller and simpler search spaces,
frame-based LSTMs trained with CTC decode faster than segmental models.
While frame-based LSTMs train and decode faster,
segmental models are more flexible,
so they might be improved with additional feature functions.
Explicitly hypothesizing segments can serve as a prior,
so segmental models might perform better in low-resource settings.

\begin{table}
\caption{Phoneme error rates (\%) for end-to-end training from
    random initialization comparing CTC
    and segmental models trained with marginal log loss.}
\label{tbl:seg-vs-ctc}
\begin{center}
\begin{tabular}{l|ll|ll}
         & \multicolumn{2}{p{2cm}|}{regular \newline LSMT}    & \multicolumn{2}{p{2cm}}{pyramid \newline LSTM} \\
\hline
\hline
         & dev    & test   & dev      & test    \\
\hline
CTC      & 17.4   &        & 16.6     & 18.7    \\
1-state HMM
         & 82.2   &        & 17.5     &         \\
2-state HMM
         & 78.1   &        & 20.6     &         \\
\hline
seg FC   & 16.7   & 19.6   & 17.9     &         \\
seg FCB  & 17.0   &        & 17.7     &         \\
seg MLP  &        &        & 17.1     & 19.2    
\end{tabular}
\end{center}
\end{table}

\begin{table}
\caption{Average number of minutes for one epoch of end-to-end training on TIMIT.}
\label{tbl:seg-vs-ctc-train}
\begin{center}
\begin{tabular}{ll|p{1.5cm}p{1.5cm}}
       &          & regular \newline LSTM    & pyramid \newline LSTM  \\
\hline
\hline
CTC    &          & 104    & 44       \\
\hline                           
seg FC & hinge    & 119    & 38       \\ 
       & log      & 177    & 44       \\ 
       & MLL      & 260    & 47       \\ 
\hline                           
seg MLP
       & hinge    &        & 83       \\ 
       & log      &        & 110      \\ 
       & MLL      &        & 116 
\end{tabular}
\end{center}
\end{table}

\begin{table}
\caption{Real-time factors for decoding comparing CTC and segmental models.}
\label{tbl:seg-vs-ctc-decode}
\begin{center}
\begin{tabular}{l|p{1.5cm}p{1.5cm}}
          & regular \newline LSTM & pyramid \newline LSTM \\
\hline
CTC       & 0.12  & 0.07 \\
\hline
seg FC    & 0.38  & 0.12 \\
seg MLP   &       & 0.59
\end{tabular}
\end{center}
\end{table}

\section{ASL Experiments}

For American Sign Language (ASL) fingerspelling experiments,
we follow \citep{K+2017} and compare end-to-end models
in both signer-dependent setting and signer-independent setting.
The input is a sequence of 128-dimensional 
histograms of oriented gradients (HoG) feature vectors 
computed over $128 \times 128$ hand shape images.
The output is a sequence of English letters,
and the label set consists of the 26 English characters
plus two labels for initial and final non-signing regions.
The evaluation metric is the edit distance between
the predicted letter sequence and the
ground-truth letter sequence.

For the signer-dependent setting, the data for each signer
is split into ten folds, in which eight are for training,
one is for development, and one is for testing.
We report the letter error rate averaged
over the 10 experiments.
We train LSTMs with CTC and segmental models with the FCB weight
function.  Both models are trained with marginal log loss from
random initialization.
The encoder is a one-layer 128-unit bidirectional LSTM.
For CTC, we run vanilla SGD with step sizes in $\{0.1, 0.05, 0.025\}$
and gradient clipping with norm 5 for 200 epochs.
Dropout is used with a rate of 0.2,
and no other explicit regularizer is used except early stopping.
Step sizes and early stopping are tuned on the development set for each
individual experiment.  For segmental models, we
run vanilla SGD with step size 0.1 for 75 epochs.
The maximum segment length is 75 frames.
Other hyperparameters, such as dropout and gradient clipping,
are the same as for CTC.
Results are shown in Table \ref{tbl:asl-signer-dep}.
The frame-based LSTMs trained with CTC are better than Tandem HMMs,
but are behind the segmental models.
The segmental models trained end to end are on par with the
two-stage system in \citep{K+2017}, but are behind
the discriminative segmental cascades (DSC).

\begin{table}
\caption{Letter error rates (\%) for signer-dependent models averaged
     over ten folds.}
\label{tbl:asl-signer-dep}
\begin{center}
\renewcommand{\arraystretch}{1.5}
\begin{tabular}{p{4cm}|ll|ll|ll|ll|ll}
     & \multicolumn{2}{l|}{Andy} & \multicolumn{2}{l|}{Drucie}
     & \multicolumn{2}{l|}{Rita} & \multicolumn{2}{l|}{Robin}
     & \multicolumn{2}{l}{Avg} \\
     & dev   & test  & dev   & test   & dev   & test   & dev   & test  & dev  & test  \\
\hline
\citep{K+2017}
     &       &       &       &        &       &        &       &       &      &       \\
\hspace{0.3cm} Tandem HMM
     &       & 13.8  &       & 7.1    &       & 26.1   &       & 11.5  &      & 14.6  \\
\hspace{0.3cm} two-stage seg FC
     &       & 8.1   &       & 7.7    &       & 9.3    &       & 10.1  &      & 8.8   \\
\hspace{0.3cm} DSC
     &       & 7.2   &       & 6.5    &       & 8.1    &       & 8.6   &      & 7.6   \\
\hline
CTC  & 6.1   & 7.8   & 7.2   & 8.4    & 11.0  & 15.0   & 10.8  & 12.6  & 8.8  & 11.0  \\
seg FCB
     & 5.1   & 7.8   & 5.6   & 8.2    & 8.3   & 11.8   & 7.5   & 9.1   & 6.6  & 9.2
\end{tabular}
\end{center}
\end{table}

For the signer-independent setting, we train on three signers and use
the last signer for development and testing.  Specifically,
the data of the last signer is split into ten folds,
and we use two folds for development and the rest of the eight folds
for testing.  Instead of dividing the accumulated error counts
by the accumulated label sequence lengths, letter error rates are averaged
over the folds.
We compare LSTMs trained with CTC and segmental models with
FCB weight function.
The loss function and training procedure stay the same except that
we only run 40 epochs of vanilla SGD for segmental models.
Results are shown in Table \ref{tbl:asl-signer-indep}.
In this setting, CTC and segmental models trained end to end
perform significantly better than Tandem HMMs and segmental models
trained with the FC weight function in two stages.

To compare signer-independent and signer-dependent settings,
we average the letter error rates over the same eight folds.
Results are shown in Table \ref{tbl:asl-summary}.
Segmental models perform significantly better than CTC in both settings.

\begin{table}
\caption{Letter error rates (\%) for signer-independent models
     averaged over eight folds.}
\label{tbl:asl-signer-indep}
\begin{center}
\renewcommand{\arraystretch}{1.5}
\begin{tabular}{p{4cm}|ll|ll|ll|ll|ll}
     & \multicolumn{2}{l|}{Andy} & \multicolumn{2}{l|}{Drucie}
     & \multicolumn{2}{l|}{Rita} & \multicolumn{2}{l|}{Robin}
     & \multicolumn{2}{l}{Avg} \\
     & dev   & test  & dev   & test   & dev   & test   & dev   & test  & dev   & test   \\
\hline
\citep{K+2017}
     &       &       &       &        &       &        &       &       &       &        \\
\hspace{0.3cm} Tandem HMM
     &       & 54.1  &       & 54.7   &       & 62.6   &       & 57.5  &       & 57.2   \\
\hspace{0.3cm} two-stage seg FC
     &       & 55.3  &       & 53.3   &       & 72.5   &       & 61.4  &       & 60.6   \\
\hline
CTC  & 47.2  & 50.2  & 55.2  & 54.2   & 50.1  & 49.3   & 55.7  & 54.4  & 52.1  & 52.0   \\
seg FCB
     & 44.5  & 43.2  & 41.8  & 43.5   & 40.6  & 44.7   & 51.0  & 48.7  & 44.5  & 45.0
\end{tabular}
\end{center}
\end{table}

\begin{table}
\caption{Letter error rates (\%) for signer-dependent and signer-independent models averaged
     over the same eight folds.}
\label{tbl:asl-summary}
\begin{center}
\renewcommand{\arraystretch}{1.5}
\begin{tabular}{l|llll|l}
     & Andy  & Drucie & Rita   & Robin & Avg  \\
\hline
CTC  & 8.1   & 8.2    & 14.5   & 12.1  & 10.7 \\
seg FCB
     & 7.9   & 8.3    & 12.0   & 8.2   & 9.1  \\
\hline
CTC  & 50.2  & 54.2   & 49.3   & 54.4  & 52.0 \\
seg FCB
     & 43.2  & 43.5   & 44.7   & 48.7  & 45.0
\end{tabular}
\end{center}
\end{table}

\section{Summary}

We have discussed how other end-to-end frame-based
models, such as CTC, HMMs trained with lattice-free MMI, ASG, and RNN transducers are
all trained with marginal log loss.
The differences among them lie in the search spaces and
the weight functions.
Drawing these connections allows us to generalize
search spaces, loss functions, and weight functions
to a broad class of models.

From the results of comparing CTC to one-state HMMs and two-state HMMs,
we have found that the blank symbol seems to play an important
role in training LSTMs end to end.
Using pyramid LSTMs improves both the performance
and the runtime of decoding and training
for CTC, one-state HMMs, and two-state HMMs, but not for segmental models.
We have also shown that segmental models with regular LSTMs are better than regular
LSTMs trained with CTC on both phonetic recognition and ASL fingerspelling recognition.

%% file: conclusion.tex
\chapter{Conclusion and Future Work}

In this thesis, we have made the following contributions
in advancing the study of discriminative segmental models.
\begin{itemize}

\item We have proposed discriminative segmental cascades for
    incorporating rich computationally expensive features
    while maintaining efficiency.
    We use max-marginal pruning to reduce the size of search spaces,
    generating sparse lattices while having low
    oracle error rates.
    We obtain improved performance over most earlier work while greatly
    improving efficiency.

\item We have explored the space of losses, multi-stage training, and end-to-end training
    for segmental models with various losses and weight functions.
    We have shown that segmental models trained with multi-stage training
    can serve as a good initialization for end-to-end training.

\item We have presented a unified framework including many end-to-end
    models, such as hidden Markov models, connectionist temporal classification,
    and segmental models, as special cases.
    Drawing this connection allows us to design general search spaces,
    loss functions, and weight functions applicable to models in a broad class.

\item We achieve competitive results on phonetic recognition and ASL fingerspelling recognition
    with a segmental model trained end to end.
    Earlier work (not reported in this thesis) by \citep{K+2017} has obtained the best
    reported results to date using our discriminative segmental cascades on signer-dependent
    fingerspelling recognition.

\end{itemize}

In this chapter, we discuss some potential future work extending
segmental models to large-vocabulary tasks and to unsupervised settings.
Another potential direction for future work is even richer feature functions.

\section{Word Recognition}

Since first-pass segmental models are computationally demanding,
it is very slow to train and decode
segmental models with label sets of size in the order of 10,000.
This poses a challenge for tasks with large label sets,
such as word recognition.
\cite{Z+2010, MXJN2015} bypassed this difficulty by using a baseline HMM recognizer
to generate word lattices,
and used segmental models to rescore these lattices, exploring various word-level features.
However, the performance is constrained by the
quality of the baseline recognizer and the quality of the lattices.

First-pass segmental models have previously been successfully
applied to word recognition \citep{G2003,HF2015}.
This previous work treats first-pass segmental models as a drop-in replacement
for HMM phoneme recognizers, because both models serve as
functions that map acoustic features to phoneme strings.
The phoneme recognizers are then composed with a lexicon and a language model
to form a word recognizer, as described in Section \ref{sec:fst-rec}.
This is also an option for extending our work to word recognition.

Recent work has explored models that directly predict characters,
avoiding the need for a lexicon~\citep{GJ2014, MGM2015}
but still allowing for improved performance when constraining the search space
with a lexicon (through FST composition)~\citep{MGM2015}.
Segmental models can also be used to predict characters
by changing the label set, but might be hampered by the poor alignment
of characters to acoustics.  Syllables might be a better option
than characters.

Instead of using intermediate discrete representations, such as
phonemes or characters, recent advances in computing power
have made it feasible to directly predict words~\citep{MMONN2012,BH2014,SLS2016,ARSPN2017}.
In this case, rather than using a pronunciation dictionary,
only a list of words is needed for decoding.
Segmental models can also be used to directly predict
words by using the list of words as the label set.
This approach is worth exploring further, although efficiency issues
make it nontrivial to train such models~\citep{ARSPN2017}.

\section{Unsupervised Sequence Prediction}

We have presented segmental models for
supervised sequence prediction in this thesis.
Our framework can be extended to unsupervised sequence prediction.
The goal in this setting is typically grouping segments of varying length
into clusters.
We define unsupervised sequence prediction as
finding a function that maps an input sequence
$x_1, \dots, x_T$ into a sequence of segments
$(\ell_1, s_1, t_1), \dots, (\ell_n, s_n, t_n)$
where $\ell_i \in L$ for $i=1, \dots, n$ and some label set $L$,
and $s_1 = 1$, $t_n = T+1$, $s_i < t_i$, $t_{i-1} = s_i$
for $i = 1, \dots, n$,
given a data set $S$ of sequences without labels.
Since we do not have labels for the data samples,
the goal in general is to design a loss function
$\mathcal{L}(\Theta; x)$
that only depends on the input sequence $x$,
or even more generally a loss function
$\mathcal{L}(\Theta; S)$ that depends on the
data set $S$.
The label set $L$ is typically predefined
to be just a set of identifiers,
and the correspondence between the input sequence
and the identifier sequence is learned from a data set.
As a result, a label in $L$ does not have a predefined meaning
unless the user assigns a post hoc meaning
or the loss function enforces one.

Generative segmental models, such as hidden semi-Markov models,
can be naturally extended to the unsupervised setting
by maximizing the marginal likelihood of the data.
For example, Bayesian segmental models have been applied
to small-vocabulary \citep{KJG2015}
and larger-vocabulary \citep{KJG2017} word discovery.
Viterbi-style training for such models has also been explored \citep{KLG2017}.
Recently, \cite{TBVMK2016} proposed unsupervised neural HMMs,
which can be easily extended to hidden semi-Markov models.
In fact, \cite{DDZLS2017} proposed a neural version of hidden semi-Markov models
similar to \cite{TBVMK2016}.
All the above approaches aim to maximize the marginal likelihood of the data.

Besides generative models, approaches for training log-linear models
in an unsupervised fashion, notably contrastive estimation \citep{SE2005,PCT2009},
noise-contrastive estimation \citep{GH2010}, and auto-encoders \citep{ADS2014},
have also been extensively studied.
These approaches are designed for discriminative models,
and can be readily applied to our segmental models.

A segmental model trained in an unsupervised fashion
can be used for segment clustering.
One major application of segment clustering
is for lexical unit discovery.
There is a long history of lexical unit discovery from
acoustic signals \citep{B1999}, sometimes called
spoken term discovery \citep{PG2008,JCH2010}.
Much of the recent work relies on dynamic time warping (DTW) \citep{JV2011,CTJH2011}
for spoken term discovery, with \cite{LOG2015} being out of the few exceptions who
took a nonparametric Bayesian approach.
DTW has also been found to help lexical discovery in an HMM system \citep{WKHR2013}.
Other related tasks that we do not cover here
include unsupervised word segmentation \citep{GGJ2006},
unsupervised part-of-speech tagging \citep{M1994}, and unsupervised dependency
parsing \citep{KM2004}.
Though segmental models were not used in the above studies,
many ideas can be borrowed and help advance the study of segmental models
in unsupervised settings.

The assumption that input sequences can be decomposed into
a sequence of segments serves as a strong inductive bias,
and it is particularly useful in unsupervised settings when little
is assumed about the data.
Segmental models are more flexible than frame-based models when
additional assumptions, such as the form of the feature functions, are needed.
Therefore, we believe segmental models have the potential to perform well
in these unsupervised settings.